\documentclass[10pt]{article}
\usepackage[preprint]{tmlr}
\usepackage{tcolorbox}

\usepackage{amsmath,amsfonts,bm}

\def\eqref#1{equation~\ref{#1}}

\def\1{\bm{1}}

\DeclareMathAlphabet{\mathsfit}{\encodingdefault}{\sfdefault}{m}{sl}
\SetMathAlphabet{\mathsfit}{bold}{\encodingdefault}{\sfdefault}{bx}{n}

\usepackage[utf8]{inputenc}
\usepackage[T1]{fontenc}
\usepackage{hyperref}
\usepackage{url}
\usepackage{booktabs}
\usepackage{amsfonts}
\usepackage{nicefrac}
\usepackage{microtype}
\usepackage{xcolor}
\newcommand{\cbmbench}{\textsc{CBM-Bench}}

\usepackage{microtype}
\usepackage{graphicx}
\usepackage{booktabs}
\usepackage{multirow}
\usepackage{longtable}
\usepackage{tabularx}
\usepackage{makecell}
\usepackage{caption}
\usepackage{subcaption}        
\usepackage{array}
\usepackage[dvipsnames]{xcolor}
\usepackage{xspace}
\usepackage{etoc}
\usepackage{placeins}
\usepackage{wrapfig}
\usepackage[accsupp]{axessibility}
\usepackage{longtable}
\usepackage{amsmath}
\usepackage[nameinlink,capitalize]{cleveref}
\usepackage{amssymb}
\usepackage{mathtools}

\usepackage{algorithm}
\usepackage{algorithmic}

\usepackage{sidecap}
\title{\cbmbench{}: Hitchhiker's Guide to Evaluating Concept Bottleneck Models}

\author{\name Eyad Alshami \email ealshami@mpi-inf.mpg.de \\
      \addr Max Planck Institute for Informatics, Saarland Informatics Campus, Germany \\
      RTG Neuroexplicit Models of Language, Vision, and Action, Saarbrücken, Germany
      \AND
      \name Shashank Agnihotri \email shashank.agnihotri@uni-mannheim.de \\
      \addr Data and Web Science Group, University of Mannheim, Germany
      \AND
      \name Wolfgang Stammer \email wolfgang.stammer@mpi-inf.mpg.de \\
      \addr Max Planck Institute for Informatics, Saarland Informatics Campus, Germany
      \AND
      \name Margret Keuper \email keuper@mpi-inf.mpg.de \\
      \addr Max Planck Institute for Informatics, Saarland Informatics Campus, Germany \\
      Data and Web Science Group, University of Mannheim, Germany
      \AND
      \name Jonas Fischer \email jonas.fischer@mpi-inf.mpg.de \\
      \addr Max Planck Institute for Informatics, Saarland Informatics Campus, Germany
      \AND
      \name Bernt Schiele \email schiele@mpi-inf.mpg.de \\
      \addr Max Planck Institute for Informatics, Saarland Informatics Campus, Germany \\
      RTG Neuroexplicit Models of Language, Vision, and Action, Saarbrücken, Germany}

\newcommand{\response}[1]{#1}

\def\month{MM}
\def\year{YYYY}
\def\openreview{\url{https://openreview.net/forum?id=XXXX}}

\begin{document}

\maketitle

\begin{figure*}[th]
    \vspace{-.2cm}
    \centering

    \includegraphics[width=0.32\textwidth]{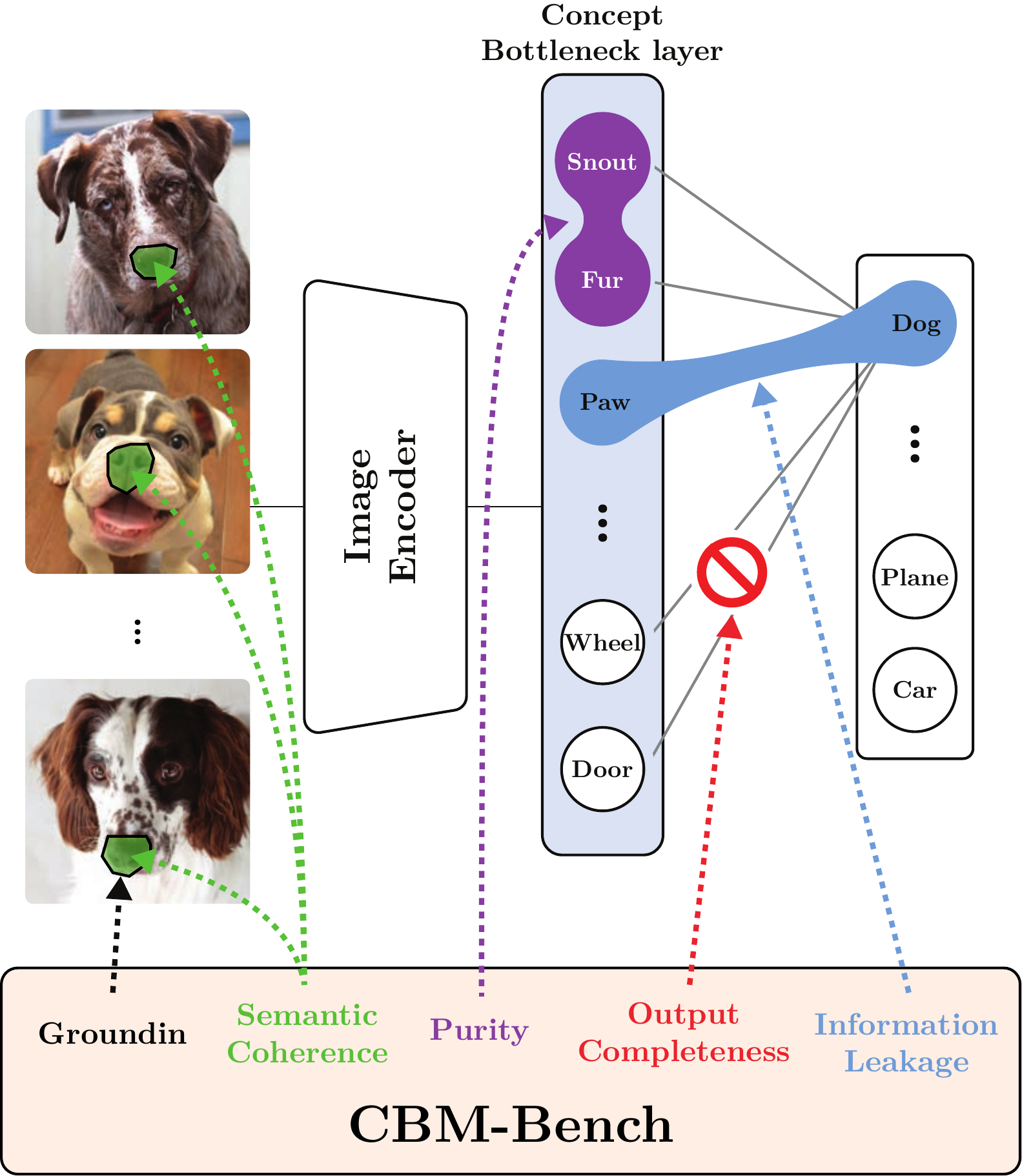}
    \hfill
    \includegraphics[width=0.65\textwidth]{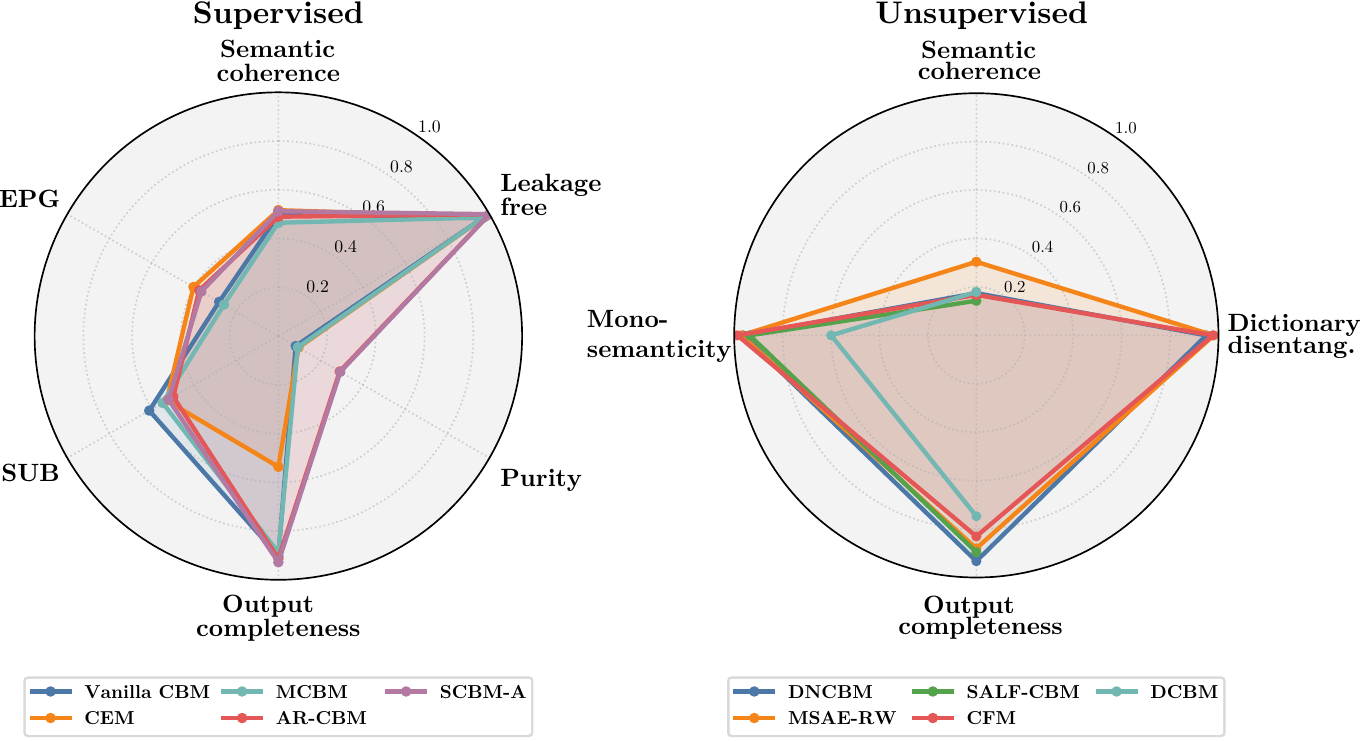}

    \caption{ \textbf{CBM-Bench reveals hidden trade-offs in concept-based models.}
Left: CBM-Bench evaluates supervised CBMs across five key properties: spatial concept grounding, semantic coherence, concept purity, information leakage, and output completeness. For unsupervised CBMs, it evaluates semantic coherence, and output completeness, introdues visual polysemanticity, and further evaluates dictionary disentanglement for SAE-based models.
Right: Results on CUB-200 and Pascal-Part on supervised and unsupervised CBMs, respectively, with higher values indicating better performance.
}
    \label{fig:intro}
    \vspace{-.18cm}
\end{figure*}

\begin{abstract}
Concept Bottleneck Models (CBMs) aim to make predictions explainable and interventional by routing decisions through human-understandable concepts without sacrificing performance. Recent studies have revealed failure modes that undermine these promises, including information leakage and weak spatial grounding. However, CBMs are still most often judged by concept and task accuracy alone, while complementary properties that probe these failure modes are evaluated only partially and without a unified protocol. Thus, we introduce \cbmbench{}, a unified suite of metrics for evaluating CBMs beyond accuracy. For supervised CBMs, \cbmbench{} evaluates five key properties: (i) spatial grounding and locality, (ii) information leakage, (iii) purity of learned concepts, (iv) output completeness, and (v) semantic coherence. For unsupervised CBMs, \cbmbench{} evaluates four complementary properties: (i) output completeness, (ii) semantic coherence, (iii) polysemanticity, and (iv) dictionary disentanglement. Beyond unifying existing metrics, \cbmbench{} introduces new metrics that better capture the purity and monosemanticity of concepts. Applying these metrics to representative supervised and unsupervised CBMs reveals failure modes and trade-offs that are not visible from concept and task accuracy alone. In supervised models, strong concept and task accuracy can coincide with weak spatial grounding, low semantic coherence, and information leakage. In unsupervised models, \cbmbench{} highlights the advantages of SAE-based CBMs in discovering more monosemantic and semantically coherent concepts. Overall, our work offers a practical, extensible suite of metrics for analyzing CBM behavior beyond accuracy, enabling interpretability claims to be supported by a standardized suite of complementary metrics. \textbf{Code is available at:} \response{\href{https://github.com/eyad-al-shami/CBM-Bench}{github.com/eyad-al-shami/CBM-Bench}}
\end{abstract}

\section{Introduction}
\label{sec:intro}

In \textit{The Hitchhiker’s Guide to the Galaxy}, the supercomputer ``Deep Thought'' famously produces the answer ``42'' without an explanation anyone can interpret. Modern deep learning systems often behave similarly: they make highly accurate predictions, yet their internal reasoning remains opaque. Concept Bottleneck Models (CBMs) \citep{losch-gcpr20, cbm, losch2021semantic, knab2026survey} are proposed as an explicit remedy to this black-box decision-making by routing predictions through human-interpretable concepts. This makes the decision process auditable by design. However, we still lack a rigorous benchmark to evaluate whether these concept representations truly improve interpretability.

CBMs in principle explain their predictions by first predicting human-defined concepts from the input and then using only these concepts for the final task prediction. This two-stage design aims to enable explainability through named concepts and interventions by allowing users to correct mispredicted concepts at test time. 
These properties are meaningful only if the predicted concepts are themselves trustworthy: they should correspond to the intended visual evidence, represent semantically distinct features, and avoid hiding additional task-relevant information. Recent work shows that CBMs can fail in ways not visible from accuracy alone, including \textit{weak spatial grounding} \citep{respect_locality, learn_as_intended, trust_worthiness}, \textit{information leakage} \citep{leakage, makonnen2025measuring, leakage_first_paper}, and \textit{entangled concept representations} \citep{was_never_cbm}. 
As a result, a CBM can achieve high concept accuracy while violating the assumptions needed for trustworthy explanations and interventions. Although information leakage and spatial grounding have been studied in early CBMs \citep{leakage_first_paper, learn_as_intended}, they have not been systematically adopted in later CBM evaluations, which continue to rely mainly on task and concept accuracy. In the absence of a unified suite of metrics, CBM evaluation remains partial and often lacks rigorous experiments that test the extent to which the key properties required for interpretability are satisfied.

We introduce \cbmbench, a suite of metrics for evaluating whether CBMs satisfy key properties required for reliable concept-based explanations. For supervised CBMs, where a ground-truth concept vocabulary is available, \cbmbench{} combines adapted metrics from prior work with new measures introduced in this paper. Specifically, we measure \textit{(i)} \textit{spatial grounding and locality}, testing whether concepts are grounded in the right visual evidence and remain local, using EPG \citep{epg} and SUB \citep{sub}, \textit{(ii)} \textit{information leakage}, testing whether additional task-relevant information leaks through the bottleneck, using concept-to-label leakage (CTL) \citep{leakage}, \textit{(iii)} \textit{purity}, introduced in this work, testing whether concept predictors remain semantically distinct, using a variation-of-information-based score, \textit{(iv)} \textit{output completeness}, testing whether the final prediction relies on a compact set of concepts, using concepts insertion and deletion AUC \citep{co_12}, and \textit{(v)} \textit{semantic coherence}, introduced in this work, testing whether concept-attributed regions are semantically coherent across images. To demonstrate the value of these metrics, we apply them to a representative set of supervised CBMs: Vanilla CBM~\citep{cbm}, CEM~\citep{cem}, MCBM~\citep{was_never_cbm}, AR-CBM~\citep{leakage_first_paper}, and SCBM~\citep{scbm} on CUB-200-2011~\citep{cub2002011}, PartImageNet~\citep{partimagenet}, and Pascal-Part~\citep{everingham2010pascal}.

Beyond supervised CBMs, the same need for systematic evaluation extends to unsupervised CBMs, which have gained increasing attention as large-scale, foundation model-applicable alternatives that do not require costly concept labels for training. Their evaluation requires different criteria, since no ground-truth concept vocabulary is available and concept discovery is open-ended. Therefore, \cbmbench{} adapts its evaluation to this setting using four complementary metrics: \textit{(i)} \textit{Output Completeness}, testing whether the final prediction relies on a compact set of discovered concepts, \textit{(ii)} \textit{Semantic Coherence} (introduced in this work), testing whether concept-attributed regions are visually consistent, \textit{(iii)} \textit{Visual Polysemanticity} (introduced in this work), testing the semantic diversity of discovered concepts, and \textit{(iv)} \textit{Dictionary Disentanglement}, testing geometric redundancy for SAE-based models. We apply these metrics to five representative unsupervised CBMs: SALF-CBM \citep{salf_cbm}, DN-CBM \citep{dncbm}, MSAE \citep{msae}, D-CBM \citep{dcbm}, and CFM \citep{cfm}.

Using these representative models, we identify failure modes and trade-offs that are not visible from standard accuracy analysis alone. Spatial grounding and semantic coherence are consistently weak, soft concept representations leak more task-relevant information than hard ones. However, information leakage can be reduced by targeting higher concept accuracies, and hard CBMs often suffer from degraded task performance and optimization difficulties. These trade-offs show that no single metric is sufficient and highlight specific room for improvement for future developments of CBMs.

In summary, our main contributions are:
\textbf{(1)}~\cbmbench{}, a suite of metrics for evaluating CBMs beyond accuracy, supporting both supervised and unsupervised models,
\textbf{(2)}~a set of complementary evaluation metrics for supervised and unsupervised CBMs, combining adapted measures from prior work with new metrics for purity, semantic coherence, and visual polysemanticity,
\textbf{(3)}~an empirical analysis on representative CBMs showing that high task or concept accuracy can mask important failures such as weak grounding, low coherence, leakage, redundancy, and optimization difficulties, and raising open issues and future directions for building more reliable CBMs and more informative interpretability benchmarks.

\section{Concept Bottleneck Models}
\label{sec:method}

Concept Bottleneck Models (CBMs)~\citep{cbm,losch2021semantic} decompose prediction into two stages: predicting human-interpretable concepts from raw inputs, then using only these concepts for the final prediction. This routes task-relevant information through an interpretable concept layer, enabling users to understand and interact with the decision process. CBMs are part of a broader direction in inherently interpretable modeling, which aims to make model decisions transparent by design~\citep{bohle2022b, alshami2025aim}

\vspace{1mm}
\noindent\textit{\textbf{Formal Notation. }}
Following \citep{knab2026survey}, given input space $\mathcal{X}$, embedding space $\mathcal{Z} \subseteq \mathbb{R}^D$, concept space $\mathcal{C} \subseteq \mathbb{R}^K$, and output space $\mathcal{Y}$, a CBM $f \circ h_{\mathcal{S}} \circ g : \mathcal{X} \rightarrow \mathcal{Y}$ maps $x$ to $z=g(x)$, $c=h_{\mathcal{S}}(z)$, and predicts $\hat{y}=f(c)$. While $g$ and $h_{\mathcal{S}}$ are typically complex models, $f$ is usually simple (often linear and sparse)~\citep{cbm}, enforcing a bottleneck: task-relevant information should be encoded only in concepts.

\vspace{1mm}
\noindent\textit{\textbf{Properties and Capabilities of CBMs. }}
CBMs assume a \textit{Markovian structure} where $y \perp x \mid c$~\citep{leakage_first_paper}. This supports \textit{Explainability}~\citep{cbm}, \textit{Interventions} without retraining~\citep{shin2023closer, steinmann2024learning}, \textit{Problem decomposition} into concept detection ($h_{\mathcal{S}}$) and reasoning ($f$)~\citep{cem}, and \textit{Structured debugging}~\citep{stammer2021right}. In practice, however, these benefits depend on concepts being correctly grounded, semantically faithful, sufficiently disentangled, and sufficiently covering the input for the downstream task.

\vspace{1mm}
\noindent\textit{\textbf{Concept Supervision Strategies. }}
\textit{Supervised CBMs} rely on ground-truth concept annotations: $\mathcal{S}$ is predefined and $h_{\mathcal{S}}$ is trained with direct supervision, but this requires costly expert labels. \textit{Unsupervised CBMs} do not require concept annotations and include foundation model-based and pattern discovery methods. Foundation model-based CBMs~\citep{lb_cbm, yang2023language} leverage pretrained vision-language models (e.g., CLIP~\citep{radford2021learning}) or LLMs to define and ground concepts without task-specific annotations. Pattern discovery methods~\citep{dncbm, schrodi2025selective} identify patterns through objectives such as disentanglement or dictionary learning and assign semantic labels post hoc.

\vspace{1mm}
\noindent\textit{\textbf{Challenges for CBMs. }}
Recent work highlights challenges that can undermine practical effectiveness~\citep{leakage_first_paper, feng2025bayesian, sinha2025comprehensive}: \textit{Spatial grounding and input alignment}, \textit{Concept leakage}, \textit{Concept disentanglement}, and \textit{Model sensitivity to concepts}. These properties are interdependent and often in tension: more concepts can improve completeness but hurt interpretability, stronger bottlenecks can reduce leakage but limit accuracy, and optimizing for prediction can degrade grounding quality.

\vspace{1mm}
\noindent\textit{\textbf{Current Evaluation Practices. }}
Existing evaluations typically focus on task accuracy, concept prediction quality, or intervention effectiveness in idealized settings where all concepts are corrected to ground truth~\citep{cbm, shin2023closer}. Some works also use user studies~\citep{chauhan2023interactive} or assess concept leakage~\citep{leakage_first_paper, makonnen2025measuring} and prototype-based detection~\citep{marconato2022glancenets}. However, these approaches often evaluate properties in isolation, rarely study the interplay between accuracy, concept quality, grounding, and intervention reliability, and often ignore concept set size~\citep{yan2023learning}. What is missing is a standardized evaluation framework across supervised and unsupervised settings that captures predictive performance, concept quality, and reliability under realistic conditions, enabling fair comparison and revealing CBM trade-offs.

\section{\cbmbench{}: A Unified Suite of Metrics}
\label{sec:cbmbench}

\begin{table*}[th]
\centering
\renewcommand{\arraystretch}{1.25}
\caption{\textbf{CBM-Bench criteria and metrics for supervised and unsupervised CBMs.}}
\label{tab:metric-overview}
\footnotesize
\begin{tabularx}{\textwidth}{
  >{\raggedright\arraybackslash}p{2.2cm}
  >{\raggedright\arraybackslash}p{2.7cm}
  >{\raggedright\arraybackslash}p{4.1cm}
  >{\raggedright\arraybackslash}X
}
\toprule
\textbf{Setting}
 & \textbf{Criterion}
 & \textbf{Metric(s)}
 & \textbf{What it measures} \\
\midrule

\multirow{5}{*}{\textbf{Supervised}}
 & \textbf{Spatial Grounding \& Locality}
 & Energy Pointing Game (EPG)~\citep{epg}; SUB ($\mathcal{S}^+$,\,$\mathcal{S}^-$)~\citep{sub}
 & Whether concept predictions and attributions localize within the intended image regions. \\
\cmidrule(lr){2-4}

 & \textbf{Information Leakage}
 & Concept-to-Label Leakage (CTL)~\citep{leakage}
 & Whether concept representations encode task-relevant information beyond their intended semantics. \\
\cmidrule(lr){2-4}

 & \textbf{Purity}
 & VI-based purity ($P$)
 & Whether each ground-truth concept maps mainly to one predicted concept. \\
\cmidrule(lr){2-4}

 & \textbf{Output Completeness}
 & Insertion/Deletion AUC; AOPC~\citep{co_12}
 & Whether the concepts capture enough task-relevant evidence for prediction. \\
\cmidrule(lr){2-4}

 & \textbf{Semantic Coherence}
 & Average pairwise cosine similarity $S(c_k)$ of concept-specific ROI embeddings
 & Whether concept-attributed regions repeatedly localizes visually similar evidence across images with the same ground-truth concept. \\
\midrule

\multirow{4}{*}{\textbf{Unsupervised}}
 & \textbf{Concept Polysemanticity} & Visual Polysemanticity Score (VPS) & How visually diverse the evidence for a learned concept is; lower values indicate more visually monosemantic concepts. \\
\cmidrule(lr){2-4}

 & \textbf{Output Completeness}
 & Label-free insertion/deletion AUC
 & Whether the learned concepts capture enough task-relevant evidence for prediction. \\
\cmidrule(lr){2-4}

 & \textbf{Semantic Coherence}
 & Label-free semantic coherence $S_{\mathrm{LF}}(c)$
 & Whether concept attributions repeatedly highlight visually similar evidence across images with the same ground-truth concept. \\
\cmidrule(lr){2-4}

 & \textbf{Dictionary Geometry}
 & Dictionary Disentanglement
 & Whether SAE decoder dictionary columns are geometrically distinct rather than duplicate or near-duplicate directions. \\

\bottomrule
\end{tabularx}\vspace{-1em}
\end{table*}

To address the evaluation gaps identified in \autoref{sec:method}, we introduce \cbmbench{}, a suite of metrics for comprehensive CBM evaluation. Since supervised and unsupervised CBMs differ in how they detect concepts and train the concept bottleneck layer, we organize the benchmark by setting. Specifically, supervised metrics assess alignment with annotated concepts, whereas unsupervised metrics evaluate the discovered concepts directly, without assuming a fixed concept vocabulary. \autoref{tab:metric-overview} summarizes the properties evaluated in \cbmbench{} and the metrics used to assess them, which are described in detail in the following sections.

\response{Overall, we organize CBM-Bench around complementary properties that concept and task accuracy do not capture. These properties provide a diagnostic profile rather than an exhaustive definition of interpretability or a universal model ranking.}

\subsection{Supervised Metrics}
\label{subsec:cbmbench-supervised-metrics}

\subsubsection{Grounding and Spatial Localization}
\label{subsec:existing_metrics}

A key requirement for CBMs is that concept predictions are grounded in the image regions that actually express the corresponding concepts. However, learned concepts may instead depend on spurious or semantically unrelated evidence~\citep{learn_as_intended}, which undermines trustworthiness~\citep{respect_locality, trust_worthiness}. Rather than relying on perturbation-based locality metrics~\citep{respect_locality}, which can introduce artifacts, we use three complementary evaluations: SUB tests sensitivity to controlled concept edits~\citep{sub}, EPG tests overlap between concept attributions and annotated regions~\citep{epg}, and semantic coherence tests whether concepts repeatedly highlight similar visual evidence across images.

\vspace{2mm}
\noindent\textit{\textbf{SUB \citep{sub}.}}
\label{paragraph:sub}
SUB evaluates concept predictions of CUB-trained models under the synthetic, fine-grained attribute substitutions introduced in \citep{sub}. We report the two SUB scores $\mathcal{S}^{+}$ and $\mathcal{S}^{-}$, which measure whether the model detects the substituted-in attribute and suppresses the removed attribute, respectively. We defer the full setup and formal definition to Appendix~\ref{appendix:metrics-sub}.

\vspace{2mm}
\noindent\textit{\textbf{Energy Pointing Game (EPG) \citep{epg}. }}
\label{paragrph:epg}
The Energy Pointing Game (EPG) score uses attribution methods \citep{gradcam, IxG,integrated_gradient}) to measure the concentration of attribution of a concept in an annotated area. 
Formally,
let  $A_k(x)\in \mathbb{R}^{H\times W}$ be the attribution map for concept $k$ on image $x$, and $M_k(x)\in\{0,1\}^{H\times W}$ the ground-truth region for that concept (e.g., an annotated mask). Taking the positive attribution values $A_k^+(x)=\max(A_k(x),0)$, the EPG score is
$
\mathrm{EPG}_k(x)
\;=\;
\frac{\sum_{h,w} M_k(x)_{h,w}\, A_k^+(x)_{h,w}}
     {\sum_{h,w} A_k^+(x)_{h,w}}
\label{eq:epg_bench} $

where higher EPG score for a concept indicates that the CBM’s evidence is more tightly concentrated on the concept’s relevant region.

\vspace{2mm}
\noindent\textit{\textbf{Semantic coherence.}}
\label{paragrph:sc}
Semantic coherence measures whether regions attributed to a predicted concept are \emph{semantically consistent across images containing that concept}. Since this requires comparing visual evidence across different images, we embed the attributed regions in a semantic feature space rather than comparing raw pixels or masks. We use DINOv2~\citep{dinov2} features for this purpose because they provide strong semantic visual representations. We compute concept-specific region embeddings by attribution-weighted pooling. Each concept is scored by the average pairwise cosine similarity of these embeddings across images, where higher values indicate more consistent visual semantics. The full definition is given in Appendix~\ref{appendix:metrics-semantic-coherence}.

\subsubsection{Information Leakage}
\label{paragraph:ctl}
A CBM exhibits information leakage when its concept bottleneck carries predictive information beyond the human-interpretable concepts it is meant to represent.
Following \citep{leakage}, we evaluate information leakage using the Concept-to-label leakage (CTL) score, which quantifies excess label information in a predicted concept $c_i$ relative to its ground-truth counterpart $c_i^*$. We compute the concept-wise score
\begin{equation}
\mathrm{CTL}_i
=
\max\!\left(
0,\
\frac{\mathrm{I}(c_i, y)}{\mathrm{H}(y)} -
\frac{\mathrm{I}(c^*_i, y)}{\mathrm{H}(y)}
\right),
\label{eq:ctl_i_bench}
\end{equation}
and report its average across concepts. Larger values indicate greater leakage.

\subsubsection{Purity}
\label{subsec:Disentanglement}
Concept purity measures whether learned concept predictors correspond to distinct ground-truth concepts.
We assess \emph{purity} using an information-theoretic metric that measures whether each ground-truth concept is captured by a unique predicted concept rather than being spread across multiple predictors. We compute variation of information between each ground-truth concept and each predicted concept, convert this distance into an agreement score, and compare the best and second-best matching predictors for each ground-truth concept. Let $A_i^{(1)}$ and $A_i^{(2)}$ denote the best and second-best agreement scores for ground-truth concept $i$. We define purity as
$$
\mathrm{Purity} = \frac{1}{K}\sum_{i=1}^{K}\left(A_i^{(1)} - A_i^{(2)}\right).
$$
A larger gap indicates a more unique correspondence, and therefore a purer concept predictor. The full definition is given in Appendix~\ref{appendix:metrics-purity}.

\subsubsection{Output Completeness}
\label{paragrph:oc}
Output completeness tests whether the concepts used by a model provide sufficient evidence to support its final class prediction, with stronger completeness indicated when this evidence is concentrated in a compact subset of the learned concepts. We measure output completeness~\citep{co_12} by evaluating how task performance changes when concepts are removed from, or inserted into, the bottleneck representation. Concepts are ranked by their utility, defined as the expected change in the true-class logit when that concept is masked. We then construct deletion curves by progressively removing concepts according to this ranking, and insertion curves by starting from a fully masked bottleneck and progressively restoring concepts. A more \emph{output-complete} model recovers quickly under insertion and degrades quickly under deletion. We summarize both curves using AUC. Full details are given in Appendix~\ref{appendix:metrics-output_completeness}.

\subsection{Unsupervised Metrics}
\label{subsec:cbmbench-unsupervised-metrics}

Evaluating unsupervised CBMs is more challenging than evaluating supervised ones, since their learned concepts are not tied to predefined ground-truth concept annotations. As a result, not all metrics used for supervised CBMs are directly applicable. We therefore focus on properties that can be assessed without one-to-one concept supervision: semantic consistency and monosemanticity of the learned concepts, as well as their output completeness. Concretely, \cbmbench{} evaluates unsupervised CBMs along four complementary axes: Semantic Coherence, Concept Polysemanticity using both the Concept Polysemantic Entropy score~\citep{chain_of_explanation} and our proposed Visual Polysemantic Score, Dictionary Disentanglement for SAE-based models, and Output Completeness.

\subsubsection{Semantic Coherence}
\label{subsubsec:cbmbench-unsupervised-semantic-coherence}
For unsupervised CBMs, semantic coherence adapts the supervised metric in \Cref{paragrph:sc} by evaluating learned concepts instead of ground-truth concepts. It asks whether the attribution maps of a discovered concept highlight semantically similar regions across the images where it is active.
We followed the same pipeline in \Cref{paragrph:sc}, while taking all the learned concepts into account. We compute attribution-weighted region embeddings and score each concept by the average pairwise cosine similarity. Higher values indicate more stable visual attributions, while lower values indicate concepts whose attributed regions vary substantially across examples. The full definition is given in Appendix~\ref{appendix:metrics-unsupervised-semantic-coherence}.

\subsubsection{Concept Polysemantic Scores}
\label{subsubsec:cbmbench-unsupervised-polysemanticity}
 Since unsupervised CBMs learn to detect concepts without explicit supervision, it is difficult to assess what semantic information each concept captures, which is typically evaluated through user studies. To address this, we use two metrics: the CPE score~\citep{chain_of_explanation} and the visual polysemanticity score. We complement CPE by suggesting the Visual Polysemanticity Score (VPS), which leverages vision foundation models instead of VLM judges to get more fine-grained and mask-independent evaluation of polysemanticity.

\vspace{1mm}
\noindent\textit{\textbf{Concept Polysemantic Entropy Score (CPE)}}
\label{subsubsec:cbmbench-unsupervised-cpe}
Semantic coherence measures whether a concept localizes consistent visual evidence across images, but it does not determine whether this evidence corresponds to a single semantic meaning. A concept may still be associated with multiple named objects or attributes. We therefore use the Concept Polysemantic Entropy score (CPE)~\citep{chain_of_explanation} to measure the semantic diversity of a concept's evidence. For each concept, we rank images by concept score, retain the top-$15$ images, extract masked regions from positive attribution maps, and use a VLM to describe them with short semantic phrases. These phrases are merged into semantic atoms, and CPE is computed as the normalized entropy of the atom distribution. Lower CPE indicates a more monosemantic concept, while higher CPE indicates broader meanings. The full definition is given in Appendix~\ref{appendix:metrics-unsupervised-cpe}.

CPE measures polysemanticity through language descriptions, but language may merge visually distinct appearances under the same phrase or miss fine-grained visual concepts. It also relies on thresholding attribution maps to generate masked evidence, making the extracted evidence sensitive to threshold choice and masking artifacts. We illustrate such cases qualitatively in \Cref{fig:uns_image_patches}. We therefore introduce the Visual Polysemanticity Score (VPS).

\vspace{1mm}
\noindent\textit{\textbf{Visual Polysemanticity Score (VPS)}}
\label{subsubsec:cbmbench-unsupervised-vps}
VPS asks whether the image regions that most strongly activate a concept form one coherent visual semantic mode or several clearly separated ones. For each learned concept, we rank images by concept score, retaining the top-$15$ images, and compute attribution-weighted DINOv2 evidence embeddings using the same pooling procedure as semantic coherence. Unlike semantic coherence, which measures average pairwise similarity among these embeddings, VPS clusters them to test whether a concept separates into multiple visually distinct evidence modes. Specifically, for each concept $c \in \mathcal{C}$ we run $k$-means with different numbers of clusters $m\in\{2,3,4,5,6\}$ and select the clustering with the highest valid silhouette score~\citep{silhouette}. The silhouette score measures how well embeddings fit their assigned cluster relative to nearby clusters.
A concept is labeled visually polysemantic when its best valid silhouette score exceeds a threshold $\tau$. Rather than relying on a single arbitrary threshold, we report VPS across a range of silhouette thresholds, with model-level VPS at each threshold defined as the fraction of concepts labeled visually polysemantic. Formally, for each concept $c$, we retain the best valid silhouette score and define the model-level score at threshold $\tau$ as
\begin{equation}
\begin{aligned}
s_c^\star
&=
\max_{m\in\mathcal{M}}
\mathrm{Silhouette}\!\left(\{z_{c,i}\}_{i=1}^{N_c},m\right),
\qquad
\mathrm{VPS}(\tau)
&=
\frac{1}{|\mathcal{C}|}
\sum_{c\in\mathcal{C}}
\mathbb{I}\!\left[s_c^\star \geq \tau\right].
\end{aligned}
\label{eq:vps_main}
\end{equation}
where $\mathcal{M}$ is the set of candidate cluster counts, and $N_c$ is the number of evidence embeddings for concept $c$.
Higher VPS values indicate that discovered concepts are more often split into multiple visually separable evidence modes. The full definition is given in Appendix~\ref{appendix:metrics-unsupervised-vps}.

\subsubsection{Dictionary Disentanglement}
\label{subsubsec:cbmbench-unsupervised-dictionary-incoherence}
For SAE-based unsupervised CBMs, each learned concept corresponds to a decoder dictionary column. Dictionary disentanglement evaluates whether these columns reconstruct distinct feature directions, since highly aligned columns indicate redundant or overlapping concepts. Inspired by dictionary self-coherence in sparse coding and dictionary learning~\citep{Sigg_2012}, we formulate a metric to measure this structural independence. Let $d_1,\dots,d_K$ denote the decoder dictionary columns, and let $\hat{d}_i=d_i/||d_i||_2$ be the normalized column. For each $\hat{d}_i$, we define its per-column disentanglement as the following:
\begin{equation}
\mathrm{DI}(i)
\;=\;
1
-
\frac{1}{K-1}
\sum_{j\neq i}
\max\left(0,\hat{d}_i^\top \hat{d}_j\right).
\label{eq:dictionary_disentanglement_per_column}
\end{equation}
The final Dictionary Disentanglement score is the mean of $\mathrm{DI}(i)$ over all decoder columns.
Values close to one indicate that decoder directions are broadly well separated under positive alignment, while lower values indicate redundant or overlapping dictionary directions. This metric should be interpreted as a geometric redundancy diagnostic rather than a direct measure of semantic interpretability. The full definition is given in Appendix~\ref{appendix:metrics-unsupervised-dictionary-incoherence}.

\subsubsection{Output Completeness}
\label{subsubsec:cbmbench-unsupervised-output-completeness}
We adapt output completeness to unsupervised CBMs by applying the same insertion and deletion protocol to the model's detected concepts. For SAE-based models, we evaluate the active concepts, while for dense models we evaluate all concepts. Concept utility is computed directly from the change in the true-class logit when a concept is masked. The resulting insertion and deletion curves test whether the discovered concepts collectively carry the evidence used for prediction and whether that evidence is concentrated in a small subset of high-utility concepts. Full details are given in Appendix~\ref{appendix:metrics-unsupervised-output-completeness}.

\section{Evaluations}
\label{sec:results}

Having introduced the metric suite, we now use representative supervised and unsupervised CBMs to illustrate the kinds of failure modes and trade-offs that \cbmbench{} can reveal beyond concept and task accuracy.

\noindent\textbf{{Experimental Setup. }}
We evaluate \cbmbench{} on three annotated datasets: CUB-200-2011~\citep{cub2002011}, PartImageNet~\citep{partimagenet}, and Pascal-Part (PASCAL VOC 2010 with part masks)~\citep{chen2014detect,everingham2010pascal}. We focus on these datasets because they provide both concept/part annotations and binary masks, enabling a consistent evaluation of all benchmark metrics. They also span different concept-to-class ratios: 56\%, 25\%, and 400\% for CUB, PartImageNet, and Pascal-Part, respectively. Although CUB-200-2011 is commonly used as a fine-grained classification benchmark, our evaluation focuses on concept-related model properties rather than fine-grained recognition performance; therefore, we treat it as a concept-annotated benchmark dataset in this study.
For Pascal-Part, we use a single-object subset filtered from the original dataset (see Appendix~\ref{appendix:experimental_settings-pascal_part} for more details).

In Section \ref{subsec:supervised}, we evaluate a representative set of supervised CBMs: Vanilla CBM~\citep{cbm}, CEM~\citep{cem}, AR-CBM~\citep{leakage_first_paper}, SCBM~\citep{scbm} (amortized and global), and MCBM~\citep{was_never_cbm}, all using a ResNet50 backbone. Results are averaged over three seeds (see Appendix for configurations). In Section \ref{subsec:unsupervised}, we evaluate a representative set of unsupervised CBMs: DN-CBM~\citep{dncbm}, MSAE-RW~\citep{msae}, SALF-CBM~\citep{salf_cbm}, D-CBM~\citep{dcbm}, and CFM~\citep{cfm}. 

\response{For the EPG score and generating the attribution maps, we use Grad-CAM~\citep{gradcam} for supervised CBMs~\citep{use_gradcam1, kim2022hive_use_gradcam2}. We compare EPG across additional attribution methods in \Cref{tab:epg-attribution-sensitivity}, with their model-ranking correlations reported in Appendix~\ref{appendix:metrics-epg}.} For unsupervised CBMs, we use Chefer LRP~\citep{chefer_lrp} for DN-CBM~\citep{dncbm}, MSAE-RW~\citep{msae}, D-CBM~\citep{dcbm} and the native attribution maps of SALF-CBM~\citep{salf_cbm} and CFM~\citep{cfm}.

\begin{table*}[t]
\centering
\begin{minipage}{\textwidth}
\centering
\caption{Task and concept accuracy across datasets and methods, averaged over three seeds. 
Best CBM performance in \textbf{bold}, second-best \underline{underlined}. 
Black-box baseline (ResNet50) provided for reference.}
\resizebox{0.9\textwidth}{!}{\begin{tabular}{@{}llccc@{\hspace{1em}}ccc@{\hspace{1em}}c@{}}
\toprule
\multirow{2}{*}{\textbf{Dataset}} &
\multirow{2}{*}{\textbf{Accuracy}} &
\multicolumn{3}{c}{\textbf{Hard CBMs}} &
\multicolumn{3}{c}{\textbf{Soft CBMs}} &
\multirow{2}{*}{\textbf{Black-box}} \\
\cmidrule(lr){3-5}\cmidrule(lr){6-8}
& & \textbf{AR-CBM} & \textbf{SCBM-A} & \textbf{SCBM-G} & \textbf{Vanilla} & \textbf{CEM} & \textbf{MCBM} & \\
\midrule
\multirow{2}{*}{CUB}
& Task & 72\% & 73\% & 73\% & 70\% & 73\% & \textbf{74\%} & 73\% \\
& Concept & \textbf{96\%} & \textbf{96\%} & \textbf{96\%} & 95\% & \textbf{96\%} & 72\% & --- \\
\midrule
\multirow{2}{*}{PartImageNet}
& Task & 9\% & 10\% & 10\% & 71\% & \underline{75\%} & 72\% & \textbf{77\%} \\
& Concept & 98\% & 98\% & 98\% & \textbf{99\%} & 98\% & 98\% & --- \\
\midrule
\multirow{2}{*}{Pascal-Part}
& Task & 58\% & 70\% & 52\% & 80\% & 81\% & \underline{83\%} & \textbf{89\%} \\
& Concept & \textbf{97\%} & 96\% & 96\% & 96\% & 96\% & 77\% & --- \\
\bottomrule
\label{tab:task_concept_acc}
\end{tabular}
}
\end{minipage}
\end{table*}

\section{Supervised CBMs}
\label{subsec:supervised}
\subsection{Task and Concept Accuracies }
\label{subsec:evaluations:accuracies}
We first report task and concept accuracies for a representative set of recent supervised CBMs~\citep{cbm, cem, leakage_first_paper, scbm} in \Cref{tab:task_concept_acc}, alongside a black-box ResNet50 baseline. On CUB, task accuracy is broadly competitive with the black-box model, whereas the gap is larger on Pascal-Part and largest on PartImageNet, where binary CBMs underperform by roughly 38\% and 67\%, respectively. On PartImageNet, this occurs despite near-perfect concept accuracy, suggesting that the concept set is insufficient for the downstream task. This is consistent with the dataset design, which provides only 40 concepts for 158 classes~\citep{leakage}; an oracle trained on binary ground-truth concepts reaches only about 12\% task accuracy. Although limiting for prediction, this makes PartImageNet a useful stress test for CBMs under incomplete concept coverage.
\begin{tcolorbox}[
  colback=cyan!8,
  colframe=cyan!55,
  boxrule=0.9pt,
  arc=2.5pt,
    before skip=4pt,
  after skip=4pt,
  left=4pt,right=4pt,top=4pt,bottom=4pt
]
\noindent\textbf{Takeaway 1.}
\emph{CBM task accuracy depends critically on the quality and sufficiency of the concept set, and therefore high concept accuracy is not, by itself, indicative of high task accuracy.}
\end{tcolorbox}

\subsection{Spatial Grounding Abilities }
\label{subsubsec:evaluations:supervised:spatial_grounding}

\begin{table}[t]
\centering
\caption{\textbf{SUB generalization is poor.} SUB $\mathcal{S}^+$ / $\mathcal{S}^-$ results. Best in \textbf{bold}, second-best \underline{underlined}. All models perform below random chance on $\mathcal{S}^+$, and higher $\mathcal{S}^+$ is often accompanied by lower $\mathcal{S}^-$, consistent with class-specific concept-vector memorization~\citep{sub}.}
\small
\resizebox{\textwidth}{!}{\begin{tabular}{lcccccc}
\toprule
\textbf{Mode} & \textbf{ARCBM} & \textbf{CBM} & \textbf{CEM} & \textbf{SCBM-G} & \textbf{SCBM-A} & \textbf{MCBM} \\
\midrule
\textbf{S+} & \underline{33.86\% (1.77)} & 28.25\% (1.48) & 32.61\% (1.53) & 30.74\% (0.98) & \textbf{33.88\% (1.07)} & 32.89\% (18.19) \\
\textbf{S-} & 50.17\% (1.78) & \textbf{61.19\% (1.20)} & 52.29\% (4.74) & 51.11\% (4.63) & 52.10\% (1.69) & 58.93\% (9.32) \\
\bottomrule
\end{tabular}
}

\label{tab:sub_grounding_results}
\end{table}
\begin{table}[t]
\centering

\caption{\textbf{Models perform similarly within each dataset, with lower coherence on datasets containing small, non-centric concepts.} Semantic coherence scores for supervised CBMs.}
\resizebox{\textwidth}{!}{\begin{tabular}{l c c c c c c}
\toprule
\textbf{Dataset} & \textbf{ARCBM} & \textbf{CBM} & \textbf{CEM} & \textbf{SCBM-G} & \textbf{SCBM-A} & \textbf{MCBM} \\
\midrule
  CUB & 49\% ($\pm$0.02) & 50\%  ($\pm$0.02) & 51\%  ($\pm$0.01) & 51\%  ($\pm$0.03) & \textbf{52\%  ($\pm$0.01)} & 46\%  ($\pm$0.02) \\
  PartImageNet & \textbf{45\%  ($\pm$0.03)} & 42\%  ($\pm$0.05) & \underline{45\%  ($\pm$0.01)} & 44\%  ($\pm$0.00) & 44\%  ($\pm$0.01) & 45\%  ($\pm$0.02) \\
  Pascal-part & \textbf{40\%  ($\pm$0.01)} & 36\%  ($\pm$0.02) & 37\% ($\pm$0.02) & 35\%  ($\pm$0.01) & 36\%  ($\pm$0.03) & 34\% ($\pm$0.02) \\
\bottomrule
\end{tabular}
}

\label{tab:semantic-coherence-supervised}
\end{table}

We next assess concept spatial grounding using SUB~\citep{sub}, and EPG~\citep{epg}.

\noindent\textit{\textbf{SUB. }} The results in (\Cref{tab:sub_grounding_results}) show that methods can not identify inserted concepts, generalizing poorly to the new concept combinations introduced. Also the absence of the substituted concept was poorly detected across the board. This aligns with findings in~\citep{sub}, suggesting that the models tend to memorize class-specific concept vectors.

\setlength{\abovecaptionskip}{0pt}
\setlength{\belowcaptionskip}{1pt}
\begin{figure}[t]
    \centering
    \includegraphics[width=0.9\textwidth]{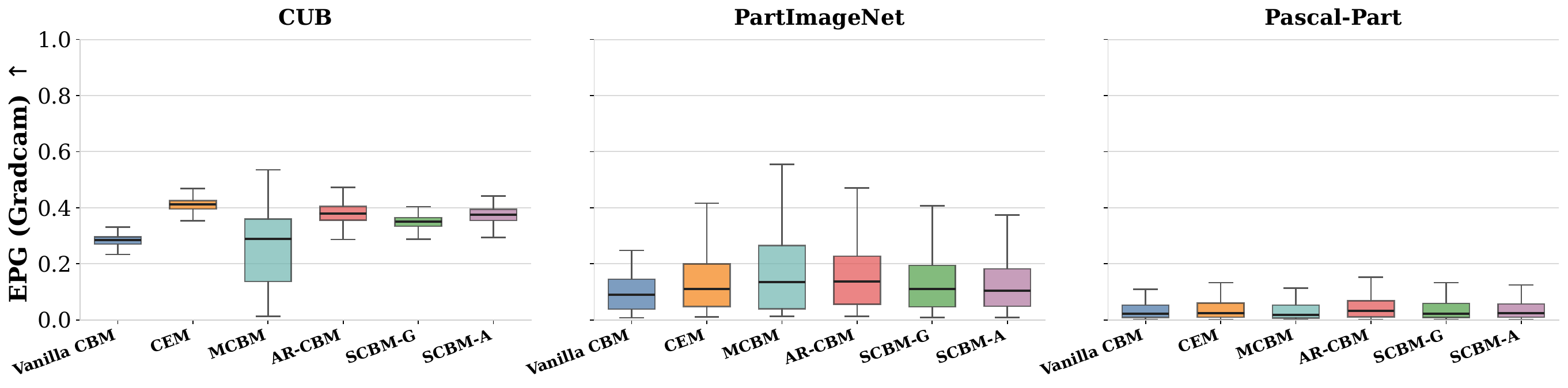}
    \caption{\textbf{Supervised CBMs show low spatial localization.} Distribution of EPG scores of the supervised CBMs across annotated concepts.}  \label{fig:concept_epg_gradcam_all_datasets}
\end{figure}

\noindent\textit{\textbf{EPG. }} \Cref{fig:concept_epg_gradcam_all_datasets} shows that high concept accuracy does not imply strong spatial localization. Despite very high concept accuracies, all models obtain low EPG scores across datasets. On CUB, binary CBMs perform similarly at around 40\%, while soft CBMs vary more, with CEM scoring highest. On PartImageNet, both hard and soft CBMs remain similarly low at around 16\%, and on Pascal-Part, all models localize poorly, with EPG scores around 5\%. (see qualitative examples in Appendix~\ref{appendix:attribution_maps}).

\response{
\noindent\textbf{Attribution-method dependence.}
EPG can be computed using any method that produces a spatial attribution map, but its value may depend on the selected method. We therefore report EPG using Grad-CAM, Guided Grad-CAM, and Input$\times$Gradient, applying each method consistently across models. The corresponding correlations between model rankings are reported in Appendix~\ref{appendix:metrics-epg}.
}

\begin{table*}[t]
\centering
\small
\setlength{\tabcolsep}{4.5pt}
\caption{\response{\textbf{EPG scores depend on the attribution method.} Model-level EPG scores obtained with three attribution methods, averaged over concepts and three seeds. Higher is better.}}
\label{tab:epg-attribution-sensitivity}
\response{\begin{tabular}{llcccccc}
\toprule
\textbf{Dataset} & \textbf{Attribution} & \textbf{CBM} & \textbf{CEM} & \textbf{MCBM} & \textbf{AR-CBM} & \textbf{SCBM-A} & \textbf{SCBM-G} \\
\midrule
\multirow{3}{*}{CUB}
 & Grad-CAM               & 0.280 & 0.403 & 0.260 & 0.376 & 0.371 & 0.344 \\
 & Guided Grad-CAM        & 0.652 & 0.676 & 0.525 & 0.646 & 0.684 & 0.663 \\
 & Input$\times$Gradient  & 0.219 & 0.288 & 0.293 & 0.271 & 0.293 & 0.258 \\
\midrule
\multirow{3}{*}{PartImageNet}
 & Grad-CAM               & 0.112 & 0.145 & 0.174 & 0.164 & 0.135 & 0.141 \\
 & Guided Grad-CAM        & 0.197 & 0.221 & 0.220 & 0.230 & 0.213 & 0.213 \\
 & Input$\times$Gradient  & 0.114 & 0.118 & 0.117 & 0.123 & 0.109 & 0.118 \\
\midrule
\multirow{3}{*}{Pascal-Part}
 & Grad-CAM               & 0.064 & 0.066 & 0.056 & 0.074 & 0.067 & 0.062 \\
 & Guided Grad-CAM        & 0.082 & 0.085 & 0.075 & 0.099 & 0.084 & 0.080 \\
 & Input$\times$Gradient  & 0.050 & 0.051 & 0.059 & 0.056 & 0.048 & 0.052 \\
\bottomrule
\end{tabular}}
\end{table*}

\response{The attribution method affects EPG magnitudes and, in some cases, model ordering. Across all three methods, EPG scores are lower on PartImageNet and Pascal-Part than on CUB.}

\noindent\textit{\textbf{Semantic Coherence. }} \Cref{tab:semantic-coherence-supervised} suggests that models are often consistent enough to achieve non-trivial semantic coherence, yet, in light of the EPG qualitative results in Appendix~\ref{appendix:attribution_maps}, they fail to focus tightly on the exact ground-truth region. The attributions are likely too broad, or coarse, and can be consistently wrong, e.g., attending to the whole object, or a larger surrounding region instead of the precise part region. This also explains the further drop on Pascal-Part, where smaller parts make the benchmark less forgiving to spatial spread and context leakage.

\begin{tcolorbox}[
  colback=cyan!8,
  colframe=cyan!55,
  boxrule=0.9pt,
  arc=2.5pt,
    before skip=4pt,
  after skip=4pt,
  left=4pt,right=4pt,top=4pt,bottom=4pt
]
\noindent\textbf{Takeaway 2.}
\emph{Despite high concept accuracies, supervised CBMs show limited spatial grounding across attribution methods and grounding metrics, struggling to detect concept substitutions and localize concepts.}
\end{tcolorbox}

\subsubsection{Purity }
\label{subsubsec:evaluations:supervised:disentanglement}
\begin{figure}[t]
    \centering
    \includegraphics[width=0.9\textwidth]{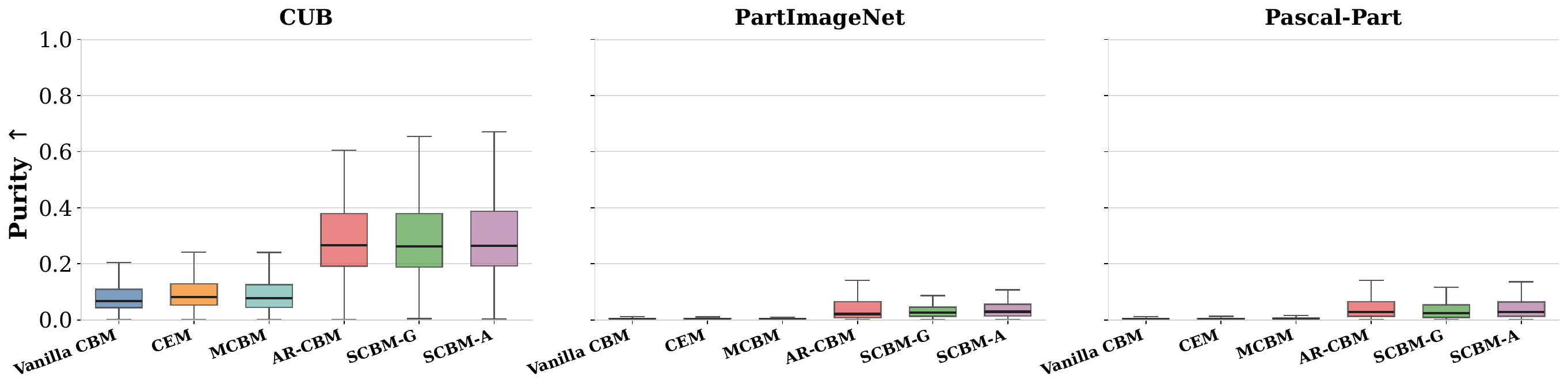}
    \caption{\textbf{Hard CBMs achieve higher purity, indicating better one-to-one alignment with ground-truth concepts.} Purity scores of supervised CBMs show substantially lower purity on datasets with coarser or spatially small concepts, suggesting reduced semantic specialization of the concept predictors.}
    \label{fig:concept_disentaglement_all_datasets}
\end{figure}

\Cref{fig:concept_disentaglement_all_datasets} shows that hard CBMs consistently obtain higher purity scores than soft CBMs across datasets. This indicates that hard CBMs learn predicted concepts that map more uniquely to individual ground-truth concepts. While the hard CBMs show better purity than the soft ones, their lower scores on PartImageNet and PASCAL-Part indicate that these part annotations are harder to recover exclusively, which is a dataset property.

\setlength{\abovecaptionskip}{0pt}
\setlength{\belowcaptionskip}{1pt}

\begin{tcolorbox}[
  colback=cyan!8,
  colframe=cyan!55,
  boxrule=0.9pt,
  arc=2.5pt,
    before skip=4pt,
  after skip=4pt,
  left=4pt,right=4pt,top=4pt,bottom=4pt
]
\noindent\textbf{Takeaway 3.}
\emph{Hard CBMs show higher concept purity than soft CBMs, indicating more distinct mappings between predicted and ground-truth concepts.}
\end{tcolorbox}

\subsubsection{Leakage }
\label{subsubsec:evaluations:supervised:leakage}
\begin{figure}[t]
    \centering
    \includegraphics[width=0.9\textwidth]{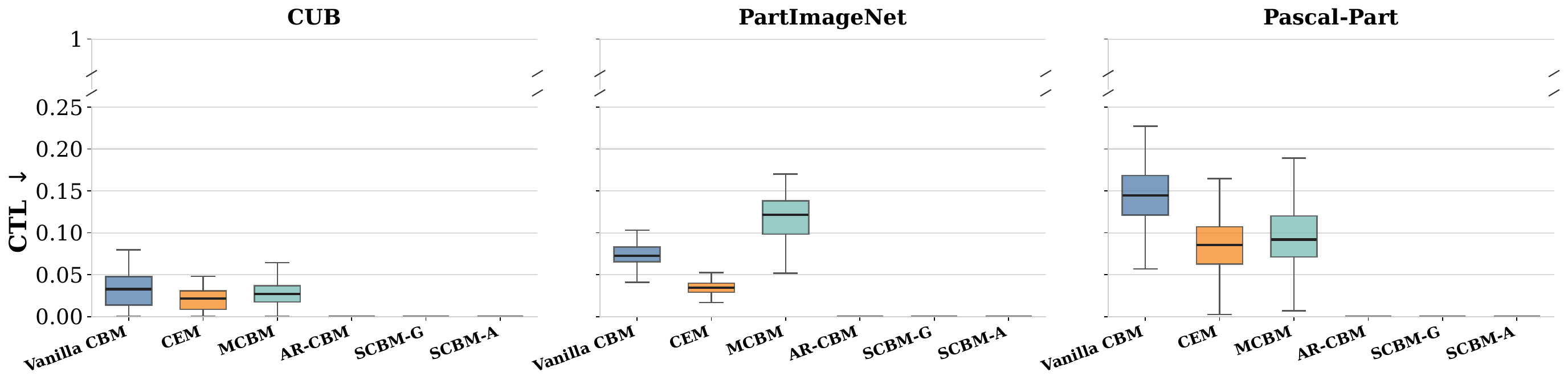}
    \caption{\textbf{Binary CMBs show no leakage.} Leakage scores measured by CTL of the supervised models across datasets. While soft CBMs leak more information, high concept accuracies can limit the leakage. While leakage-free, binary CBMs show usually much lower task performance (Table \ref{tab:task_concept_acc}).}
    \label{fig:concept_leakage_all_datasets}

\end{figure}

\Cref{fig:concept_leakage_all_datasets} shows that hard CBMs exhibit almost no leakage across datasets, whereas soft CBMs consistently attain higher CTL scores. Although stronger leakage might be expected for soft CBMs, their near-perfect concept accuracies likely limit the amount of additional task-relevant information encoded outside the bottleneck. Leakage nevertheless remains higher on PartImageNet and Pascal-Part, where a small set of coarse part annotations shared across classes makes the target class less identifiable from the concepts alone.

\begin{tcolorbox}[
  colback=cyan!8,
  colframe=cyan!55,
  boxrule=0.9pt,
  arc=2.5pt,
    before skip=4pt,
  after skip=4pt,
  left=4pt,right=4pt,top=4pt,bottom=4pt
]
\noindent\textbf{Takeaway 4.}
\emph{Consistent with their higher concept purity, binary CBMs show minimal leakage, suggesting that discrete, pure concepts limit unintended task-relevant information. Soft concept representations are less pure and leak more, especially on datasets with coarse, shared-part annotations and lower concept coverage.}
\end{tcolorbox}

\setlength{\abovecaptionskip}{0pt}
\setlength{\belowcaptionskip}{1pt}

\subsection{Output Completeness }
\label{subsubsec:evaluations:supervised:output_completeness}

\begin{figure}[t]
    \centering
    \includegraphics[width=1.0\textwidth]{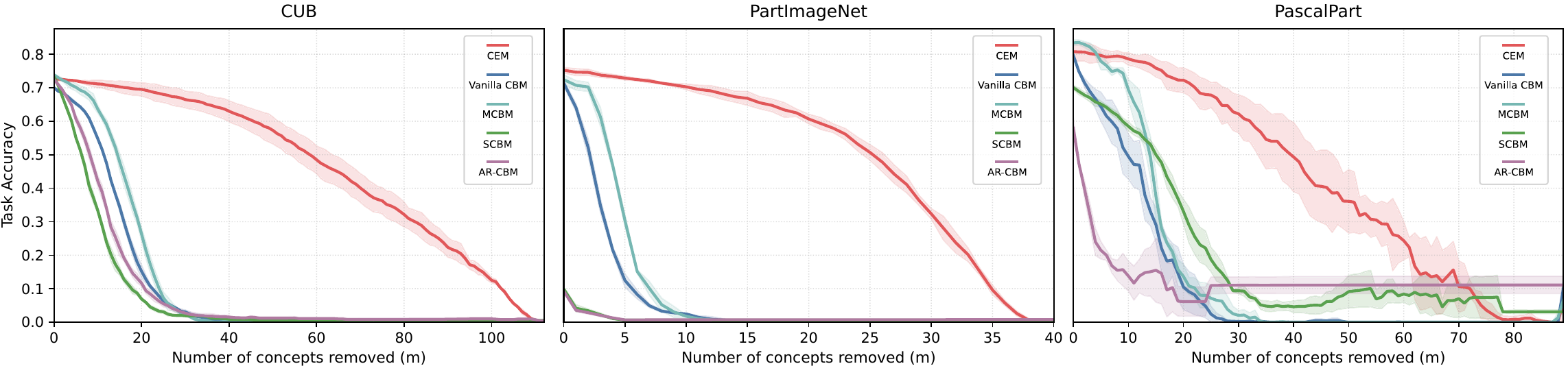}
    \caption{\textbf{Output completeness via concept deletion.} Utility-ordered concept deletion curves show how task accuracy changes as concepts are removed.}
    \label{fig:task_acc_vs_concept_insertion_deletion}
\end{figure}

\Cref{fig:task_acc_vs_concept_insertion_deletion} shows how the task performance changes while deleting concepts (for the insertion concepts see Appendic \ref{appendix:metrics-output_completeness}). On CUB dataset, the hard CBMs show sharpest decline in the performance, which is in-line with the purity results that suggests hard CBMs map more uniquely to individual ground-truth concepts (see Appendix for the metrics correlation heatmap \ref{appendix:metrics-supervised-correlation}). While the soft CBMs, show less sensitivity to the deletion of concepts with CEM being the least sensitive. The same trend goes with the PartImageNet and Pascal-Part dataset where CEM is always the least sensitive model, indicating the CEM spreads task-relevant evidence more diffusely across its concept representation. This is also shown in the table of the AUC values (see Appendix \ref{appendix:metrics-output_completeness})
\begin{tcolorbox}[
  colback=cyan!8,
  colframe=cyan!55,
  boxrule=0.9pt,
  arc=2.5pt,
    before skip=4pt,
  after skip=2pt,
  left=4pt,right=4pt,top=4pt,bottom=4pt
]
\noindent\textbf{Takeaway 5.}
\emph{Output-completeness trends align with the purity and leakage results: hard CBMs rely more sharply on a small set of purer, less leaky concepts, whereas soft CBMs distribute task-relevant evidence more broadly across a wider set of concepts.}
\end{tcolorbox}

\setlength{\abovecaptionskip}{6pt}
\setlength{\belowcaptionskip}{0pt}

\subsection{The Intervention Risk }
\label{subsubsec:evaluations:supervised:intervention_risk}
\setlength{\abovecaptionskip}{6pt}
\setlength{\belowcaptionskip}{0pt}
\begin{figure}[t]
    \centering
    \includegraphics[width=0.9\textwidth]{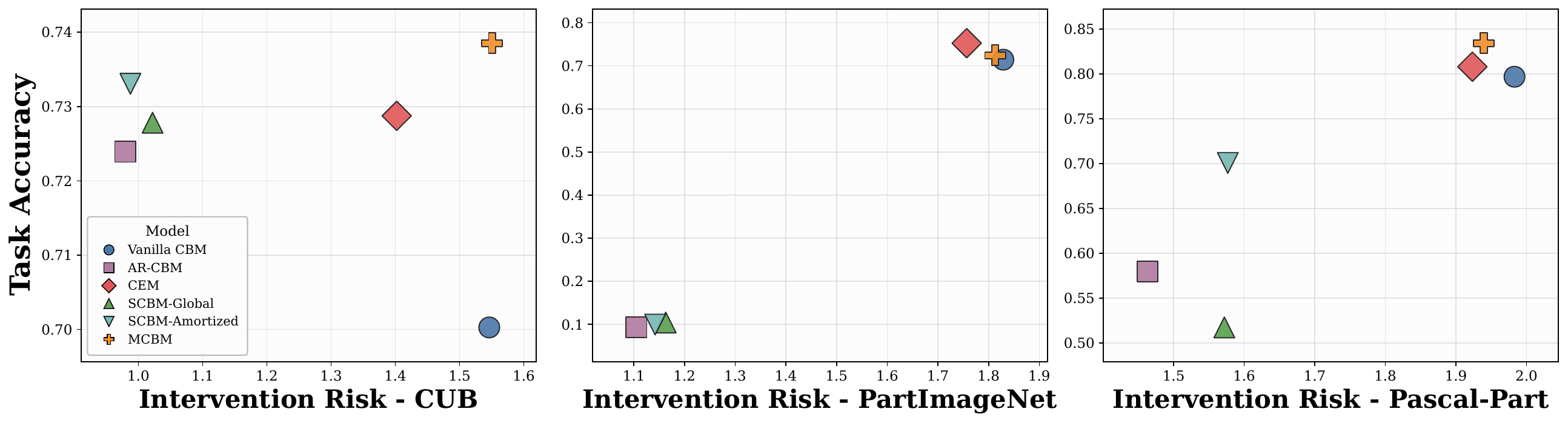}
    \caption{\textbf{Accuracy vs.\ Intervention Risk.} We aggregate three intervention-relevant failure modes into a composite score, $\text{Risk}=\text{CTL}+(1-\text{EPG})+(1-\text{Purity})$ (lower is better), and compare against task accuracy.}
    \label{fig:pareto_front_acc_vs_intervention_risk}
    \vspace{-1em}
\end{figure}

Intervention is a core promise of CBMs, but its reliability depends on how faithfully concepts are detected. We define a heuristic intervention risk as $\mathrm{IR} = \mathrm{CTL} + (1 - \mathrm{EPG}) + (1 - \mathrm{Purity})$, where lower values indicate more faithful concept-based interventions. 
\response{Notably, IR is a summary heuristic and does not replace its separately reported components. Since all terms lie in $[0,1]$, equal weighting gives a transparent default on a common scale. Application-specific weights can instead be fixed before evaluation.}
\Cref{fig:pareto_front_acc_vs_intervention_risk} shows a clear trade-off: soft CBMs improve task accuracy at the cost of higher intervention risk, while hard CBMs achieve lower intervention risk but are often accuracy-limited, notably on PartImageNet. This highlights that accuracy alone can obscure how reliably a model responds to human interventions.

\begin{tcolorbox}[
  colback=cyan!8,
  colframe=cyan!55,
  boxrule=0.9pt,
  arc=2.5pt,
    before skip=4pt,
  after skip=4pt,
  left=4pt,right=4pt,top=4pt,bottom=4pt
]
\noindent\textbf{Takeaway 6.}
\emph{Soft CBMs often trade higher task accuracy for higher intervention risk, while hard CBMs reduce risk but can be accuracy-limited in scenarios of insufficient concept coverage.}
\end{tcolorbox}

\section{Unsupervised CBMs}
\label{subsec:unsupervised}

We now move on to evaluating unsupervised CBMs, which do not rely on predefined concept labels. We evaluate them using four metrics that characterize semantic stability, polysemanticity, geometric redundancy, and output completeness.

We evaluate five representative recent models: three SAE-based models, namely DN-CBM, MSAE, and CFM, while the other two, SALF-CBM and D-CBM, use CLIP visual prompting techniques and foundational segmentation models to train the concept detection module, respectively. For the SAE-based models, the test active-concept counts are summarized in \Cref{appendix-tab:unsupervised_active_concepts}.

\subsection{Semantic Coherence}
\label{subsubsec:evaluations:unsupervised:semantic_coherence}

\begin{figure}[t] 
\centering 
\includegraphics[width=0.9\textwidth]{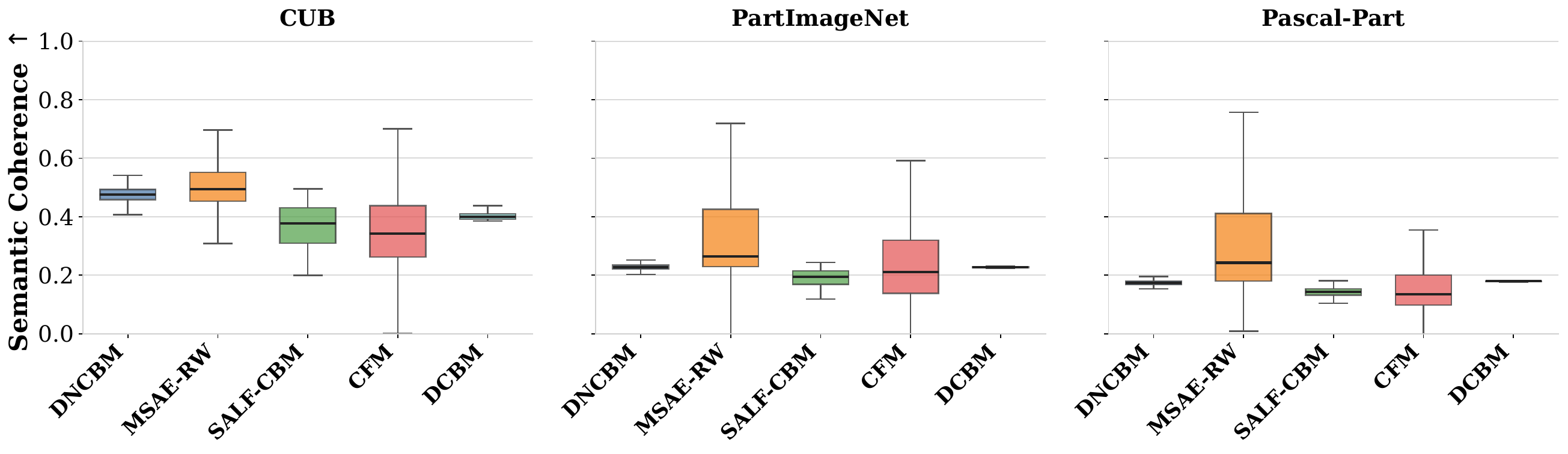} 
\caption{\textbf{Semantic coherence of unsupervised CBMs.} Semantic coherence seem to be reduced on dataset with spatially smaller concepts namely the PartImageNet and Pascal-Part.}
\label{fig:unsupervised_semantic_coherence} 
\end{figure}

\Cref{fig:unsupervised_semantic_coherence} illustrates the results of the Semantic Coherence metric, defined in \Cref{subsubsec:cbmbench-unsupervised-semantic-coherence}. Overall, semantic coherence remains low across all evaluated unsupervised models, with the MSAE model achieving the highest average coherence scores across datasets while also showing the widest variance, ranging from completely incoherent to highly coherent detected concepts. For datasets with spatially smaller concepts, namely PartImageNet and Pascal-Part, a clear trend emerges in which coherence scores drop uniformly compared to CUB (See Appendix~\Cref{app-fig:dataset_mask_ratios} for concepts' spatial size statistics). In these datasets, the Non-Matryoshka SAE-based models, namely DNCBM, SALF-CBM, and DCBM, show uniformly low semantic coherence across all concepts, while CFM and MSAE remain notable exceptions that maintain a wider spread. We attribute this sustained variance to two key architectural choices. First, the Matryoshka SAE~\citep{matryoshka} inherently captures more monosemantic concepts across multiple levels of abstraction. Second, the choice of encoder features is critical, as most models rely on the CLS token or post-pooling features, which discard finer spatial details, while CFM utilizes pre-pooling spatial features.

\begin{tcolorbox}[
    colback=green!6,
    colframe=green!50!black,
  boxrule=0.9pt,
  arc=2.5pt,
    before skip=4pt,
  after skip=4pt,
  left=4pt,right=4pt,top=4pt,bottom=4pt
]
\noindent\textbf{Takeaway 7.}
\emph{All evaluated models show weak semantic coherence, especially on datasets with spatially small concepts. Nevertheless, Matryoshka SAE-based models show, on average, better semantic coherence across datasets.}

\end{tcolorbox}

\setlength{\abovecaptionskip}{2pt}

\subsection{Concept Polysemantic Metrics}
\label{subsubsec:concept_polysemantic_monosemantic_metrics}

Semantic coherence evaluates whether the concept-attributed region remains semantically consistent across images that activate the concept. However, coherence alone does not imply monosemanticity, since a visually stable concept can still combine multiple meanings. To quantify this issue, we adopt the Concept Polysemantic Entropy score (CPE)~\citep{chain_of_explanation} and show that it yields results inconsistent with the monosemantic properties reported for the unsupervised models we benchmark~\citep{msae, cfm, dncbm}. We therefore propose a \textit{Visual Polysemanticity Score} that better reflects both user-study outcomes and the expected behavior of these models.

\begin{figure}[t] 
\centering 
\includegraphics[width=0.9\textwidth]{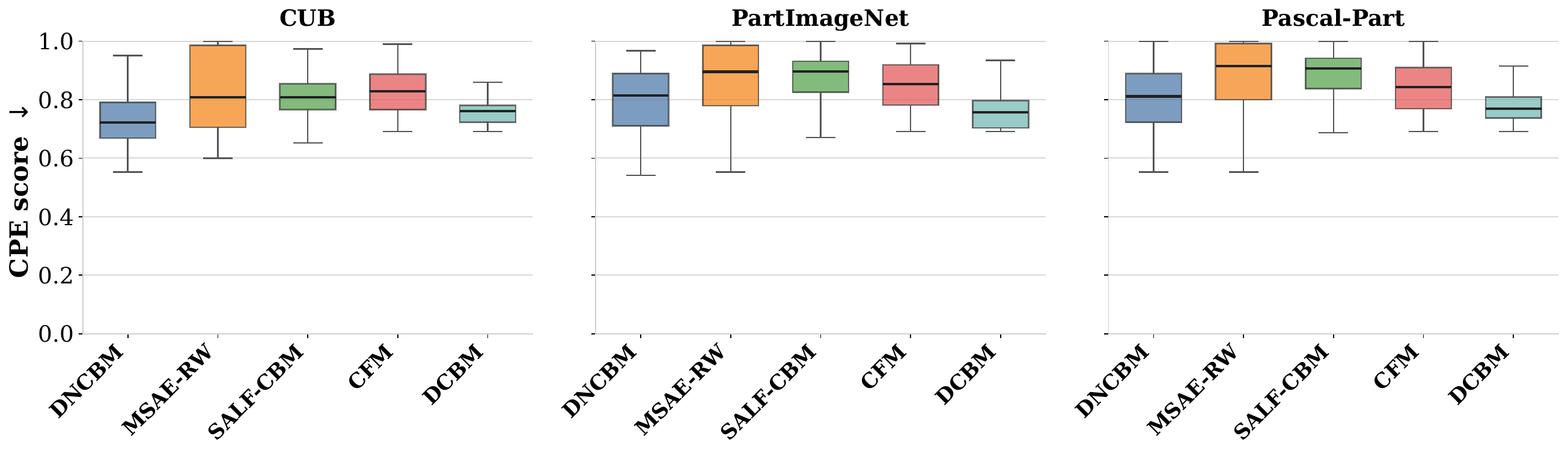} 
\caption{\textbf{Concept Polysemantic Entropy scores characterize all models as polysemantic.} The CPE score consistently contradicts user studies and results reported by the benchmarked unsupervised CBMs, showing very high polysemanticity for all methods and datasets. We allocate this to the technical limitations of this metric, motivating our visual polysemanticity score below.}
\label{fig:unsupervised_cpe} 
\end{figure}

\begin{figure}[ht]
  \centering

  \begin{subfigure}{0.49\textwidth}
    \centering
    \includegraphics[width=\linewidth]{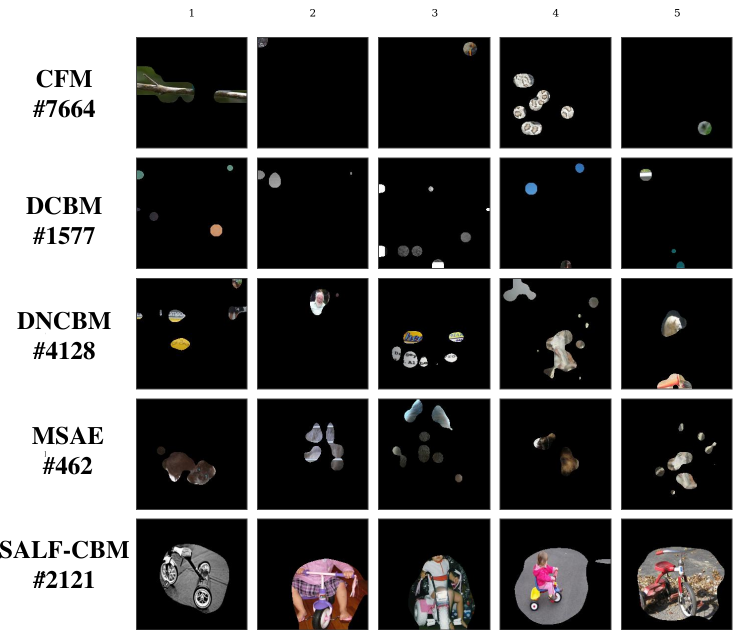}
  \end{subfigure}
  \hfill
  \begin{subfigure}{0.49\textwidth}
    \centering
    \includegraphics[width=\linewidth]{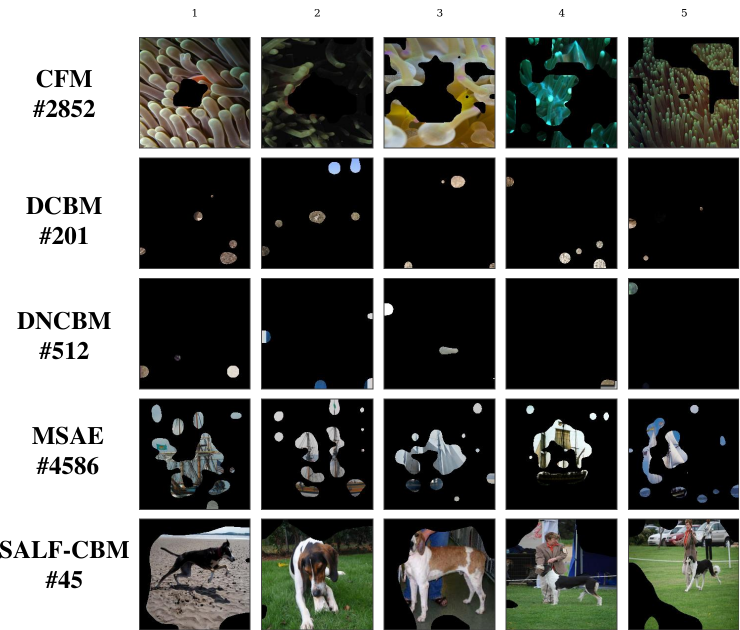}
  \end{subfigure}

  \caption{\textbf{Image patches show semantically similar detected concepts across images.} Sample of concepts across models and their top-5 activating image patches from Part-ImageNet dataset.}
  \label{fig:uns_image_patches}
\end{figure}

\noindent\textit{\textbf{Concept Polysemantic Entropy score (CPE).}}
\label{subsubsec:evaluations:unsupervised:concept_polysemantic_entropy}
Following~\citep{chain_of_explanation}, CPE quantifies the semantic entropy of VLM-generated descriptions for the top-activating image patches of each concept predictor. Higher CPE scores indicate greater polysemanticity, further details are provided in Appendix~\ref{appendix:metrics-unsupervised-cpe}.
We observe that the CPE results in \Cref{fig:unsupervised_cpe} conflict with both the reported results and user studies of the benchmarked models~\citep{msae, cfm, dncbm}, which suggest that these models are highly monosemantic. In particular, CPE ranks CFM and MSAE among the most polysemantic methods. However, qualitative examples of their top-activating patches, shown in \Cref{fig:uns_image_patches}, support the monosemantic behavior reported by the authors~\citep{msae, cfm, dncbm}.

\setlength{\abovecaptionskip}{1pt}

\setlength{\abovecaptionskip}{3pt}

\begin{figure}[t] 
\centering 
\includegraphics[width=0.9\textwidth]{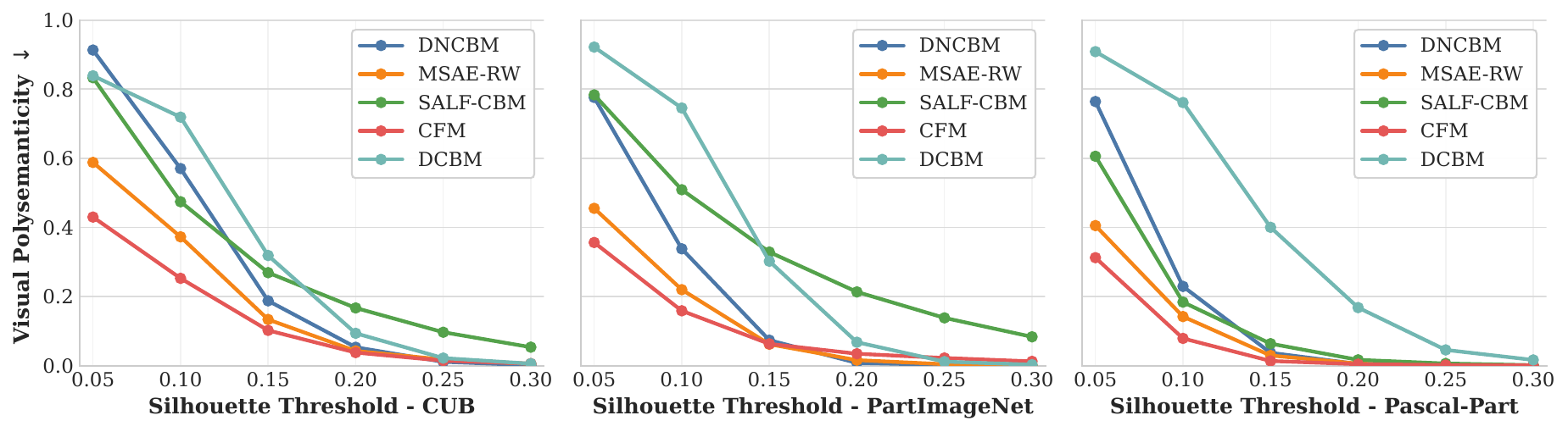} 
\caption{
\textbf{Visual Polysemanticity Score for unsupervised CBMs.}
Visual Polysemanticity scores across silhouette thresholds on CUB, PartImageNet, and Pascal-Part. Lower VPS values indicate stronger visual monosemanticity, CFM and MSAE consistently achieve the best scores, while the scores of non-SAE-based models indicate more polysemantic representations.
}
\label{fig:unsupervised_visual_mono_semantic} 
\end{figure}

\noindent\textit{\textbf{Visual Polysemanticity Score.}}
\label{subsubsec:evaluations:unsupervised:visual_monosemantic}
We observe that the DINOv2-based results in \Cref{fig:unsupervised_visual_mono_semantic} for the updated Visual Polysemanticity scores are consistent with the user studies reported in prior work on these models, which suggest that these methods indeed learn largely monosemantic concepts. The results also provide a clearer separation between methods. Across datasets, the Matryoshka SAE-based models, CFM and MSAE-RW, consistently exhibit the lowest fraction of visually polysemantic concept predictors, indicating that their learned concepts are concentrated around fewer visual modes. DNCBM, and other SAE-based models, also becomes less polysemantic at stricter silhouette thresholds, but shows higher polysemanticity at lower thresholds. In contrast, the non-SAE-based models retain a larger fraction of concept predictors with multiple visual modes, especially at moderate and high silhouette thresholds. \response{We additionally conducted a user study to compare VPS and CPE directly with human judgments of concept polysemanticity. Both metrics show statistically significant positive correlations with human judgments, with a numerically stronger association for VPS. The study design, participant agreement, and complete correlation results are reported in Appendix~\ref{appendix:vps-user-study}.}

\response{We also evaluate the sensitivity of VPS to the visual feature
extractor using OpenCLIP-L/14 (LAION-2B) and DINOv1-B/16 as alternatives
to DINOv2. With both alternative backbones, we observe the same trend of Maytroshka-based SAEs showing lower polysemanticity scores compared to other methods. The corresponding plots and discussion are
provided in Appendix~\ref{appendix:vps-backbone-sensitivity}.}

\begin{tcolorbox}[
  colback=green!6,
  colframe=green!50!black,
  boxrule=0.9pt,
  arc=2.5pt,
    before skip=2pt,
  after skip=2pt,
  left=4pt,right=4pt,top=4pt,bottom=4pt
]
\noindent\textbf{Takeaway 8.}
\emph{All models show visually monosemantic behavior at stricter silhouette thresholds, while Matryoshka SAE-based CBMs remain largely monosemantic even at lower thresholds where other models become strongly polysemantic.}
\end{tcolorbox}

\setlength{\abovecaptionskip}{4pt}

\subsection{Output Completeness}
\label{subsubsec:evaluations:unsupervised:outputcompeleteness} 
\begin{figure}[th] 
\centering 
\includegraphics[width=1.0\textwidth]{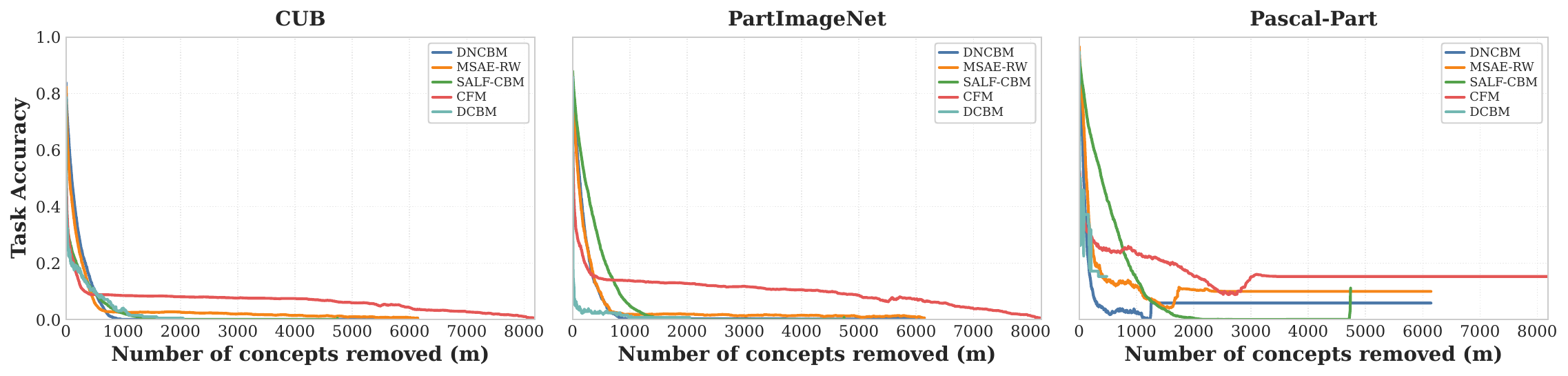} 
\caption{\textbf{Output completeness through utility-ordered concept deletion.} Utility-ordered deletion curves for on CUB, PartImageNet, and Pascal-Part. A sharper drop in task accuracy indicates that predictive evidence is concentrated in a compact subset of high-utility concepts. Most models degrade rapidly after deleting only a few concepts, with CFM showing a longer tail on CUB and PartImageNet.
}
\label{fig:unsupervised_output_completeness} 
\end{figure}

\Cref{fig:unsupervised_output_completeness} shows the utility-ordered deletion curves, where concepts are progressively removed according to their effect on the true-class logit. For most models and datasets, accuracy drops sharply after removing only a small number of high-utility concepts, indicating that task-relevant evidence is concentrated in a compact subset of the learned concepts. This behavior suggests high output completeness under the deletion protocol: the concepts most important for prediction capture a large fraction of the model's discriminative evidence.
The main exception is CFM, whose accuracy decreases more gradually on CUB and PartImageNet. This indicates that its predictive evidence is distributed across a larger set of active concepts, rather than being concentrated in only the top-ranked ones. On Pascal-Part, several models retain a non-zero accuracy after extensive concept deletion, which likely reflects residual prediction from non-concept terms such as the classifier bias, rather than evidence preserved in the removed concepts.

\setlength{\abovecaptionskip}{4pt}

\begin{tcolorbox}[
  colback=green!6,
  colframe=green!50!black,
  boxrule=0.9pt,
  arc=2.5pt,
    before skip=1pt,
  after skip=1pt,
  left=4pt,right=4pt,top=4pt,bottom=4pt
]
\noindent\textbf{Takeaway 9.}
\emph{Unsupervised CBMs are largely output-complete, with prediction driven by a compact set of high-utility concepts.}
\end{tcolorbox}

\subsection{Dictionary Disentanglement}
\label{subsubsec:evaluations:unsupervised:dictionary_disentanglement}

\begin{wrapfigure}{r}{0.5\textwidth}
\centering 
\captionsetup{font=footnotesize, skip=0.0pt}
\includegraphics[width=0.33\textwidth]{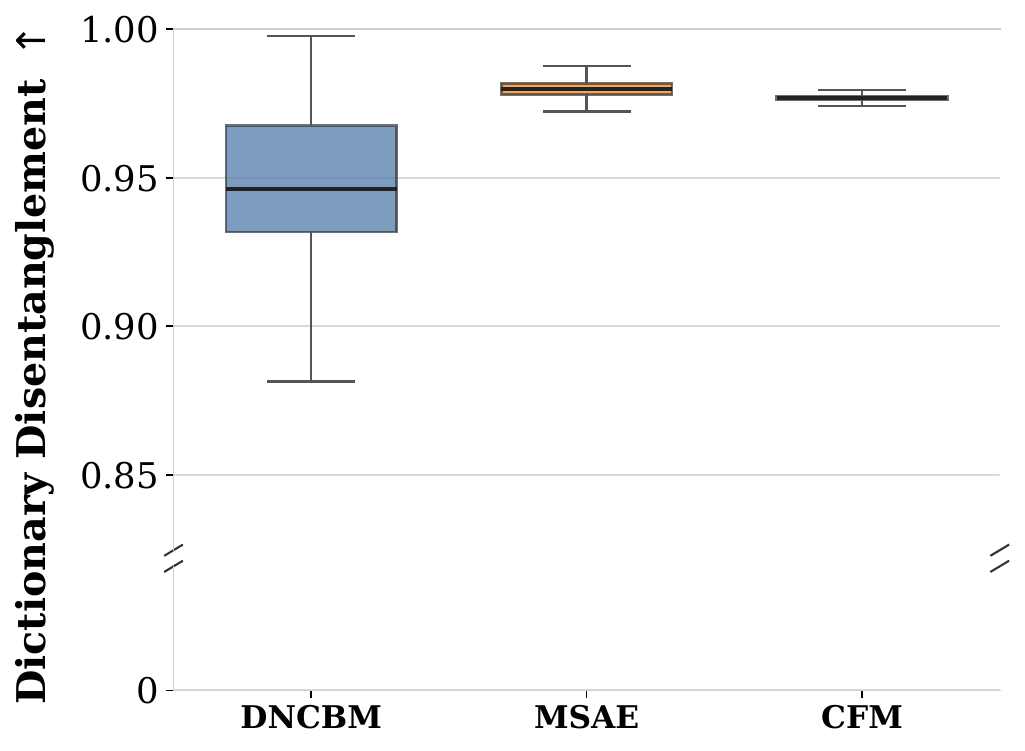} 
\caption{\textbf{Dictionary disentanglement in unsupervised CBMs.} Higher scores indicate more distinct decoder vectors. All SAE-based models achieve high disentanglement.}
\label{fig:unsupervised_dict_incoherence} 
\end{wrapfigure}

\Cref{fig:unsupervised_dict_incoherence} shows that all three SAE-based models achieve high dictionary disentanglement scores, indicating that their decoder dictionaries contain largely distinct columns rather than duplicate or highly overlapping directions. DNCBM already exhibits high disentanglement score, but its distribution is more variable than those of the Matryoshka SAE-based models and shows lower average value. In contrast, CFM and MSAE achieve scores closer to one with tighter distributions, suggesting lower geometric redundancy among their decoder columns. This trend is consistent with the Visual Polysemanticity results, where models with better disentangled dictionary columns also tend to exhibit better visual monosemanticity (lower VPS). However, this metric should be interpreted as a decoder-level geometric diagnostic, not as a standalone measure of semantic interpretability.

\begin{tcolorbox}[
  colback=green!6,
  colframe=green!50!black,
  boxrule=0.9pt,
  arc=2.5pt,
    before skip=4pt,
  after skip=4pt,
  left=4pt,right=4pt,top=4pt,bottom=4pt
]
\noindent\textbf{Takeaway 10.}
\emph{SAE-based unsupervised CBMs learn highly non-redundant decoder dictionaries, with Matryoshka SAE variants showing the strongest separation among decoder columns.}
\end{tcolorbox}

\section{Limitations}
\label{sec:limitations}

With \cbmbench{}, we introduce a suite of metrics that target key properties underlying trustworthy explanations in CBMs. By applying these metrics to representative models, our analysis quantifies several failure modes and identifies findings that can inform future CBM research. At the same time, we highlight the challenges and limitations in these evaluation metrics below.

For supervised CBMs, \cbmbench{} is limited by the availability of datasets that jointly provide task labels, concepts, and spatial localization annotations.
We require these annotations jointly to assess different aspects of the same model on the same dataset, providing a more holistic view of its behavior and of how different properties relate to each other. For example, grounding metrics such as EPG require localization annotations, while leakage and purity metrics require concept annotations. Evaluating these on the same dataset lets us analyze how different model behaviors relate to each other. However, existing annotations can also be imperfect, with concept vocabularies that may be incomplete and labels that may be noisy or coarse. For example, CUB-200 provides class-level concept vectors even when a concept is occluded or absent in a specific image, which can encourage memorization-like behavior, while PartImageNet provides 40 coarse concepts for 158 classes.

Overall, the metrics in \cbmbench{} should be interpreted as diagnostic measures rather than exhaustive tests of interpretability. Several scores depend on the choice of attribution maps (EPG and semantic coherence), masking (output completeness), or specific feature representations (semantic coherence and VPS), which can influence the benchmark metrics. 
Yet, while not an exhaustive measure of interpretability it enables identifying failure modes and shortcomings of existing methods.

Within these constraints, \cbmbench{} evaluates a representative set of supervised and unsupervised CBMs covering different concept-learning settings. The set of supervised models includes deterministic, stochastic, and embedding-based approaches, while the set of unsupervised models includes different concept discovery approaches. Extending \cbmbench{} to additional datasets, model classes, multi-label settings, and broader studies of how users interpret and intervene on learned concepts remains an important direction for future work.

\section{Conclusion}
We introduced \cbmbench{}, a unified set of metrics for evaluating concept bottleneck models beyond concept and task accuracy. By applying these metrics to representative supervised CBMs, we show that accuracy alone can hide important failure modes: high concept accuracy does not necessarily imply strong task performance, good spatial grounding, low leakage, or low intervention risk. This is especially clear on PartImageNet, where limited concept coverage leads to a large gap between concept prediction and downstream task accuracy, making the dataset a useful stress test for CBMs under incomplete concept vocabularies.

Our supervised evaluation reveals consistent differences between hard and soft bottlenecks. Hard CBMs generally achieve higher purity, lower leakage, sharper output-completeness behavior, and lower intervention risk, suggesting that their concept representations are more constrained. However, these benefits often come with reduced task performance, especially when the concept set is insufficient. Soft CBMs tend to achieve stronger task accuracy and smaller gaps to black-box baselines, but often distribute task-relevant evidence across more entangled concept representations, resulting in higher leakage and intervention risk. These results show that interpretability and intervenability require more than accurate concept prediction.

For unsupervised CBMs, \cbmbench{} provides complementary diagnostics for evaluating the semantic consistency, monosemanticity, dictionary disentanglement, and output completeness of discovered concepts. Our results show that semantic coherence remains challenging, especially on datasets with spatially small concepts, although Matryoshka SAE-based models generally perform better. We also find that CPE can disagree with qualitative evidence and prior user studies, while our Visual Polysemanticity Score better reflects the observed monosemantic behavior of SAE-based models. In addition, dictionary disentanglement suggests that SAE-based models learn largely non-redundant decoder dictionaries, and output-completeness results show that predictive evidence is often concentrated in a compact subset of high-utility concepts.

\response{\cbmbench{} does not provide a single ranking of the models. Instead, it helps practitioners compare models based on the properties that matter for their application. For example, grounding, leakage, and purity are relevant when a model will be used for concept interventions, while semantic coherence and polysemanticity are relevant when evaluating discovered concepts. The results can also show which aspects of a model require additional improvements.}

Overall, \cbmbench{} highlights that CBM evaluation requires a multi-faceted view of model behavior. Concept accuracy and task accuracy are important, but they do not fully capture grounding, leakage, purity, output completeness, or the quality of discovered concepts. By bringing these dimensions into a unified evaluation framework, \cbmbench{} provides a practical suite of metrics for analyzing CBM behavior and for guiding future work toward models whose concepts are not only predictive, but also better grounded, more specialized, and more useful for interpretation and intervention.

\section*{\response{Acknowledgements}}
\response{Funded in part by the DFG (German Research Foundation, RTG 2853/1).}

\bibliography{main_cleaned_verified}
\bibliographystyle{tmlr}

\newpage

\appendix

\begin{center}
    {\LARGE \bfseries \cbmbench{}: Hitchhiker's Guide to Evaluating Concept Bottleneck Models \par}
    \vspace{1em}
    {\Large \bfseries Supplementary Material \par}
\end{center}
\section{Appendix}
\etocsetnexttocdepth{subsubsection}
\localtableofcontents

\subsection{\cbmbench{} Metrics}
\label{appendix:metrics}

\subsubsection{Supervised Metrics}
\label{appendix:metrics-supervised}

\paragraph{Spatial Localization: SUB.}
\label{appendix:metrics-sub}
Following \citep{sub}, SUB evaluates the \emph{faithfulness of concept predictions} under synthetic, fine-grained attribute edits. Each sample consists of an \emph{edited} image in which one specific concept $k^-$ has been substituted by another concept $k^+$, yielding two accuracy-like metrics:
\begin{align}
\mathcal{S}^+ &=
\frac{1}{|\mathcal{D}_{\text{sub}}|}
\sum_{(x,k^+)\in \mathcal{D}_{\text{sub}}}
\mathbb{I}\!\left[\hat{c}_{k^+}(x)=1\right],
&&
\text{(detect the substituted-in attribute)}
\\
\mathcal{S}^- &=
\frac{1}{|\mathcal{D}_{\text{sub}}|}
\sum_{(x,k^-)\in \mathcal{D}_{\text{sub}}}
\mathbb{I}\!\left[\hat{c}_{k^-}(x)=0\right],
&&
\text{(do \emph{not} predict the removed attribute),}
\end{align}
where $\hat{c}_k(x)$ denotes the model's predicted presence of concept $k$ on input $x$.

\paragraph{Semantic coherence.}
\label{appendix:metrics-semantic-coherence}
Semantic coherence measures whether the visual evidence attributed to a concept is consistent across images containing the same ground-truth concept. The metric is defined in visual feature space and complements EPG, where instead of measuring overlap with an annotated mask, it evaluates whether a concept consistently highlights the same type of evidence across examples.

The goal is to leverage the strong semantic features of foundation models to measure the similarity of the regions highlighted by an attribution method for a given concept. 

To this end, we use DINOv2~\citep{dinov2}. For each input image $x$ and concept $k$, we compute an attribution map $A_k(x)\in\mathbb{R}^{H\times W}$ with respect to that concept, and simultaneously extract the DINOv2 features of the same image. Since DINOv2 features have lower spatial resolution than the input image, we upsample them using AnyUp~\citep{anyup} to obtain a high-resolution DINO feature map $F_{\mathrm{HR}}(x)\in\mathbb{R}^{D\times H\times W}$. DINOv2 provides a strong generic visual representation, while AnyUp aligns these features with the attribution map at image resolution. We then perform attribution-weighted feature pooling after normalizing the attribution map into a probability distribution by dividing by the total positive attribution:

\begin{equation}
p_k(x)_{ij} \;=\; \frac{\max(A_k(x)_{ij},\,0)}{\sum_{u=1}^{H}\sum_{v=1}^{W}\max(A_k(x)_{uv},\,0)\;+\;\epsilon},
\label{eq:semcoh_pk_bench}
\end{equation}
with $\epsilon>0$ for numerical stability. We compute a concept-specific ROI embedding by attribution-weighted pooling of DINO features followed by $\ell_2$ normalization:
\begin{align}
v_k(x) \;&=\; \sum_{i=1}^{H}\sum_{j=1}^{W} p_k(x)_{ij}\;F_{\mathrm{HR}}(x)_{:,i,j}\in\mathbb{R}^{D},
\label{eq:semcoh_vk_bench}
\\
u_k(x) \;&=\; \frac{v_k(x)}{\|v_k(x)\|_2+\epsilon}\in\mathbb{R}^{D}.
\label{eq:semcoh_uk_bench}
\end{align}
Let $\mathcal{I}_k$ be the set of images in which the ground-truth concept $k$ is present and produces a valid evidence embedding. We compute semantic coherence through average pairwise cosine similarity:
\begin{equation}
S(c_k)\;=\;\frac{1}{\binom{|\mathcal{I}_k|}{2}}\sum_{\substack{x_i,x_j\in\mathcal{I}_k\\ i<j}}
u_k(x_i)^\top u_k(x_j),
\label{eq:semcoh_pairwise_bench}
\end{equation}
We report the resulting per-concept scores, and summarize them at the model level by averaging over the concepts. Higher values indicate that the concept repeatedly highlights the same type of visual evidence.

\paragraph{Purity.}
\label{appendix:metrics-purity}
To assess how each concept is aligned with the concept predictors, which we call purity, we use variation of information which measures the distance between two random variables. What we measure is the distance between the ground truth concept and the predicted ones which is then converted into a similarity metric that gives us how each ground truth is aligned w.r.t the concept predictions. We evaluate how uniquely each ground-truth concept is captured by a predicted concept. Let $C^*\in\{0,1\}^{N\times K}$ denote the annotated ground-truth concepts and $C\in\mathbb{R}^{N\times K}$ the model’s predicted concept set on the same data.

For each ground-truth concept $C^*_i$ and predicted concept $\tilde{C}_j$, we compute the variation of information (VI) from dataset marginal frequencies. Specifically, we estimate $\hat{H}(C^*_i\mid \tilde{C}_j)$ and $\hat{H}(\tilde{C}_j\mid C^*_i)$, and define
\begin{equation}
\widehat{\mathrm{VI}}(C^*_i,\tilde{C}_j) \;=\; \hat{H}(C^*_i\mid \tilde{C}_j) + \hat{H}(\tilde{C}_j\mid C^*_i),
\label{eq:vi_disent_bench}
\end{equation}
We then convert VI into an interpretable agreement score in $(0,1]$:
\begin{equation}
A_{i,j} \;=\; \frac{1}{1+\widehat{\mathrm{VI}}(C^*_i,\tilde{C}_j)} \;\;\in\;(0,1],
\label{eq:vi_agreement_bench}
\end{equation}
which measures agreement between ground-truth and predicted concept distributions.

Finally, for each ground-truth concept $i$, let $A_i^{(1)}$ and $A_i^{(2)}$ denote the highest and second-highest agreement scores. We define the purity score as
\begin{equation}
P \;=\; \frac{1}{K}\sum_{i=1}^{K}\left(A_i^{(1)} - A_i^{(2)}\right).
\label{eq:purity_bench}
\end{equation}
Larger values indicate more unique correspondence between ground-truth and predicted concepts.

\paragraph{Output Completeness. }
\label{appendix:metrics-output_completeness}

In output Completeness we measure how the task performance of the model get affected by concept insertion and deletion, respectively. A model is more output complete if inserting/deleting a small number of high-utility concepts quickly recovers/deteriorates full model behavior. The order in which we select which concept to delete and which to insert is determined by a computed utility for each concept.

For an input $x$ with ground-truth label $y$, let $s(x)$ denote the task score defined as the \emph{true-class logit} (the logit corresponding to $y$). For concept $k$, define its utility as the expected absolute change in the task score when concept $k$ is masked:
\begin{equation}
u(k)
\;=\;
\mathbb{E}_{(x,y)\sim \mathcal{D}}
\Big[
\big| s(x) - s^{(-k)}(x) \big|
\Big],
\label{eq:utility_bench}
\end{equation}
where $s^{(-k)}(x)$ is the true-class logit after masking concept $k$ in the representation fed to the task head. We consider a \emph{class-conditional} ordering $\pi_c$ obtained by sorting utilities computed only on samples with ground-truth class $c$.

Let $K$ denote the number of concepts and $m\in\{0,\dots,K\}$ the number of perturbed concepts. Using a fixed masking baseline of $0$ and denoting the concept representation by $\mathbf{h}(x)$, we define
\begin{align*}
\textbf{Deletion:}\quad
\mathbf{h}^{\mathrm{del}}_{m}(x)
&=
\mathbf{h}(x)\odot \mathbf{1}_{\{\pi_{m+1},\dots,\pi_{K}\}},
\\
\textbf{Insertion:}\quad
\mathbf{h}^{\mathrm{ins}}_{m}(x)
&=
\mathbf{h}(x)\odot \mathbf{1}_{\{\pi_{1},\dots,\pi_{m}\}}.
\end{align*}
Task performance is then measured as
$P_{\mathrm{del}}(m)=\mathrm{Perf}(f(\mathbf{h}^{\mathrm{del}}_{m}(x)))$
and
$P_{\mathrm{ins}}(m)=\mathrm{Perf}(f(\mathbf{h}^{\mathrm{ins}}_{m}(x)))$.
We summarize the insertion and deletion curves using AUC over the evaluated grid.

\begin{figure}[!htbp]
    \centering
    \captionsetup{font=footnotesize, skip=0.0pt}
    \includegraphics[width=1.0\textwidth]{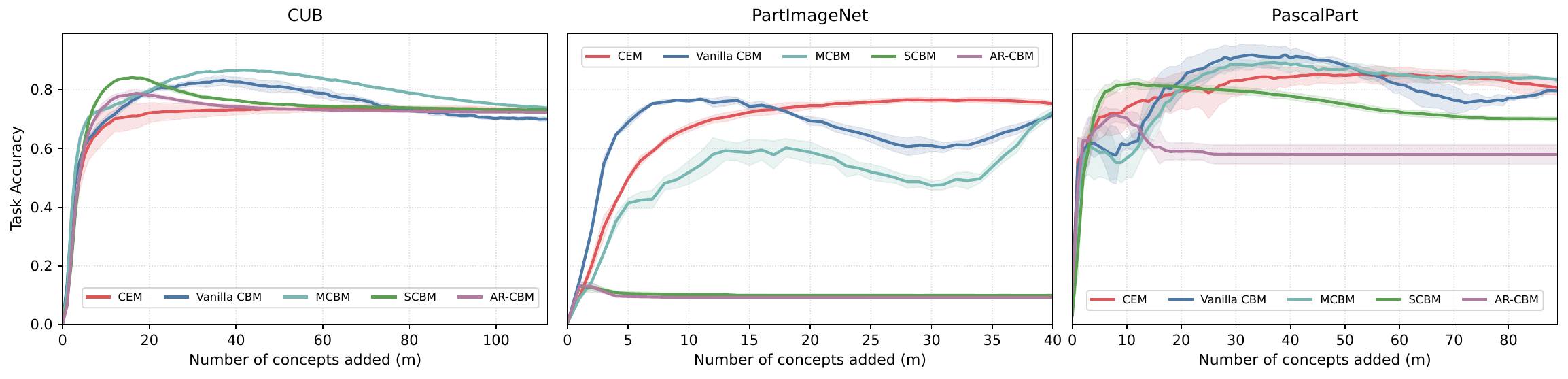}
    \caption{Utility-ordered deletion insertion curves show how task accuracy changes as concepts
are added on the three datasets.}
    \label{app-fig:output_completeness_insertion_curves}
\end{figure}

\begin{table}[H]
\centering
\caption{Supervised output-completeness AUC metrics. Values are mean (standard deviation) over seeds.}
\small
\resizebox{0.72\textwidth}{!}{\begin{tabular}{l l c c}
\toprule
\textbf{Dataset} & \textbf{Model} & \textbf{AUC$_{del}$}$\downarrow$ & \textbf{AUC$_{ins}$}$\uparrow$ \\
\midrule
  \multirow{6}{*}{CUB} & CBM & $0.097\,(0.001)$ & $0.742\,(0.012)$ \\
   & CEM & $0.465\,(0.031)$ & $0.708\,(0.022)$ \\
   & MCBM & $0.114\,(0.004)$ & $0.785\,(0.002)$ \\
   & ARCBM & $0.088\,(0.004)$ & $0.719\,(0.003)$ \\
   & SCBM-G & $0.074\,(0.001)$ & $0.726\,(0.012)$ \\
   & SCBM-A & $0.071\,(0.002)$ & $0.738\,(0.003)$ \\
\midrule
  \multirow{6}{*}{PartImageNet} & CBM & $0.062\,(0.004)$ & $0.652\,(0.008)$ \\
   & CEM & $0.502\,(0.014)$ & $0.668\,(0.009)$ \\
   & MCBM & $0.089\,(0.005)$ & $0.501\,(0.028)$ \\
   & ARCBM & $0.010\,(0.000)$ & $0.095\,(0.003)$ \\
   & SCBM-G & $0.008\,(0.000)$ & $0.108\,(0.002)$ \\
   & SCBM-A & $0.010\,(0.000)$ & $0.101\,(0.005)$ \\
\midrule
  \multirow{6}{*}{PascalPart} & CBM & $0.111\,(0.014)$ & $0.799\,(0.025)$ \\
   & CEM & $0.417\,(0.062)$ & $0.807\,(0.047)$ \\
   & MCBM & $0.143\,(0.009)$ & $0.803\,(0.029)$ \\
   & ARCBM & $0.127\,(0.013)$ & $0.591\,(0.037)$ \\
   & SCBM-G & $0.112\,(0.038)$ & $0.527\,(0.021)$ \\
   & SCBM-A & $0.191\,(0.010)$ & $0.742\,(0.011)$ \\
\bottomrule
\end{tabular}}

\label{tab:supervised-auc-class-conditional}
\end{table}

\subsubsection{Supervised Metrics Correlation Heatmap}
\label{appendix:metrics-supervised-correlation}
\begin{figure}[!htbp]
    \centering
    \captionsetup{font=footnotesize, skip=0.0pt}
    \rotatebox{270}{
    \includegraphics[width=1.33\textwidth]{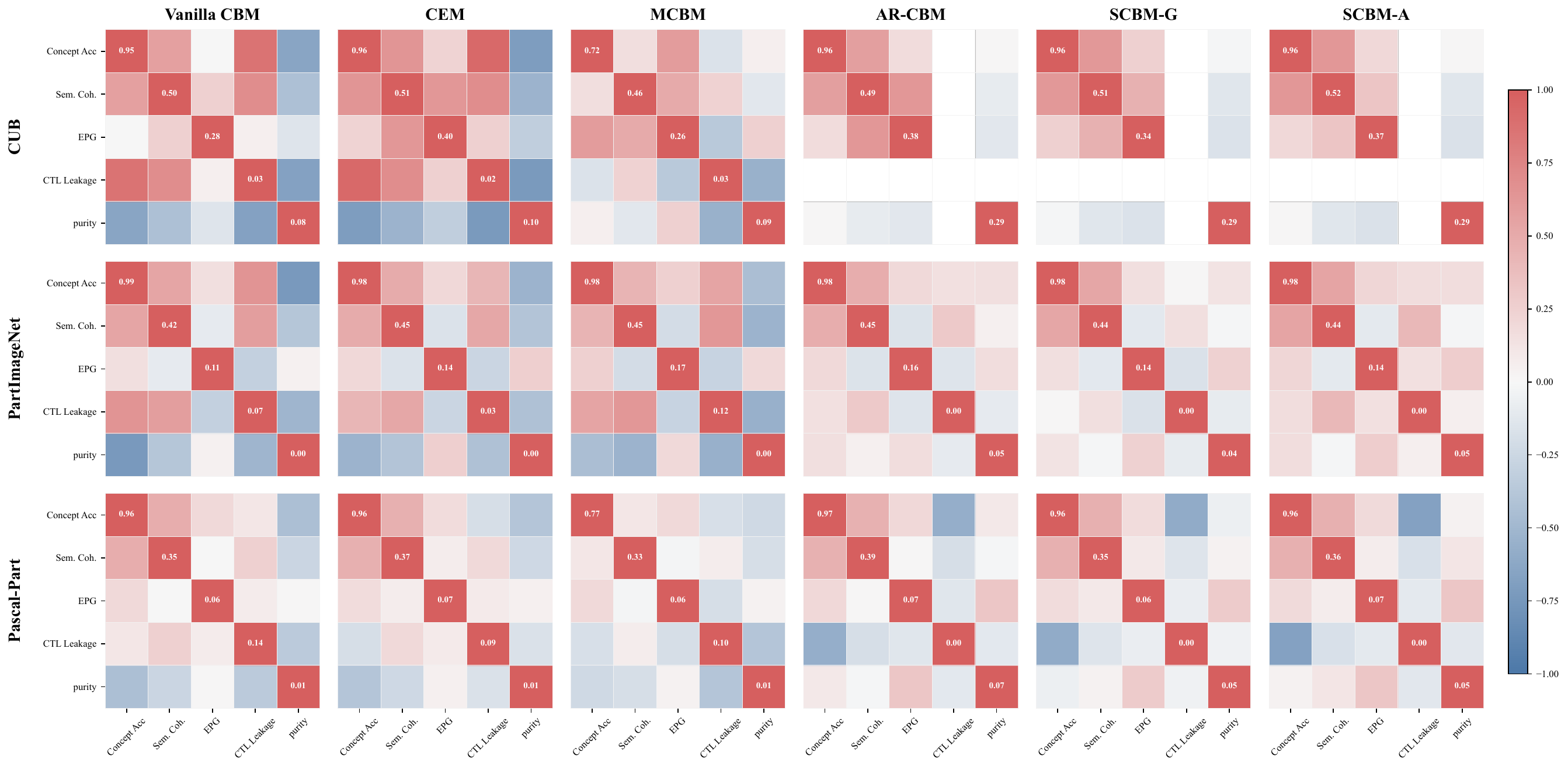}}
    \caption{Supervised metrics correlation heatmap.}

    \label{app-fig:supervised-metrics-correlation-heatmap}
\end{figure}

In \Cref{app-fig:supervised-metrics-correlation-heatmap}, we report the correlation score between every pair of supervised metrics across all the representative models and datasets.

\subsubsection{Unsupervised Metrics}
\label{appendix:metrics-unsupervised}

\begin{table}[!htbp]
\centering
\caption{\textbf{Statistics of the active concepts in the SAE-based unsupervised models.} Showing the number of active unique concepts, and the average number of active concepts per image as Act./img.}
\label{appendix-tab:unsupervised_active_concepts}
\small
\setlength{\tabcolsep}{5.5pt}
\renewcommand{\arraystretch}{1.12}
\begin{tabular}{@{}lrrrrrrr@{}}
\toprule
\multirow{2}{*}{Model}
& \multirow{2}{*}{Total}
& \multicolumn{2}{c}{CUB}
& \multicolumn{2}{c}{PartImageNet}
& \multicolumn{2}{c}{Pascal-Part} \\
\cmidrule(lr){3-4}
\cmidrule(lr){5-6}
\cmidrule(lr){7-8}
& & Unique & Act./img. & Unique & Act./img. & Unique & Act./img. \\
\midrule
DNCBM   & 6,144 & 1,257 & 615 & 1,260 & 616 & 1,260 & 615 \\
CFM     & 8,192 & 6,545 & 95  & 6,422 & 85  & 6,233 & 93  \\
MSAE-RW & 6,144 & 2,765 & 256 & 3,722 & 256 & 4,281 & 256 \\
\bottomrule
\end{tabular}
\end{table}

\paragraph{Semantic Coherence.}
\label{appendix:metrics-unsupervised-semantic-coherence}
For unsupervised CBMs, we use a label-free variant of semantic coherence that does not require any ground-truth concept annotations or any mapping from model concepts to dataset concepts. Each learned concept is treated as a native unit of analysis. For a given image $x$, we first compute the model’s concept scores. For the SAE-based models, this corresponds to the active concepts in that image, and for dense models, it corresponds to the all concepts.

Following the same approach for the supervised models, for every selected concept $c$, we compute a concept-specific attribution map $A_c(x)\in\mathbb{R}^{H\times W}$ and a high-resolution DINOv2 feature map $F_{\mathrm{HR}}(x)\in\mathbb{R}^{D\times H\times W}$. The positive attribution mass is normalized into spatial weights
\begin{equation}
p_c(x)_{ij}
\;=\;
\frac{\max(A_c(x)_{ij},0)}{\sum_{u=1}^{H}\sum_{v=1}^{W}\max(A_c(x)_{uv},0)+\epsilon},
\label{eq:lf_semcoh_weights}
\end{equation}
which are used to pool the visual features inside the attributed region. This yields one normalized evidence embedding per concept-image pair,
\begin{align}
v_c(x)
\;&=\;
\sum_{i=1}^{H}\sum_{j=1}^{W} p_c(x)_{ij} F_{\mathrm{HR}}(x)_{:,i,j},
\label{eq:lf_semcoh_embedding}
\\
u_c(x)
\;&=\;
\frac{v_c(x)}{\|v_c(x)\|_2+\epsilon}.
\label{eq:lf_semcoh_norm_embedding}
\end{align}
For each concept $c$, let $\mathcal{E}_c=\{\bar{e}_c(x)\}$ denote the set of valid evidence embeddings collected across images. Otherwise, its label-free semantic coherence is
\begin{equation}
S_{\mathrm{LF}}(c)
\;=\;
\frac{1}{\binom{|\mathcal{E}_c|}{2}}
\sum_{\substack{u_i,u_j\in \mathcal{E}_c\\ i<j}}
u_i^\top u_j.
\label{eq:lf_semcoh_score}
\end{equation}
Higher scores indicate that the learned concept repeatedly localizes visually similar evidence across images, without using concept labels.

\paragraph{Concept Polysemantic Entropy (CPE) score.}
\label{appendix:metrics-unsupervised-cpe}

Concept Polysemantic Entropy (CPE) score, introduced in~\citep{chain_of_explanation}, measures the semantic diversity of the visual evidence associated with a learned concept. It does so by using a VLM to identify named visual concepts in the evidence and computing the entropy of their distribution.

For each concept $c$, we rank images by their concept score $f_c(x)$ and retain the top-$K$ evidence examples. For each selected image, we compute the attribution map for $c$, threshold the positive attribution mass to obtain a binary evidence mask, and save the corresponding masked evidence image.

Each masked evidence image is then described by a vision-language model with short object or attribute phrases, these phrases are merged into semantic columns, producing an empirical count vector $n_{c,a}$ over columns $a$. If a concept has too few distinct columns, we use the padded naive entropy variant from \citep{chain_of_explanation} and add singleton padding columns before normalization. Let $\tilde{\mathcal{A}}_c$ denote the resulting padded column set and let
\begin{equation}
\tilde{q}_{c,a}
\;=\;
\frac{n_{c,a}}{\sum_{b\in \tilde{\mathcal{A}}_c} n_{c,b}}
\label{eq:cpe_distribution_bench}
\end{equation}
be the padded empirical distribution. We compute per-concept CPE as the normalized entropy
\begin{equation}
\mathrm{CPE}(c)
\;=\;
-\frac{1}{\log |\tilde{\mathcal{A}}_c|}
\sum_{a\in \tilde{\mathcal{A}}_c}
\tilde{q}_{c,a}\log \tilde{q}_{c,a},
\label{eq:cpe_bench}
\end{equation}
and report the model-level score as the mean of $\mathrm{CPE}(c)$ over scored concepts. Higher values indicate that a concept is associated with a broader and more diverse set of semantic descriptions.

\paragraph{Visual Polysemanticity Score (VPS).}
\label{appendix:metrics-unsupervised-vps}

Visual Polysemanticity Score (VPS) measures whether the visual evidence associated with a learned concept forms a single coherent visual mode or separates into multiple visually distinct modes. Unlike CPE, which estimates polysemanticity from language descriptions of evidence regions, VPS operates directly in a visual feature space.

For each concept $c$, we rank original images by their raw concept score $f_c(x)$ and retain the top-$K$ images. For sparse SAE-based models, only images with positive concept activations are considered in this ranking. For each selected image $x_{c,i}$, we compute an attribution map $A_c(x_{c,i})$ with respect to concept $c$. We also extract a dense DINOv2 feature map from the original image and upsample it using AnyUp. The positive attribution values are normalized into spatial pooling weights, and the concept evidence embedding is computed as
\begin{equation}
z_{c,i}
\;=\;
\frac{
\sum_{p} \alpha_{c,i,p}\,\Phi(x_{c,i})_p
}{
\left\|
\sum_{p} \alpha_{c,i,p}\,\Phi(x_{c,i})_p
\right\|_2
},
\qquad
\alpha_{c,i,p}
=
\frac{\max(A_c(x_{c,i})_p,0)}
{\sum_{q}\max(A_c(x_{c,i})_q,0)} ,
\label{eq:vps_attribution_pooled_embedding}
\end{equation}
where $\Phi(x_{c,i})_p$ is the upsampled DINOv2 feature at spatial position $p$. Images whose positive attribution mass is zero are discarded.

We then cluster the embeddings $\{z_{c,i}\}_{i=1}^{N_c}$ for each concept using $k$-means over candidate cluster counts $m\in\{2,\ldots,m_{\max}\}$. Clusterings are considered valid only if every cluster contains at least a minimum number of embeddings. For each valid clustering, we compute the silhouette score and retain the best value
\begin{equation}
s_c^\star
\;=\;
\max_{m\in\mathcal{M}}
\mathrm{Silhouette}
\bigl(\{z_{c,i}\}_{i=1}^{N_c}, m\bigr),
\label{eq:vps_best_silhouette}
\end{equation}
where $\mathcal{M}$ denotes the set of cluster counts for concept.

A concept is labeled visually polysemantic if $s_c^\star \geq \tau$, where $\tau$ is a fixed silhouette threshold, and monosemantic otherwise. Concepts with too few valid embeddings or no valid clustering are labeled uncertain. We report the model-level VPS as the fraction of concepts labeled visually polysemantic, together with the monosemantic and uncertain fractions. Higher VPS indicates that learned concepts more often decompose into multiple visually separable evidence modes.

\paragraph{Dictionary Disentanglement.}
\label{appendix:metrics-unsupervised-dictionary-incoherence}

Dictionary Disentanglement is a geometry-based diagnostic for SAE-style unsupervised CBMs. It measures whether the decoder dictionary columns learned by the model are redundant with one another. The metric is motivated by dictionary coherence in sparse coding and dictionary learning, where highly aligned dictionary columns indicate duplicate or overlapping representation directions.

For an SAE-based model with decoder dictionary vectors $d_1,\dots,d_K$, each vector corresponds to one learned decoder column, or concept direction. We compute the metric over the full decoder dictionary, rather than restricting to concepts that are active on a particular evaluation split. Each column is first normalized,
\begin{equation}
\hat{d}_i
\;=\;
\frac{d_i}{\|d_i\|_2},
\qquad i=1,\ldots,K.
\label{eq:dictionary_disentanglement_normalized_columns}
\end{equation}
We then compute the positive cosine similarity between every pair of distinct decoder columns. As in dictionary coherence diagnostics, we keep only the positive part of the cosine similarity: positively aligned columns indicate redundant directions, whereas negatively aligned columns encode opposite directions and are not treated as duplicates.

For each decoder column $d_i$, we define its per-column disentanglement as one minus its average positive cosine similarity to all other columns,
\begin{equation}
\mathrm{DI}(i)
\;=\;
1
-
\frac{1}{K-1}
\sum_{j\neq i}
\max\left(0,\hat{d}_i^\top \hat{d}_j\right).
\label{app-eq:dictionary_disentanglement_per_column}
\end{equation}
The final Dictionary Disentanglement score is the mean of $\mathrm{DI}(i)$ over all decoder columns. Higher values indicate that, on average, dictionary columns are well separated and not positively aligned with the rest of the dictionary. Lower values indicate that many columns share similar positive directions, suggesting duplicate or overlapping dictionary elements.

Dictionary Disentanglement should be interpreted as a measure of geometric redundancy, not as a direct measure of semantic interpretability. A high score means that decoder directions are well separated in feature space, but it does not guarantee that the corresponding concepts are semantically meaningful, spatially grounded, or useful for downstream prediction. We therefore interpret Dictionary Disentanglement alongside semantic coherence, CPE, EPG, visual polysemanticity, and task performance.

\paragraph{Output Completeness without Labels.}
\label{appendix:metrics-unsupervised-output-completeness}
For unsupervised CBMs, we use the same insertion and deletion protocol as in Appendix~\ref{appendix:metrics-output_completeness}, but treat the model’s active concepts as the concept set for the SAE-based CBMs and all the concepts for the SALF-CBM. The utility of a concept is computed directly from the change in the true-class logit when that native concept is masked, so no concept labels or concept mappings are required. Class-conditional utility rankings, deletion curves, insertion curves, and the AUC summaries are then computed exactly as in the supervised setting.

\subsubsection{Spatial Localization: Energy Pointing Game (EPG) using different attribution methods. }
\label{appendix:metrics-epg}
The results regarding spatial localization of the main paper indicate that high concept accuracy does not imply strong spatial grounding (\Cref{subsubsec:evaluations:supervised:spatial_grounding}). Here we verify the same conclusion under two additional attribution methods.

\begin{figure}[!htbp]
    \centering
    \captionsetup{font=footnotesize, skip=0.0pt}
    \includegraphics[width=1.0\textwidth]{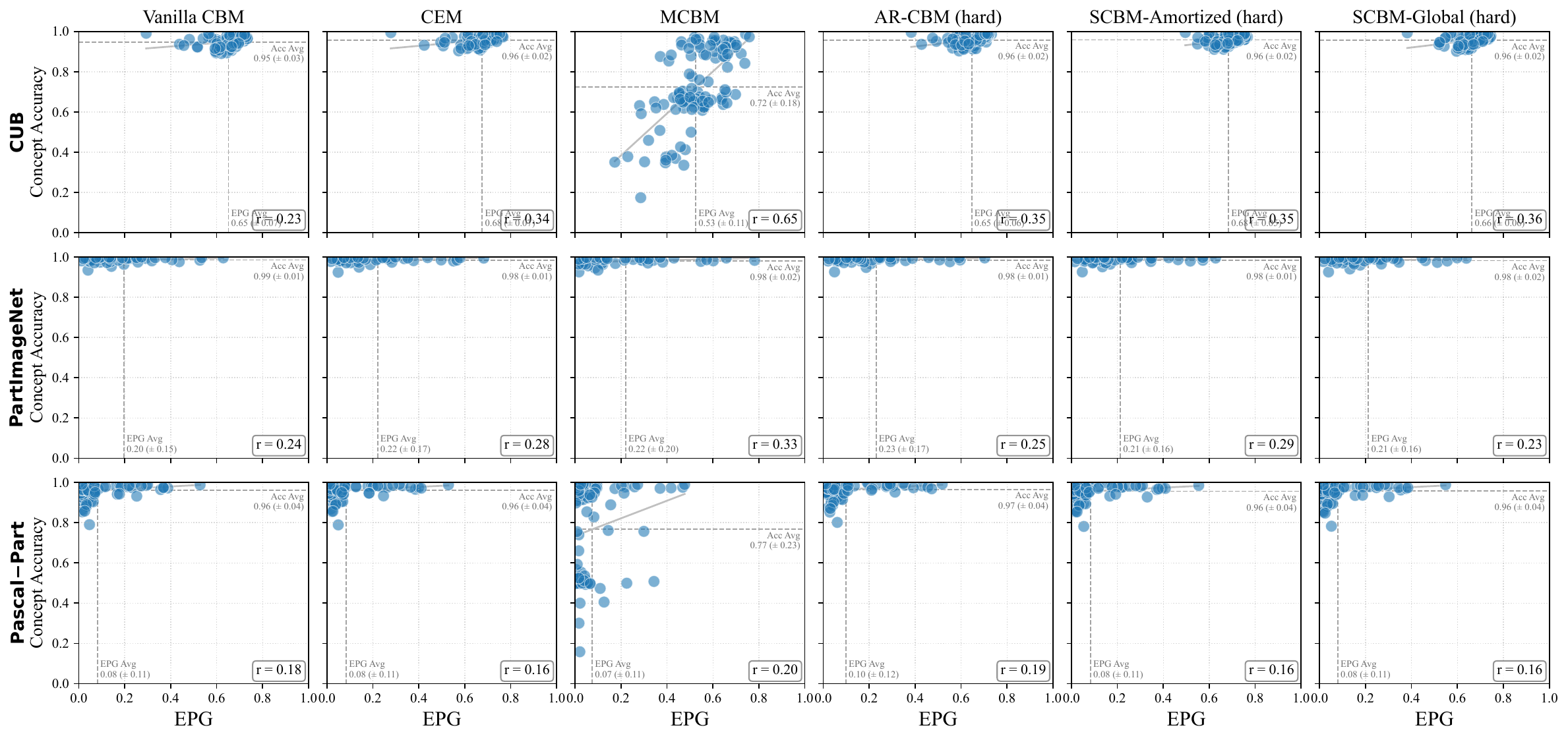}
    \caption{Concept accuracy vs. Energy Pointing Game score using Guided-GradCAM. \textbf{High concept accuracy does not guarantee spatial localization.}}
    \label{app-fig:epg-guided-gradcam}
\end{figure}

\noindent\textit{Guided-GradCAM.} \Cref{app-fig:epg-guided-gradcam} shows the same qualitative pattern as in the main text: concepts can have high accuracy while EPG remains modest, especially for datasets with smaller part masks.

\begin{figure}[!htbp]
    \centering
    \captionsetup{font=footnotesize, skip=0.0pt}
    \includegraphics[width=1.0\textwidth]{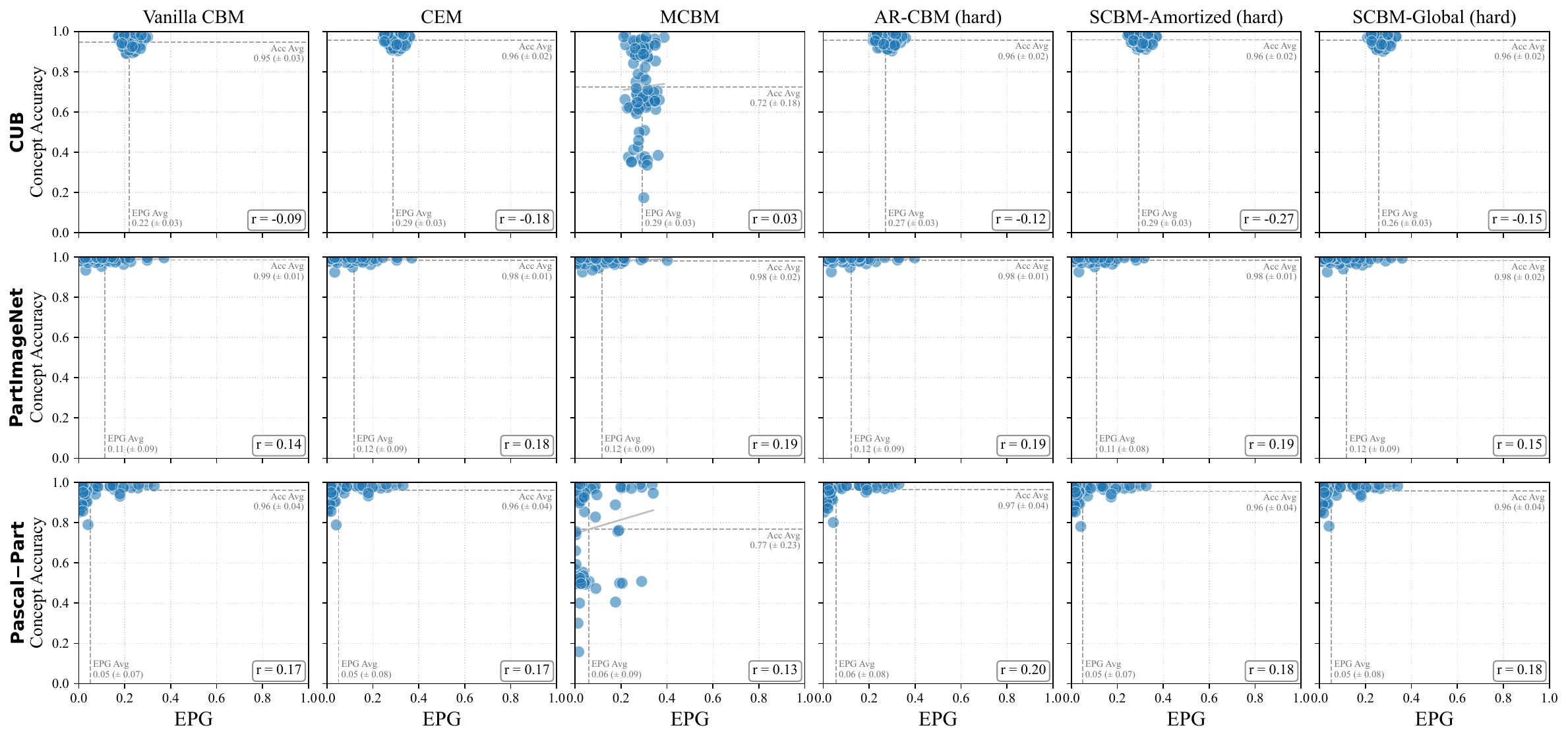}
    \caption{Concept accuracy vs. Energy Pointing Game score using Input$\times$Gradient. \textbf{High concept accuracy does not guarantee spatial localization.}}
    \label{app-fig:epg-inputxgrad}
\end{figure}

\noindent\textit{Input$\times$Gradient.} \Cref{app-fig:epg-inputxgrad} is consistent with Guided-GradCAM: the localization gap persists, indicating that weak EPG is not an artifact of a single attribution method.

\response{\Cref{tab:epg-attribution-rank-correlation} summarizes the agreement between the model rankings induced by the three attribution methods.}

\begin{table}[!htbp]
\centering
\caption{\response{Spearman correlations between the rankings of six supervised CBMs by EPG under different attribution methods. Model-level EPG scores are averaged over concepts and three seeds. The mean is the arithmetic mean of the three pairwise correlations.}}
\label{tab:epg-attribution-rank-correlation}
\response{\begin{tabular}{lrrrr}
\toprule
\textbf{Dataset} & \textbf{Grad.--Guided} & \textbf{Grad.--Input$\times$Grad.} & \textbf{Guided--Input$\times$Grad.} & \textbf{Mean} \\
\midrule
CUB          & 0.54 & $-0.03$ & $-0.03$ & 0.16 \\
PartImageNet & 0.77 & 0.54    & 0.54    & 0.62 \\
Pascal-Part  & 0.94 & $-0.37$ & $-0.26$ & 0.10 \\
\bottomrule
\end{tabular}}
\end{table}

\subsubsection{Task Accuracy of the Unsupervised Models.} 
\label{appendix:metrics-unsupervised_model_accurcay}

For completeness, we report test task accuracy for the unsupervised models across all three datasets in \Cref{tab:unsup_test_acc}.

\begin{table}[th]
\centering
\caption{Test task accuracy for unsupervised models}
\begin{tabular}{lccc}
\hline
Model & CUB & PartImageNet & PascalPart \\
\hline
DNCBM    & 83.90 & 45.91 & 96.17 \\
MSAE-RW  & 82.46 & 43.95 & 96.68 \\
SALF-CBM & 48.33 & 40.16 & 92.64 \\
CFM & 59.15 & 78.17 & 60.87 \\
DCBM & 78.79 & 85.90 & 94.87 \\
DCBM & 78.79 & 85.90 & 94.87 \\
\hline
\end{tabular}

\label{tab:unsup_test_acc}
\end{table}

\subsection{Experimental Settings}
\label{appendix:experimental_settings}

\subsubsection{Supervised CBMs}
\label{appendix:experimental_settings-training_configurations}
\label{appendix:experimental_settings-supervised}

For supervised CBMs, all benchmarked models use a ResNet-50 backbone~\citep{resnet}. 
Unless specified otherwise, training runs for 300 epochs with Adam~\citep{adam}, learning rate $1\times10^{-4}$, and batch size 512. 
Images are center-cropped to $224 \times 224$. 
Each model is trained with three random seeds, and we report average scores across seeds.

\paragraph{Concept Bottleneck Model: \textbf{CBM}.}
We train a vanilla Concept Bottleneck Model (CBM)~\citep{cbm} with concept-loss weight $\alpha=1.0$.

\paragraph{Concept Embedding Model: \textbf{CEM}.}
CEM~\citep{cem} is trained jointly with a concept-embedding size of 16, followed by a linear label head on the concatenated concept embeddings.

\paragraph{Autoregressive CBM: \textbf{AR-CBM}.}
AR-CBM~\citep{leakage_first_paper} is trained in three stages: 50 epochs of autoregressive pretraining with the encoder and label head frozen, 200 epochs of concept training, and 100 epochs of target-head training. 
At inference, predictions use $M=100$ autoregressive Monte Carlo samples.

\paragraph{Stochastic Concept Bottleneck Model: \textbf{SCBM}.}
SCBM~\citep{scbm} is trained jointly with concept-learning weight $\alpha=1.0$ and $M=100$ Monte Carlo samples. 
The concept logits use a learned covariance, amortized by default, with an $\ell_1$ off-diagonal precision regularizer of weight 1. 
Concept sampling uses a straight-through \texttt{RelaxedBernoulli} estimator with temperature annealed from 1.0 to 0.5.

\paragraph{Minimal CBM: \textbf{MCBM}.}
MCBM~\citep{was_never_cbm} is trained jointly with one latent dimension per concept, i.e., $z$-dimension per concept equal to 1, and Gaussian noise injected into $z$ during training with $\texttt{var\_z}=1.0$. 
The loss weights are $\beta=1.0$ for concept BCE, $\gamma=0.3$, and $\texttt{rep\_loss\_weight}=0.2$ for the representation MSE. 
We use a fixed concept-to-latent mapping with $\texttt{z\_mapping\_scale}=6.0$ and $\texttt{z\_mapping\_bias}=-3.0$, together with linear concept and label heads.

\subsubsection{Unsupervised CBMs}
\label{appendix:experimental_settings-naming-unsupervised}
\label{appendix:experimental_settings-unsupervised}

For unsupervised CBMs, models use their original concept-discovery pipelines and backbone image encoders rather than the supervised ResNet-50 training setup described above. 
We benchmark five unsupervised variants: DN-CBM~\citep{dncbm}, MSAE~\citep{msae}, SALF-CBM~\citep{salf_cbm}, CFM~\citep{cfm}, and D-CBM~\citep{dcbm}. 
Below, we summarize the configuration used for each model.

\paragraph{Discover-then-Name CBM: \textbf{DN-CBM}~\citep{dncbm}.}
DN-CBM discovers concepts by training a sparse autoencoder over CLIP image representations. 
It uses CLIP ViT-L/14 as the image encoder and applies the SAE to the CLIP output representation. 
The SAE uses an expansion factor of 8, producing 6144 discovered concepts from the 768-dimensional CLIP representation.

\paragraph{Matryoshka Sparse Autoencoder with Reverse Weighting: \textbf{MSAE}~\citep{msae}.}
MSAE-RW uses a Matryoshka sparse autoencoder with TopKReLU activations and reverse-weighted training. 
The model is built on ViT-L/14 image features and uses an expansion factor of 8, resulting in 6144 discovered concepts.
The Matryoshka structure contains three nested shells with sizes $[64,128,256]$, giving three levels of concept granularity.

\paragraph{Spatially-Aware and Label-Free CBM: \textbf{SALF-CBM}~\citep{salf_cbm}.}
SALF-CBM uses an ImageNet-pretrained ResNet-50 image encoder. 
The resulting concept bank contains 4741 discovered concepts. 

\paragraph{Concept Foundational Model: \textbf{CFM}~\citep{cfm}}
CFM uses a Matryoshka BatchTopK sparse autoencoder.
The model is built on DINO-CLIP ViT-B/16 image features and uses an expansion factor of 16, resulting in 8192 discovered concepts. 
Its Matryoshka structure is organized into six concept groups with sizes $[65,245,491,983,1966,4442]$.

\paragraph{Data-efficient Concept Bottleneck Model: \textbf{D-CBM}~\citep{dcbm}.}
D-CBM discovers concepts by clustering visual segments generated with SAM2, enabling concept discovery with limited annotation requirements.
It represents each segment using CLIP ViT-L/14 image features and constructs concept clusters via k-means, with median centroids used to obtain more robust concept prototypes.

\FloatBarrier

\subsubsection{Used Datasets. }
\label{appendix:experimental_settings-datasets}
We evaluate on CUB, PartImageNet, and a single-object subset of Pascal-Part, as these datasets provide the concept and spatial annotations required for our grounding and completeness analyses. Together, they span fine-grained and part-centric recognition settings with different concept granularities. \Cref{tab:dataset_stats} summarizes the dataset sizes, train/test splits, and concept counts used in our benchmark.

\begin{table}[!htbp]
\centering
\caption{Dataset statistics for the benchmark.}
\resizebox{0.9\textwidth}{!}{\begin{tabular}{lccccc}
\toprule
Dataset & \#Images (Total) & \#Train & \#Test & \#Classes & \#Concepts \\
\midrule
CUB & 11,788 & 4,796 & 5,794 & 200 & 112 \\
PartImageNet & 24,067 & 20,457 & 2,405 & 158 & 40 \\
PascalPart (single-object) & 4,737 & 2,358 & 2,379 & 20 & 89 \\
\bottomrule
\end{tabular}
}

\label{tab:dataset_stats}
\end{table}

\noindent{\textbf{Pascal-Part Single-Object Subset Dataset. }}
\label{appendix:experimental_settings-pascal_part}
Beyond CUB \citep{cub2002011} and PartImageNet \citep{partimagenet}, Pascal-Part \citep{chen2014detect,everingham2010pascal} is one of the few benchmarks that provides rich object- and part-level annotations. However, Pascal-Part is originally a multi-label dataset, where multiple annotated objects and their parts may co-occur in the same image. Although EPG can be computed in such a setting, the resulting localization scores are more difficult to interpret, as evidence assigned to a target concept may be influenced by the presence of other objects and their parts. To remove this ambiguity and maintain consistency with CUB and PartImageNet, we construct a single-object subset of Pascal-Part by retaining only images containing one annotated object. The resulting subset contains 4,737 images, split into 2,358 training and 2,379 validation images, covering 20 object classes and 89 part concepts.

\begin{figure}[th]
    \centering
    \includegraphics[width=0.5\textwidth]{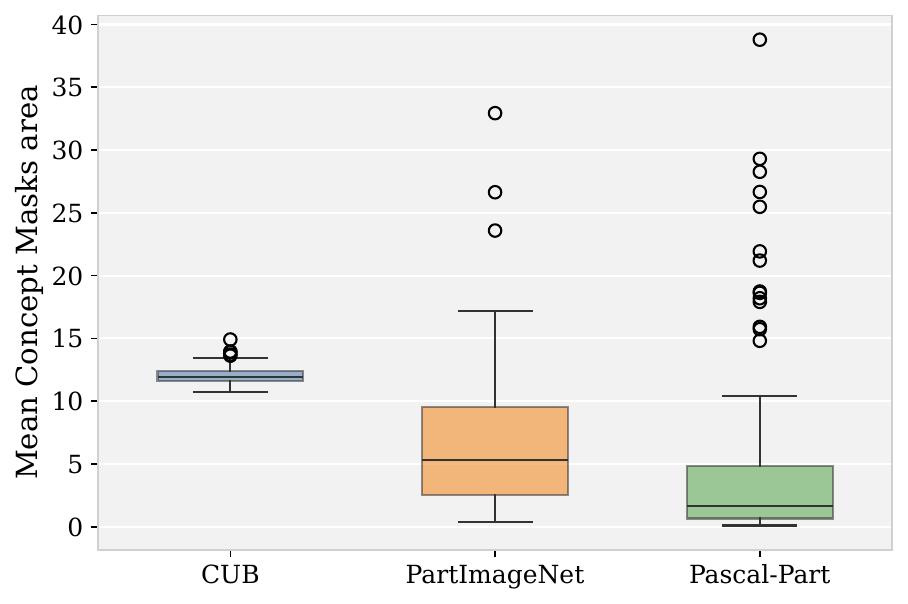}
    \caption{Distribution of concept sizes across datasets, quantified as the average concept-mask area.}
    \label{app-fig:dataset_mask_ratios}
\end{figure}

\FloatBarrier

\subsection{\response{Human Validation of VPS and CPE}}
\label{appendix:vps-user-study}

\begin{figure}[!htbp]
    \centering
    \includegraphics[width=\textwidth]
    {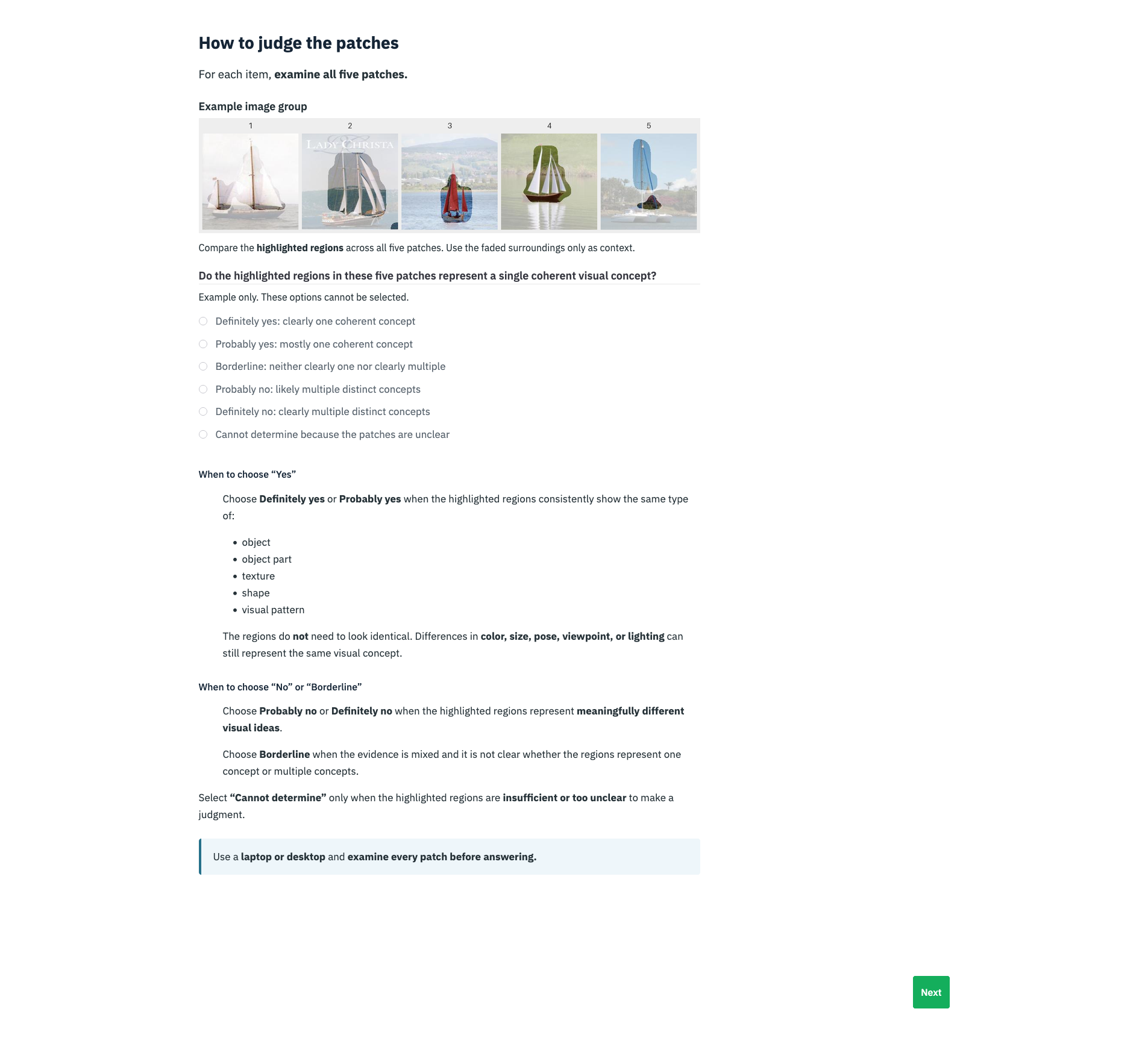}
    \caption{\response{Instructions shown to participants
    before answering the survey questions.}}
    \label{app-fig:survey-introduction}
\end{figure}

\begin{figure}[!htbp]
    \centering
    \includegraphics[width=\textwidth]
    {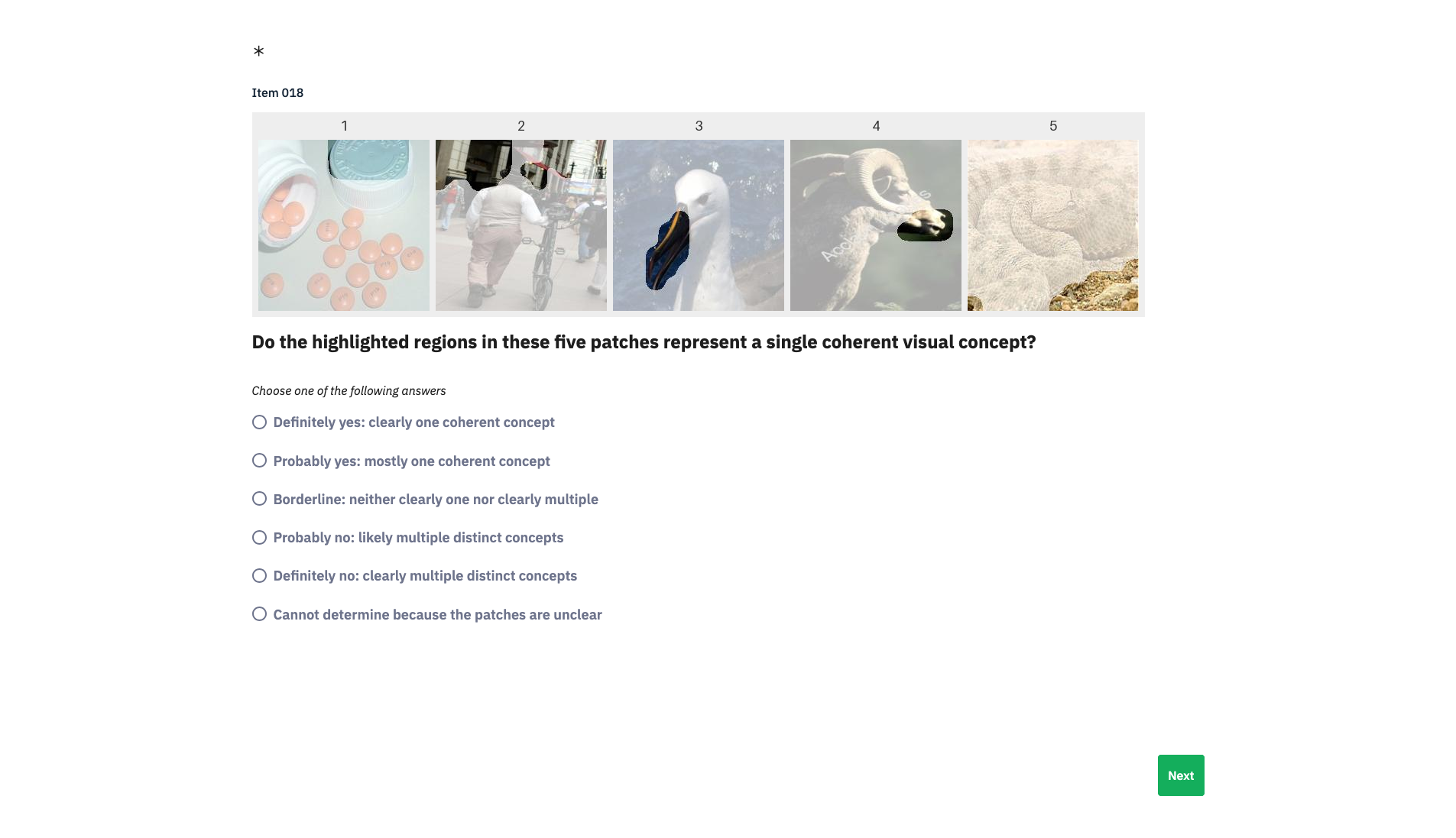}
    \caption{\response{An example survey question showing
    five highlighted image regions and the available
    answer options.}}
    \label{app-fig:survey-example}
\end{figure}

\paragraph{\response{Study design.}}
\response{We conducted a user study with 45 participants to assess the
alignment of VPS and CPE with human judgments of concept
polysemanticity. The study included 60 concepts divided into three
groups, with groups presented at random to participants. The survey instructions and an example question
are shown in \Cref{app-fig:survey-introduction,app-fig:survey-example}.}

\paragraph{\response{Participant agreement and metric alignment.}}
\response{Participant agreement measures how consistently different
participants rate the same concepts. We measured it using interval
Krippendorff's $\alpha$, obtaining $\alpha = 0.698$.
\Cref{app-tab:vps-user-study} reports the Spearman correlations between
human judgments and the two metrics. Both correlations are positive
and statistically significant, with a numerically larger coefficient
for VPS. The reported $p$-values concern the individual associations;
they do not establish a statistically significant difference between
the two correlations.}

\begin{table}[!htbp]
    \centering
    \caption{\response{Spearman correlations between human judgments
    of concept polysemanticity and VPS or CPE.}}
    \label{app-tab:vps-user-study}
    \small
    \begin{tabular}{lcc}
        \toprule
        \response{Metric}
        & \response{Spearman correlation ($\rho$)}
        & \response{$p$-value} \\
        \midrule
        \response{VPS}
        & \response{$+0.513$}
        & \response{$0.023$} \\
        \response{CPE}
        & \response{$+0.175$}
        & \response{$0.002$} \\
        \bottomrule
    \end{tabular}
\end{table}

\clearpage
\subsection{\response{VPS Sensitivity to the Visual Feature Extractor}}
\label{appendix:vps-backbone-sensitivity}

\begin{figure}[!htbp]
    \centering
    \captionsetup{font=footnotesize}

    \includegraphics[width=\textwidth]
    {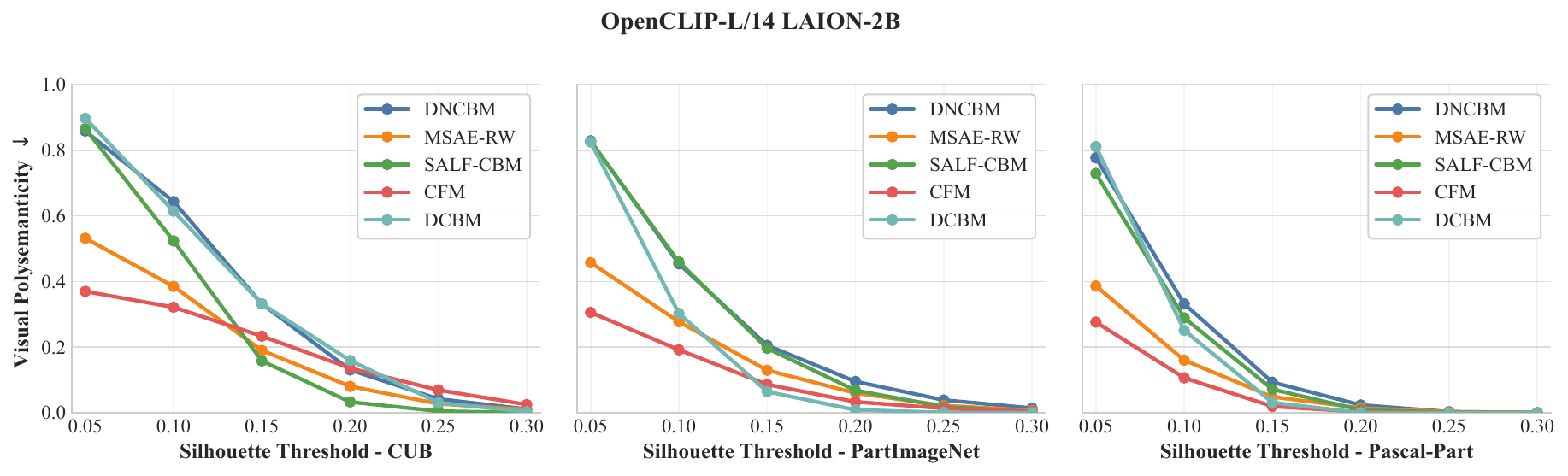}

    \par\medskip

    \includegraphics[width=\textwidth]
    {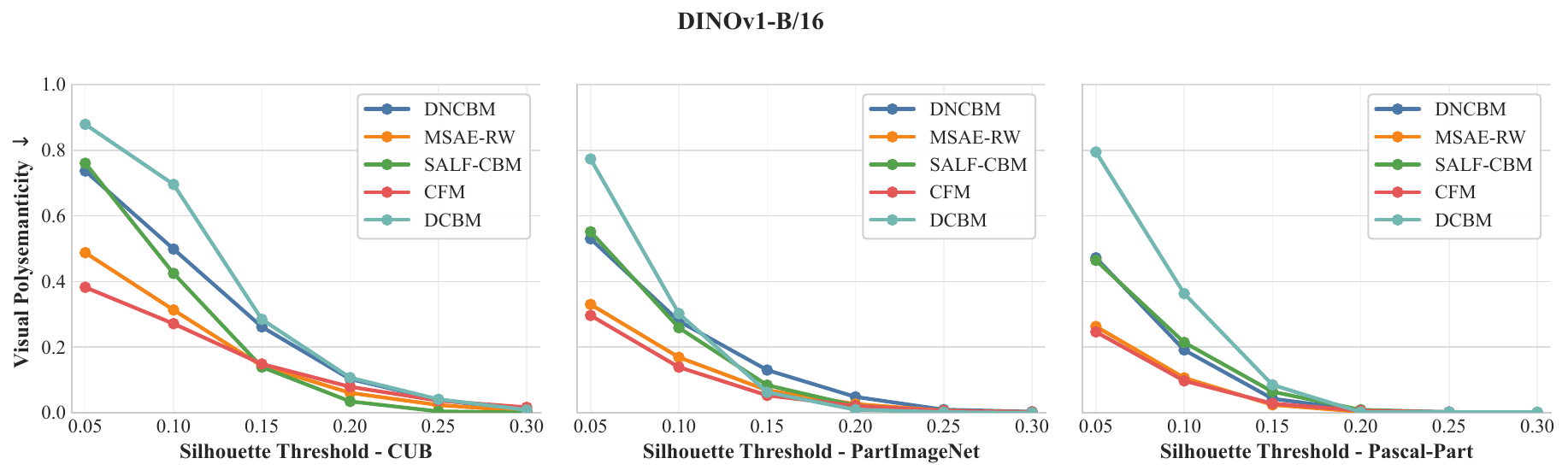}

    \caption{\response{\textbf{VPS sensitivity to the visual feature
    extractor and silhouette threshold.}
    VPS computed using OpenCLIP-L/14 (LAION-2B), top, and DINOv1-B/16,
    bottom. Columns show CUB, PartImageNet, and Pascal-Part,
    respectively. Lower values indicate a smaller fraction of
    concepts classified as visually polysemantic at the specified
    threshold. CFM and MSAE-RW have the lowest values at low
    thresholds, while separation and relative rankings can change
    as the threshold increases.}}
    \label{app-fig:vps-backbone-sensitivity}
\end{figure}

\response{To assess the dependence of VPS on the visual feature
extractor, we evaluate it using OpenCLIP-L/14 (LAION-2B) and
DINOv1-B/16, complementing the DINOv2-based results in
\Cref{fig:unsupervised_visual_mono_semantic}.
\Cref{app-fig:vps-backbone-sensitivity} presents the resulting VPS
curves over silhouette thresholds on CUB, PartImageNet, and
Pascal-Part.}

\response{For both alternative backbones and all three datasets,
the two Matryoshka-based models, CFM and MSAE-RW, have the two lowest VPS values at the low silhouette thresholds $\tau \in \{0.05, 0.10\}$. As the threshold increases, differences between methods generally narrow and some rankings change, especially on CUB dataset.}

\clearpage
\subsection{\response{Evaluation on MS COCO}}
\label{appendix:coco-evaluation}

\response{We evaluate the five unsupervised CBMs on all 5,000 images
in the MS COCO 2017 validation set without additional training.
\Cref{app-tab:coco-evaluation} reports VPS using DINOv2 features,
the top-$15$ examples per concept, and a silhouette threshold of
$\tau=0.10$, alongside mean semantic coherence.}

\begin{table}[!htbp]
    \centering
    \small
    \caption{\response{VPS and mean semantic coherence for unsupervised
    CBMs on MS COCO.}}
    \label{app-tab:coco-evaluation}
    \response{\begin{tabular}{lcc}
        \toprule
        \textbf{Model} & \textbf{VPS (\%) $\downarrow$}
        & \textbf{Mean Semantic Coherence $\uparrow$} \\
        \midrule
        CFM      & 33.08 & 0.267 \\
        DN-CBM   & 61.72 & 0.185 \\
        MSAE     & 39.48 & 0.349 \\
        SALF-CBM & 65.24 & 0.160 \\
        D-CBM    & 38.13 & 0.194 \\
        \bottomrule
    \end{tabular}}
\end{table}

\response{\Cref{app-tab:coco-evaluation} shows that CFM retains
the lowest VPS, while MSAE achieves the highest mean semantic
coherence, followed by CFM. The same trend continues here, where the Matryoshka-based CBM models, namely CFM and MSAE are the best models in terms of monosemanticity, while now we have D-CBM being the second-best model, which can be attributed to its training scheme of building prototypes by clustering segmented image regions~\citep{dcbm}. This evaluation extends the use of both metrics to multi-object images without requiring concept annotations.}

\subsection{\response{Computational Cost}}
\label{appendix:computational-cost}

\response{We measure each evaluation on all 2,405 PartImageNet test
images. \Cref{app-tab:computational-cost}
reports measurements on one NVIDIA L40S GPU with four CPU threads and
an image batch size of four.}
\response{Times exclude common model and dataset setup and include
metric-specific feature loading and result writing. GPU memory is peak PyTorch tensor allocation during
evaluation. CPU RAM is peak process memory, including initialization.}

\begin{table}[!htbp]
    \centering
    \small
    \caption{\response{Wall-clock time and peak memory for VPS and
    semantic coherence on the PartImageNet test set.}}
    \label{app-tab:computational-cost}
    \response{\begin{tabular}{llrrr}
        \toprule
        \textbf{Model} & \textbf{Metric}
        & \makecell{\textbf{Time} \\ \textbf{(min)}}
        & \makecell{\textbf{Peak GPU memory} \\ \textbf{(GiB)}}
        & \makecell{\textbf{Peak CPU RAM} \\ \textbf{(GiB)}} \\
        \midrule
        \multirow{2}{*}{CFM}
         & VPS                & 35.20  & 4.46  & 3.25 \\
         & Semantic Coherence & 8.43   & 4.48  & 3.02 \\
        \midrule
        \multirow{2}{*}{DN-CBM}
         & VPS                & 38.67  & 6.35  & 4.77 \\
         & Semantic Coherence & 142.20 & 10.04 & 4.77 \\
        \midrule
        \multirow{2}{*}{MSAE}
         & VPS                & 51.68  & 6.35  & 4.77 \\
         & Semantic Coherence & 60.24  & 10.04 & 4.77 \\
        \midrule
        \multirow{2}{*}{SALF-CBM}
         & VPS                & 35.88  & 4.00  & 2.90 \\
         & Semantic Coherence & 126.96 & 4.00  & 2.52 \\
        \midrule
        \multirow{2}{*}{D-CBM}
         & VPS                & 46.19  & 6.30  & 4.77 \\
         & Semantic Coherence & 468.32 & 9.99  & 4.77 \\
        \bottomrule
    \end{tabular}}
\end{table}

\response{\Cref{app-tab:computational-cost} shows that runtime varies substantially with the model and metric, despite using the same images and hardware.
VPS takes 35.20 to 51.68 minutes, whereas semantic coherence ranges from 8.43 minutes for CFM to 468.32 minutes for D-CBM.
Semantic coherence is therefore the more time-consuming evaluation for four of the five models, with CFM as the exception.
Across both metrics, peak GPU tensor memory ranges from 4.00 to 10.04\,GiB and peak CPU RAM from 2.52 to 4.77\,GiB.
These results show why one runtime estimate cannot represent the cost of evaluating every CBM.
Reporting time and memory separately for each model and metric provides concrete budgets for incorporating these concept-quality checks into evaluation pipelines.
}

\clearpage
\subsection{Qualitative Examples of Attribution Maps}
\label{appendix:attribution_maps}

As a complement to the quantitative EPG results for the supervised models (see \Cref{fig:concept_epg_gradcam_all_datasets}), we present attribution map overlays for selected concepts across all models and datasets. Specifically, we show results for three attribution methods: GradCAM, Guided GradCAM, and Input$\times$Gradient. Note that the EPG results reported in the main paper in \Cref{fig:concept_epg_gradcam_all_datasets} are based on GradCAM. In all qualitative panels, attributions are rendered as a red overlay on a grayscale version of the input image to highlight salient regions while preserving structural context and improving visual comparability across models and methods.

\begin{figure}[!htbp]
    \centering
    \captionsetup{font=footnotesize, skip=0.0pt}
    \includegraphics[width=1.0\textwidth]{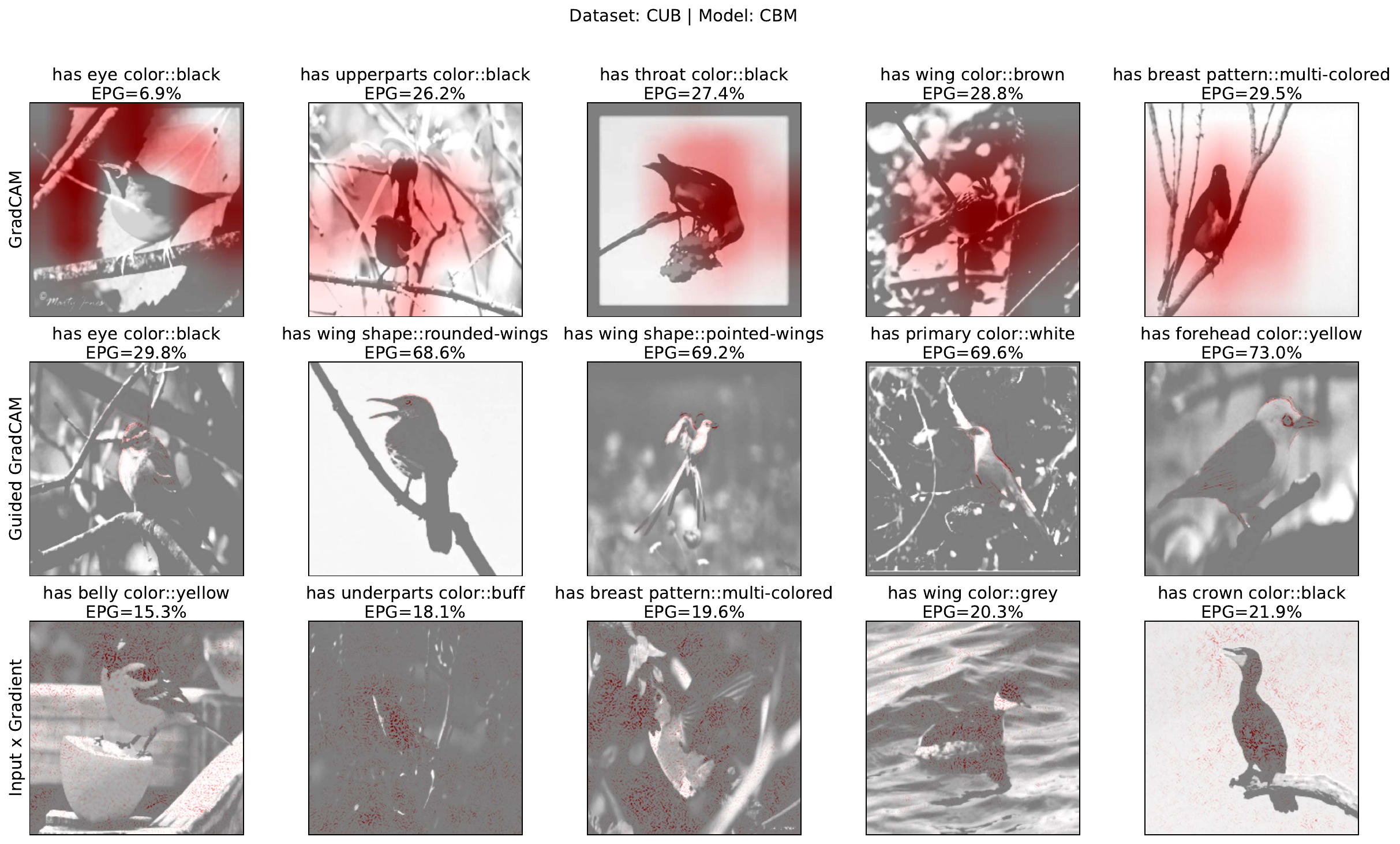}
    \caption{Attribution map visualizations for the vanilla CBM on the CUB dataset.}
    \label{app-fig:cbm-cub-attributions}
\end{figure}

\begin{figure}[!htbp]
    \centering
    \captionsetup{font=footnotesize, skip=0.0pt}
    \includegraphics[width=1.0\textwidth]{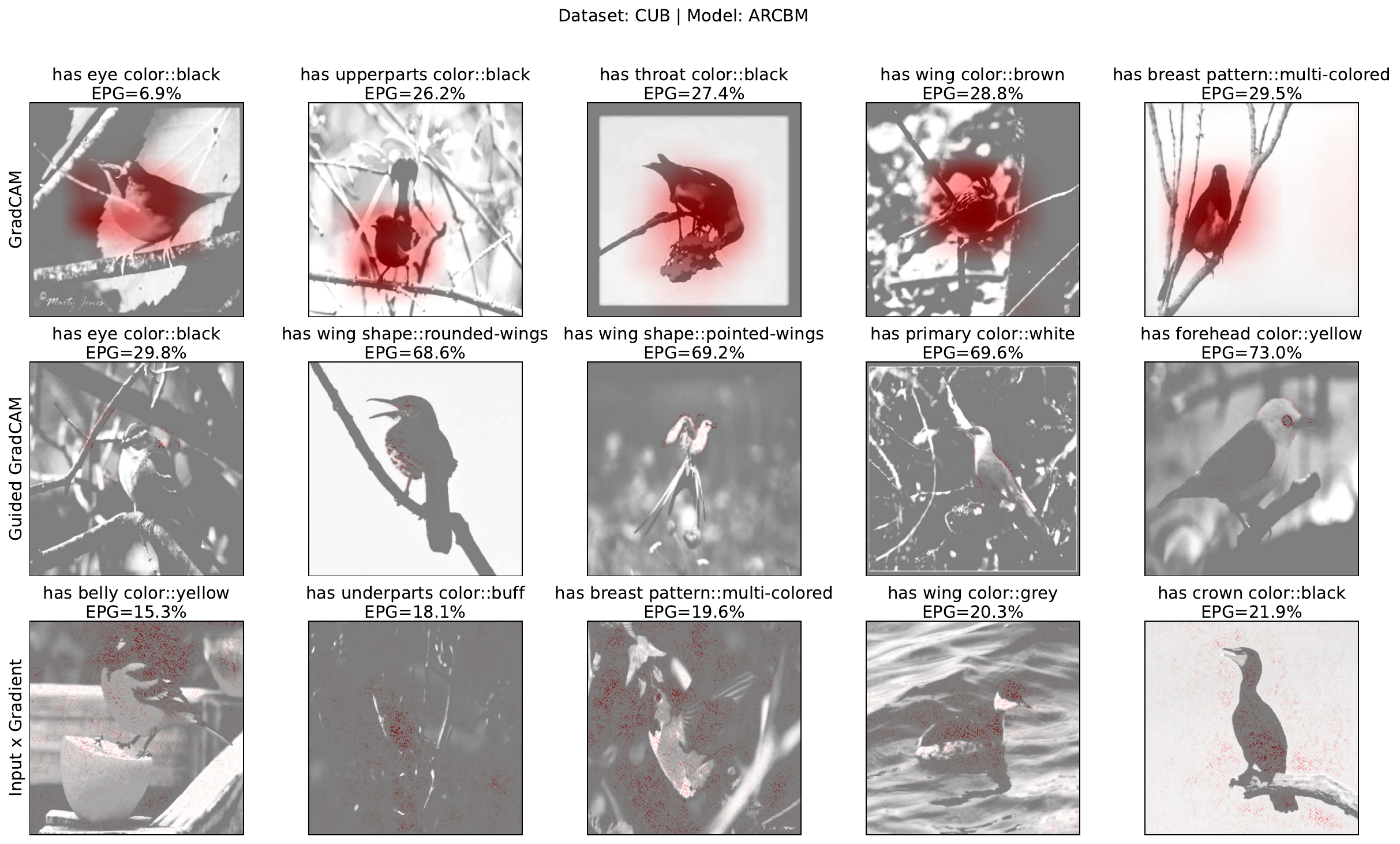}
    \caption{Attribution maps visualization for the  Auto-Regressive CBM on the CUB dataset.}
    \label{app-fig:arcbm-cub-attributions}
\end{figure}

\begin{figure}[!htbp]
    \centering
    \captionsetup{font=footnotesize, skip=0.0pt}
    \includegraphics[width=1.0\textwidth]{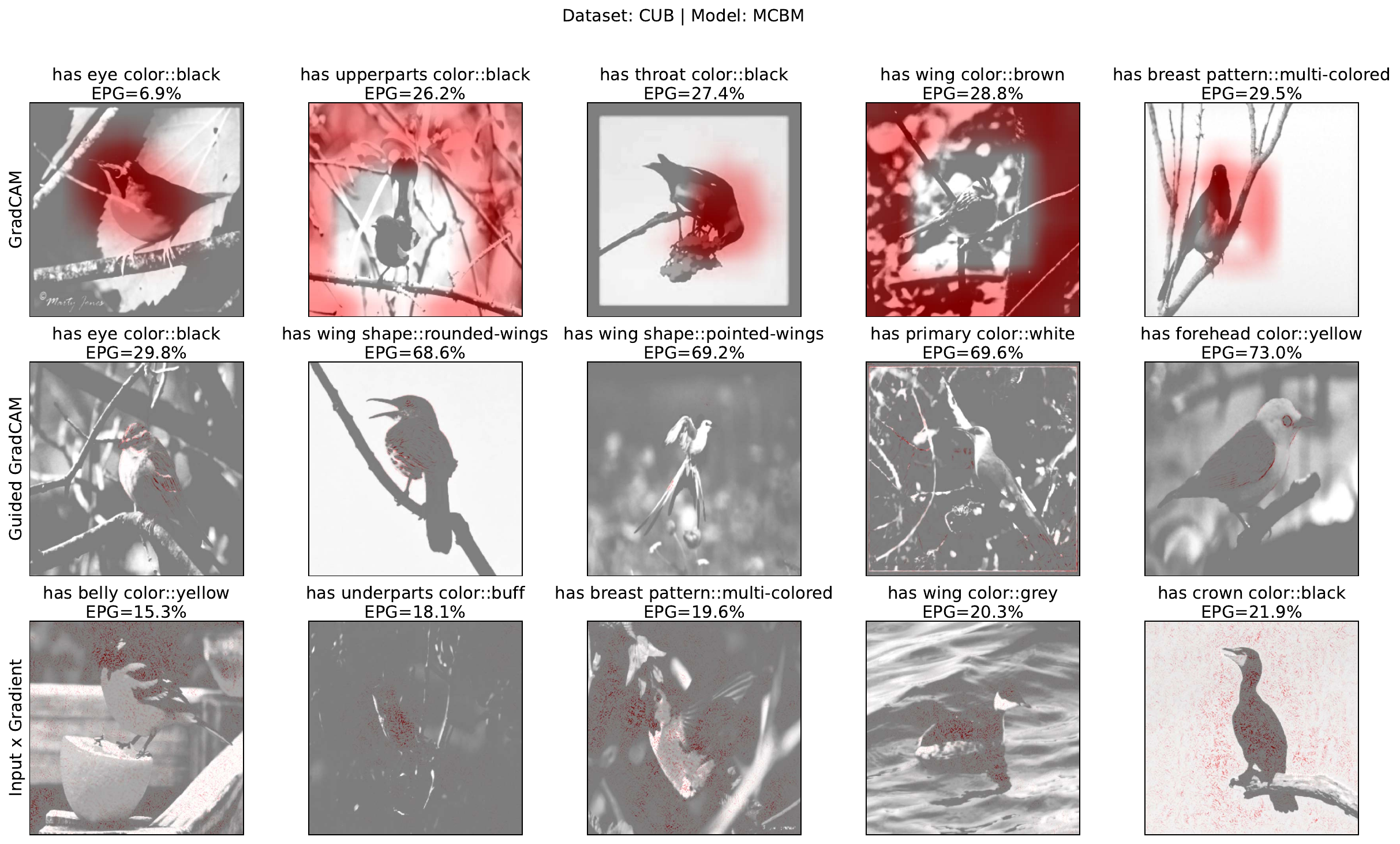}
    \caption{Attribution maps visualization for the  MCBM on the CUB dataset.}
    \label{app-fig:mcbm-cub-attributions}
\end{figure}

\begin{figure}[!htbp]
    \centering
    \captionsetup{font=footnotesize, skip=0.0pt}
    \includegraphics[width=1.0\textwidth]{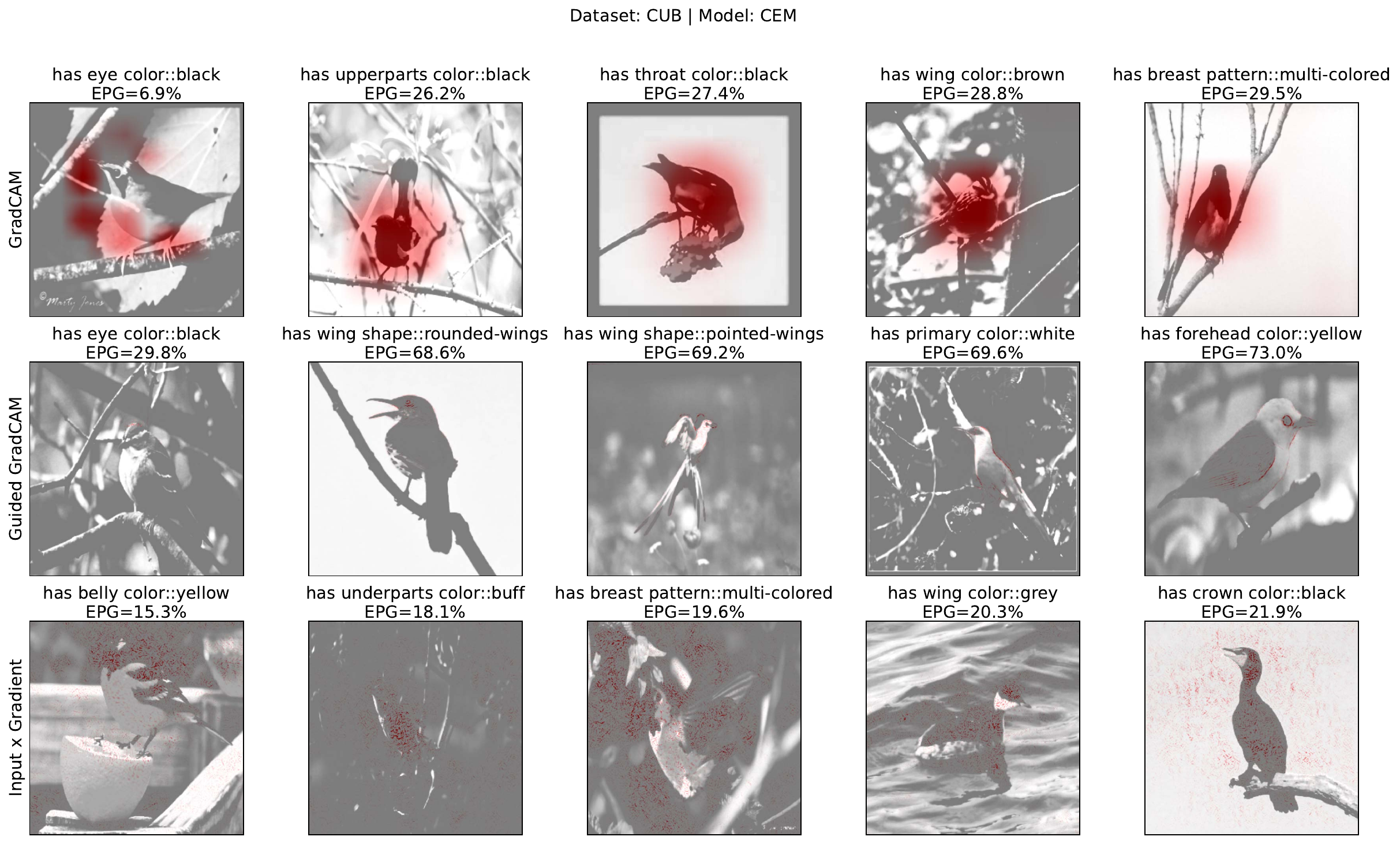}
    \caption{Attribution maps visualization for the CEM on the CUB dataset.}
    \label{app-fig:cem-cub-attributions}
\end{figure}

\begin{figure}[!htbp]
    \centering
    \captionsetup{font=footnotesize, skip=0.0pt}
    \includegraphics[width=1.0\textwidth]{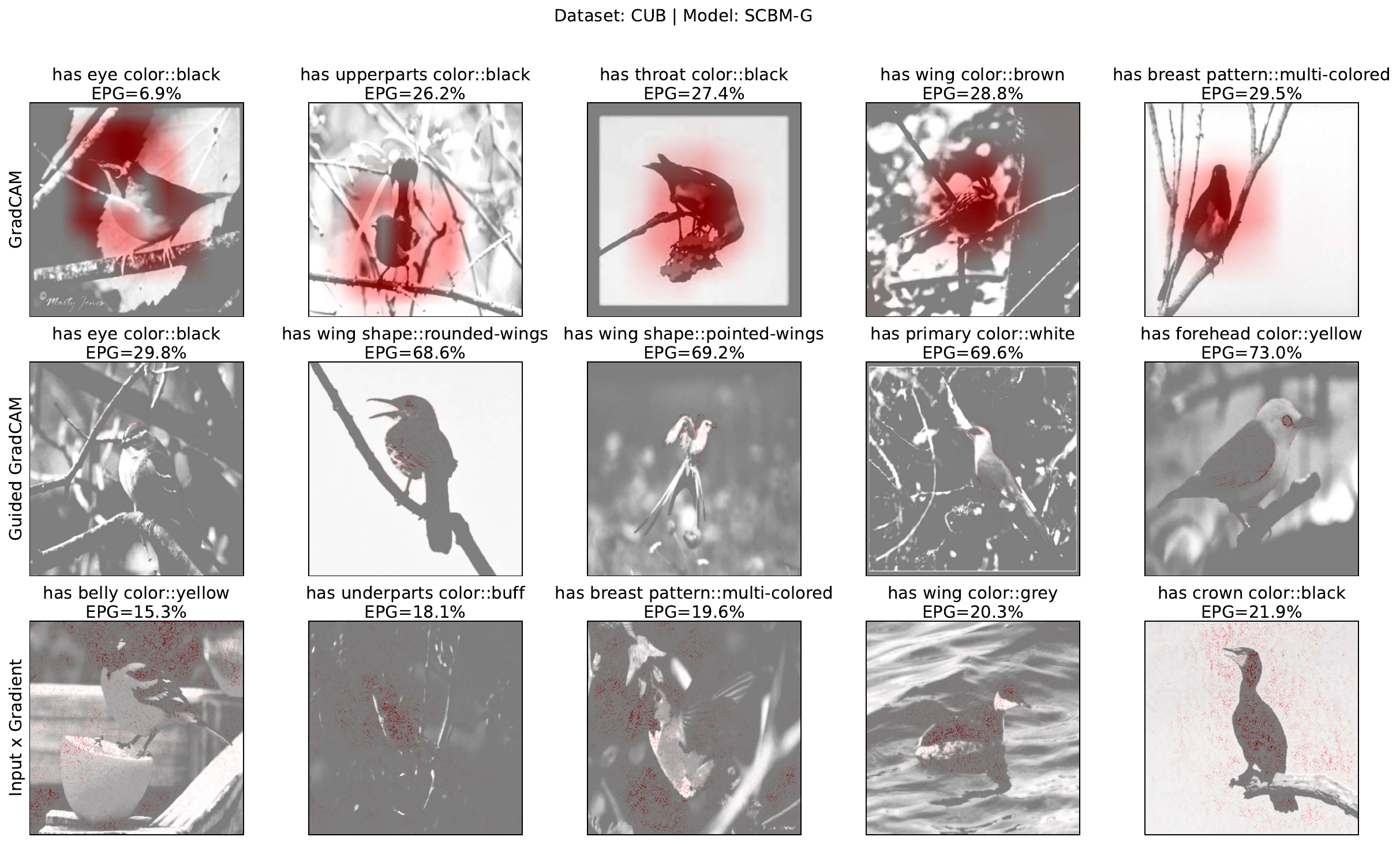}
    \caption{Attribution maps visualization for the SCBM-Amortized on the CUB dataset.}
    \label{app-fig:scbm_a-cub-attributions}
\end{figure}

\begin{figure}[!htbp]
    \centering
    \captionsetup{font=footnotesize, skip=0.0pt}
    \includegraphics[width=1.0\textwidth]{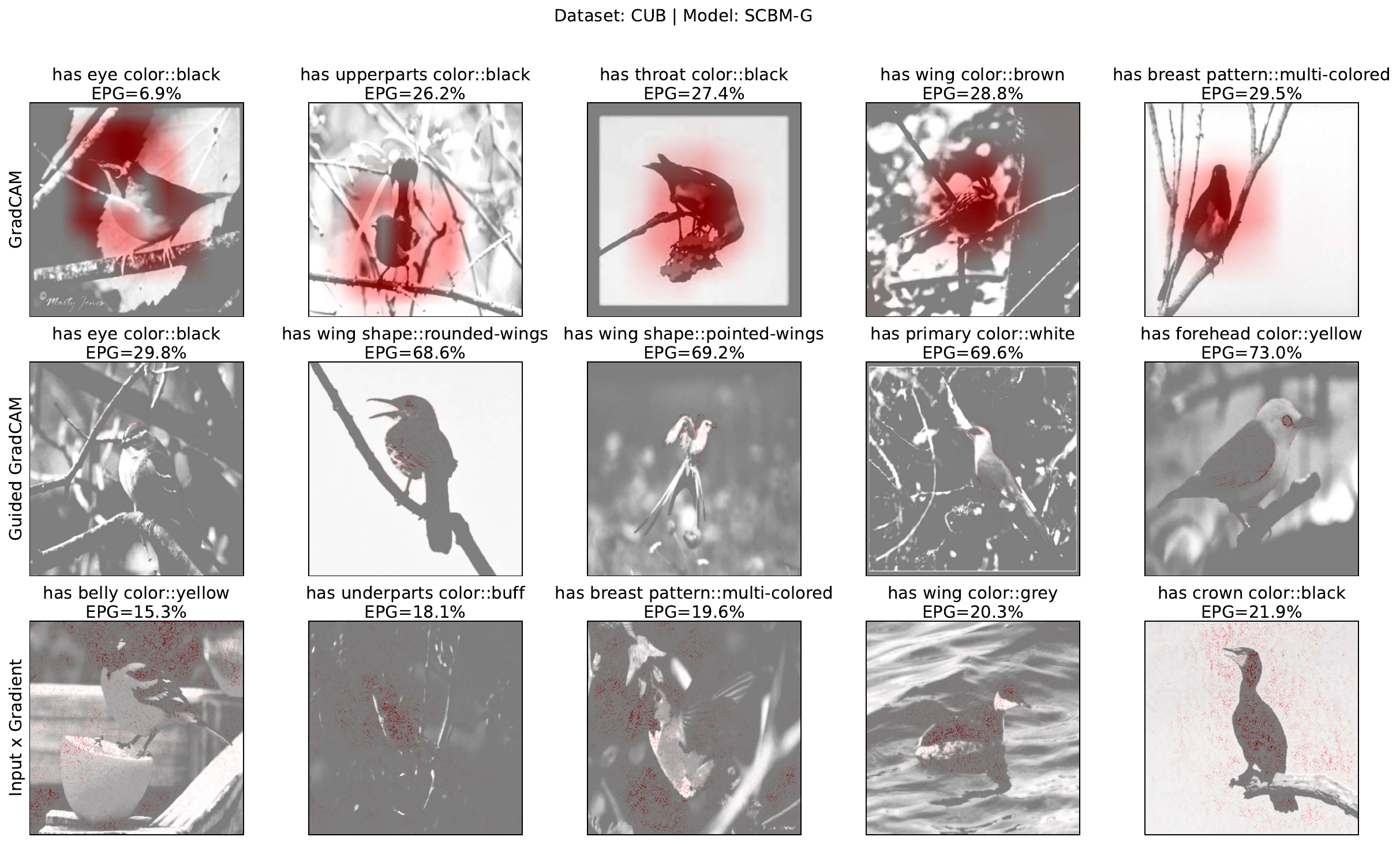}
    \caption{Attribution maps visualization for the SCBM-Global on the CUB dataset.}
    \label{app-fig:scbm_g-cub-attributions}
\end{figure}

\begin{figure}[!htbp]
    \centering
    \captionsetup{font=footnotesize, skip=0.0pt}
    \includegraphics[width=1.0\textwidth]{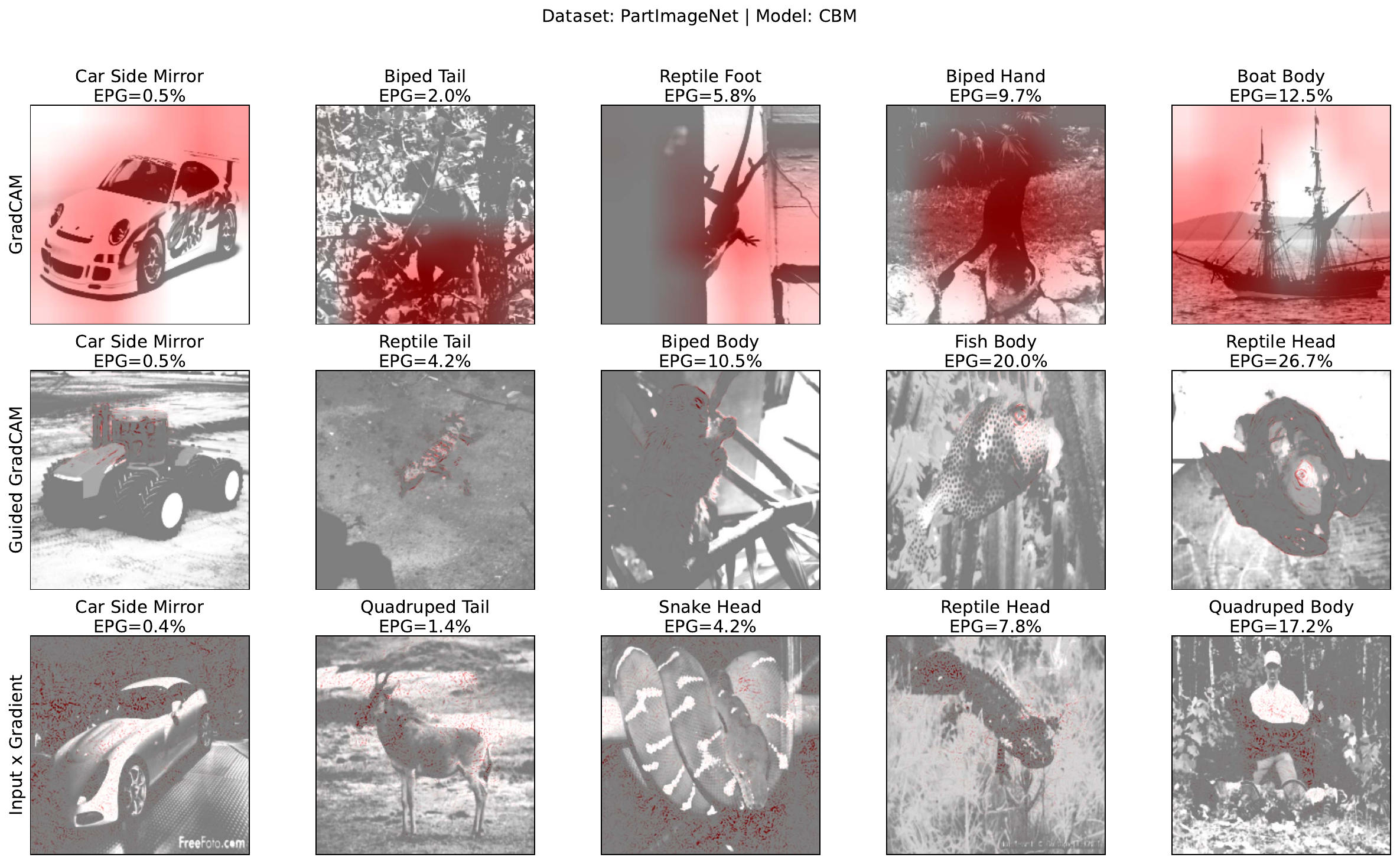}
    \caption{Attribution maps visualization for the vanilla CBM on the PartimageNet dataset.}
    \label{app-fig:cbm-partimagenet-attributions}
\end{figure}

\begin{figure}[!htbp]
    \centering
    \captionsetup{font=footnotesize, skip=0.0pt}
    \includegraphics[width=1.0\textwidth]{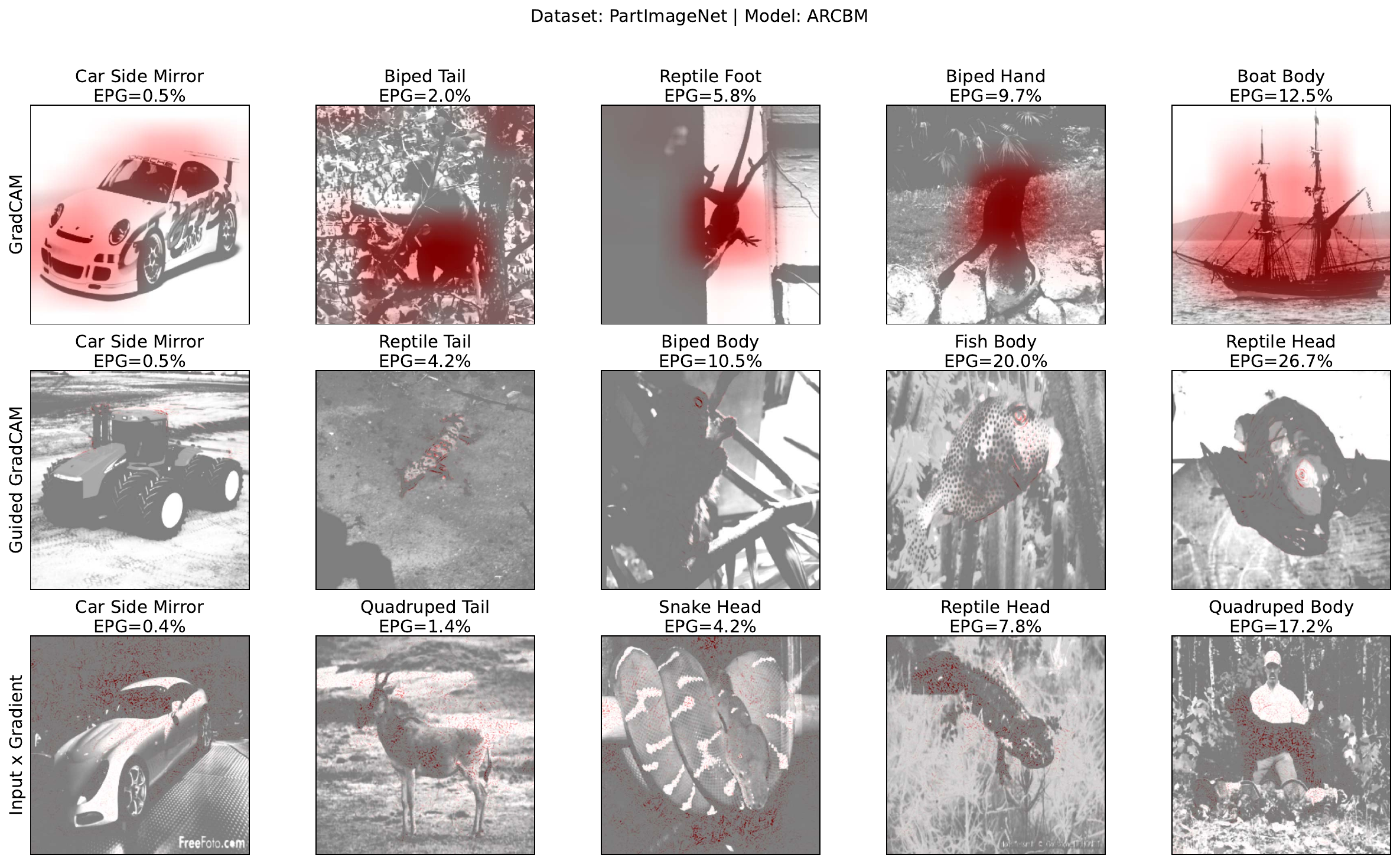}
    \caption{Attribution maps visualization for the Auto-Regressive CBM on the PartimageNet dataset.}
    \label{app-fig:arcbm-partimagenet-attributions}
\end{figure}

\begin{figure}[!htbp]
    \centering
    \captionsetup{font=footnotesize, skip=0.0pt}
    \includegraphics[width=1.0\textwidth]{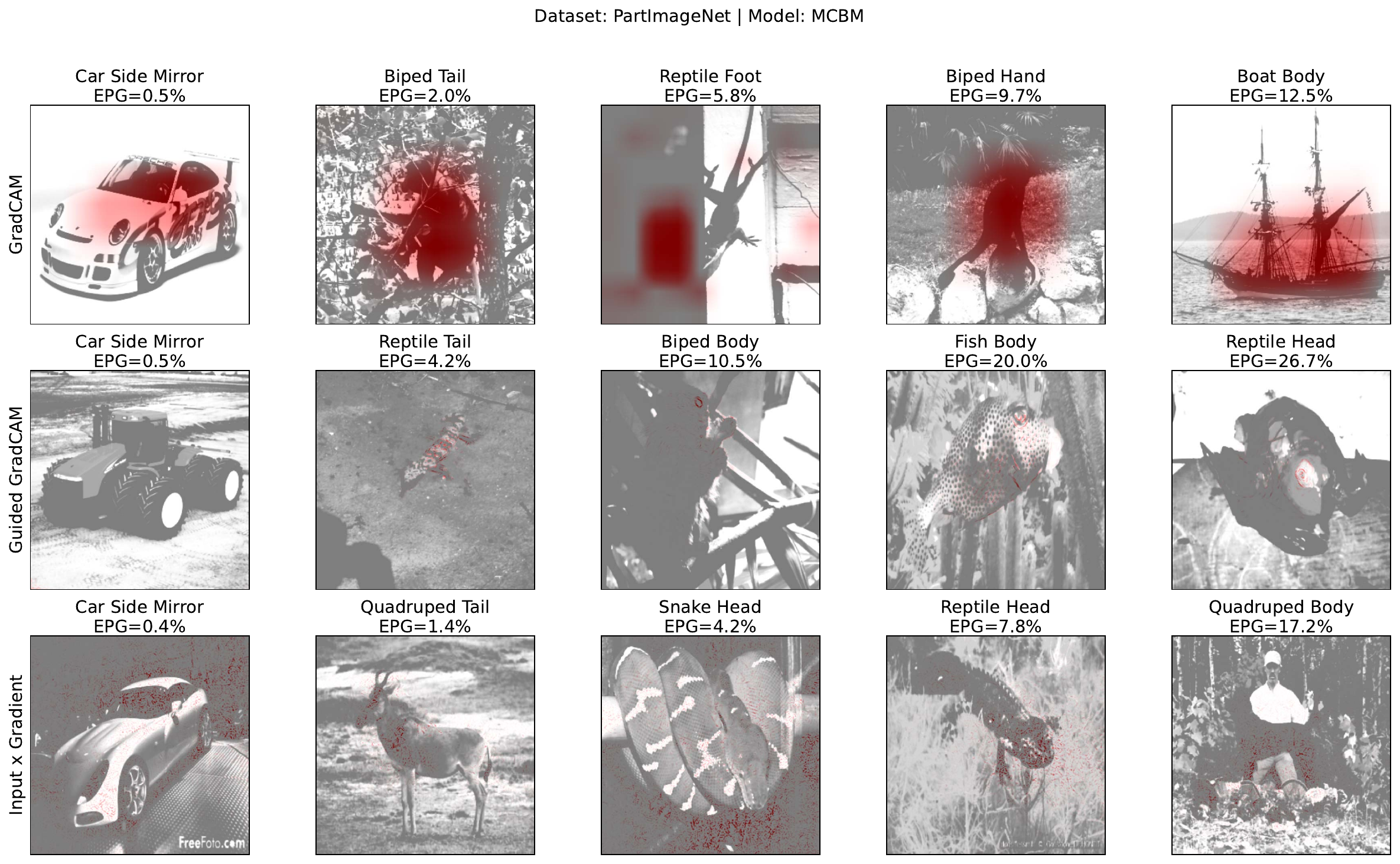}
    \caption{Attribution maps visualization for the MCBM on the PartimageNet dataset.}
    \label{app-fig:mcbm-partimagenet-attributions}
\end{figure}

\begin{figure}[!htbp]
    \centering
    \captionsetup{font=footnotesize, skip=0.0pt}
    \includegraphics[width=1.0\textwidth]{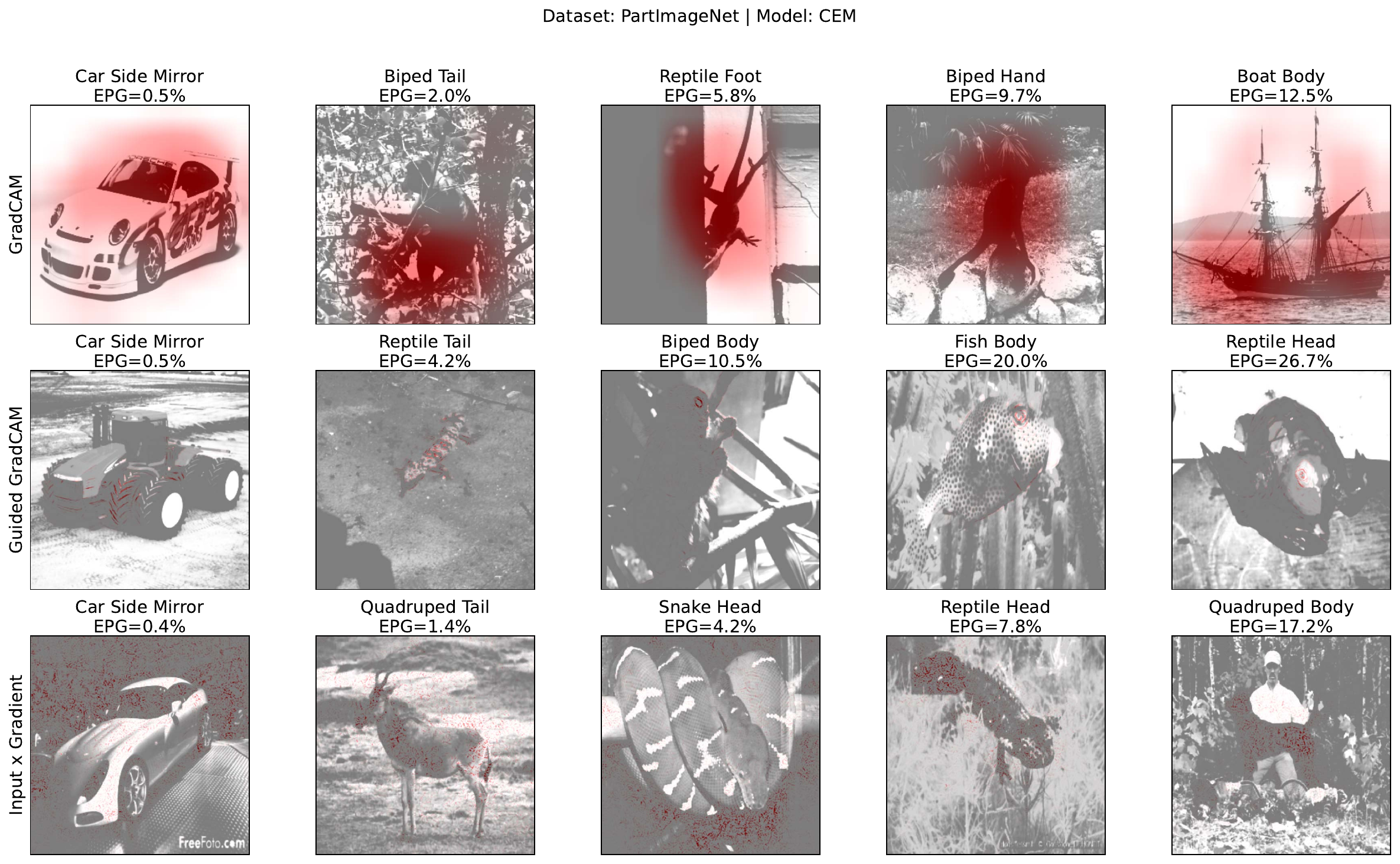}
    \caption{Attribution maps visualization for the CEM on the PartimageNet dataset.}
    \label{app-fig:cem-partimagenet-attributions}
\end{figure}

\begin{figure}[!htbp]
    \centering
    \captionsetup{font=footnotesize, skip=0.0pt}
    \includegraphics[width=1.0\textwidth]{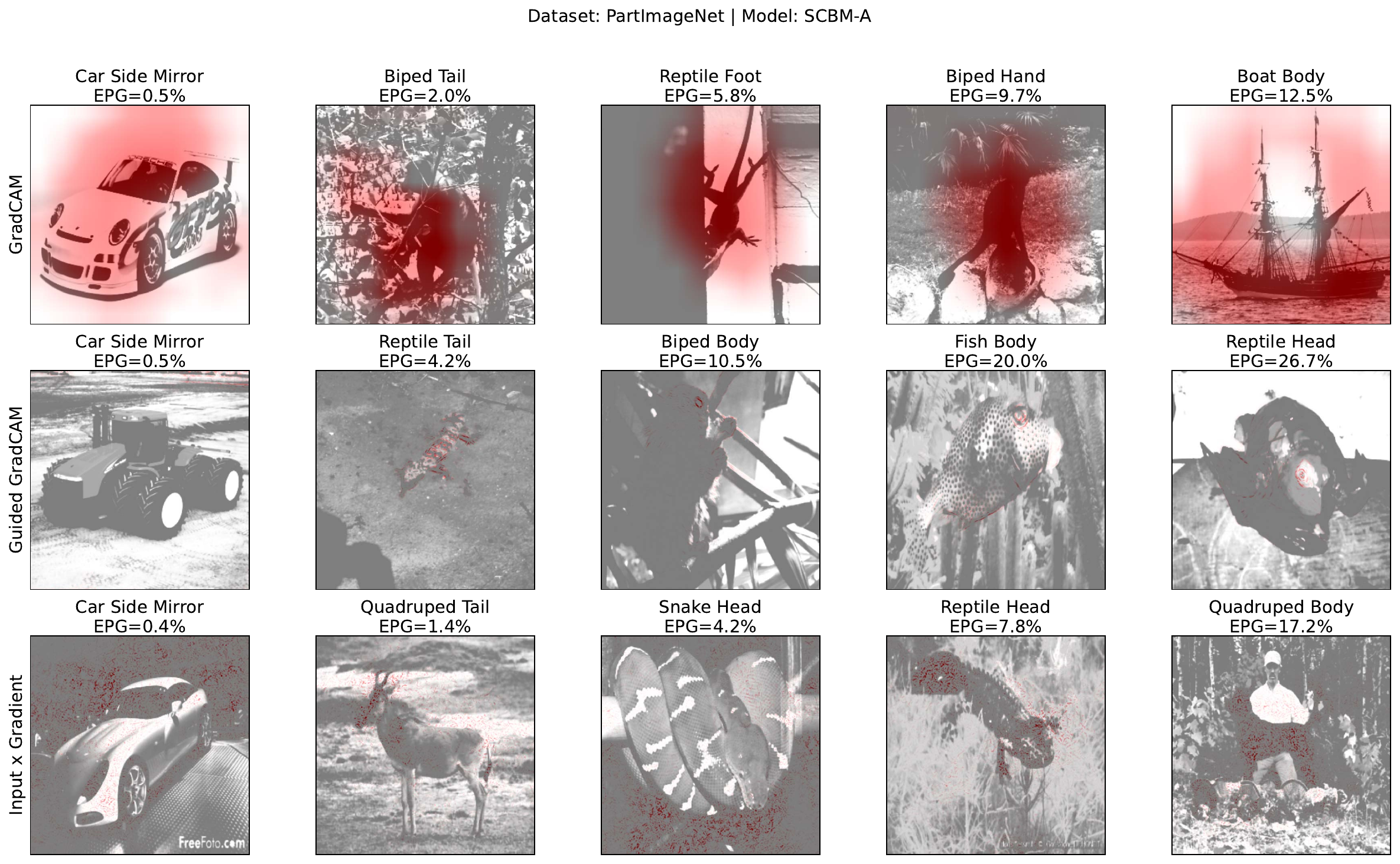}
    \caption{Attribution maps visualization for the SCBM-Amortized on the PartimageNet dataset.}
    \label{app-fig:scbm_a-partimagenet-attributions}
\end{figure}

\begin{figure}[!htbp]
    \centering
    \captionsetup{font=footnotesize, skip=0.0pt}
    \includegraphics[width=1.0\textwidth]{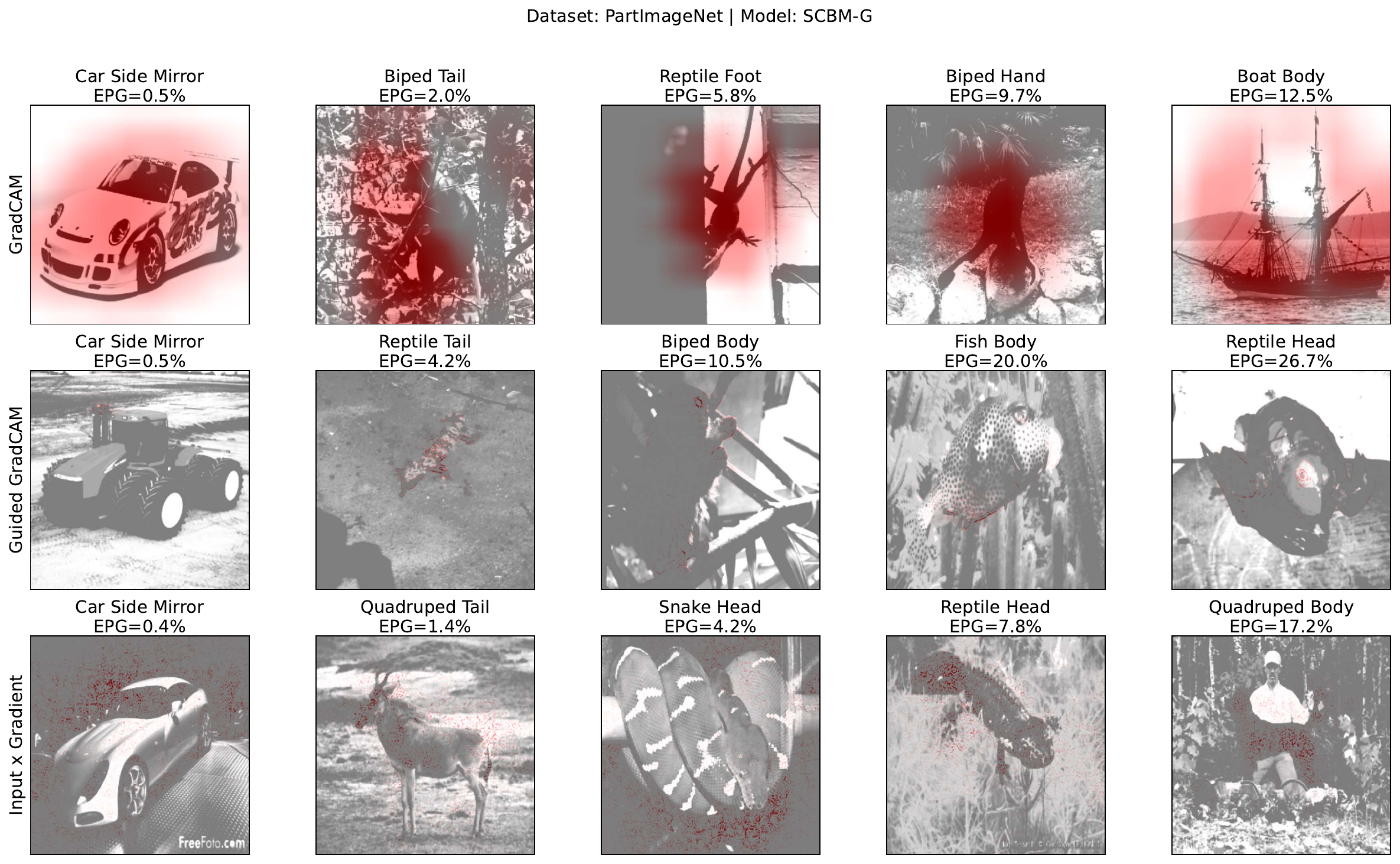}
    \caption{Attribution maps visualization for the SCBM-Global on the PartimageNet dataset.}
    \label{app-fig:scbm_g-partimagenet-attributions}
\end{figure}

\begin{figure}[!htbp]
    \centering
    \captionsetup{font=footnotesize, skip=0.0pt}
    \includegraphics[width=1.0\textwidth]{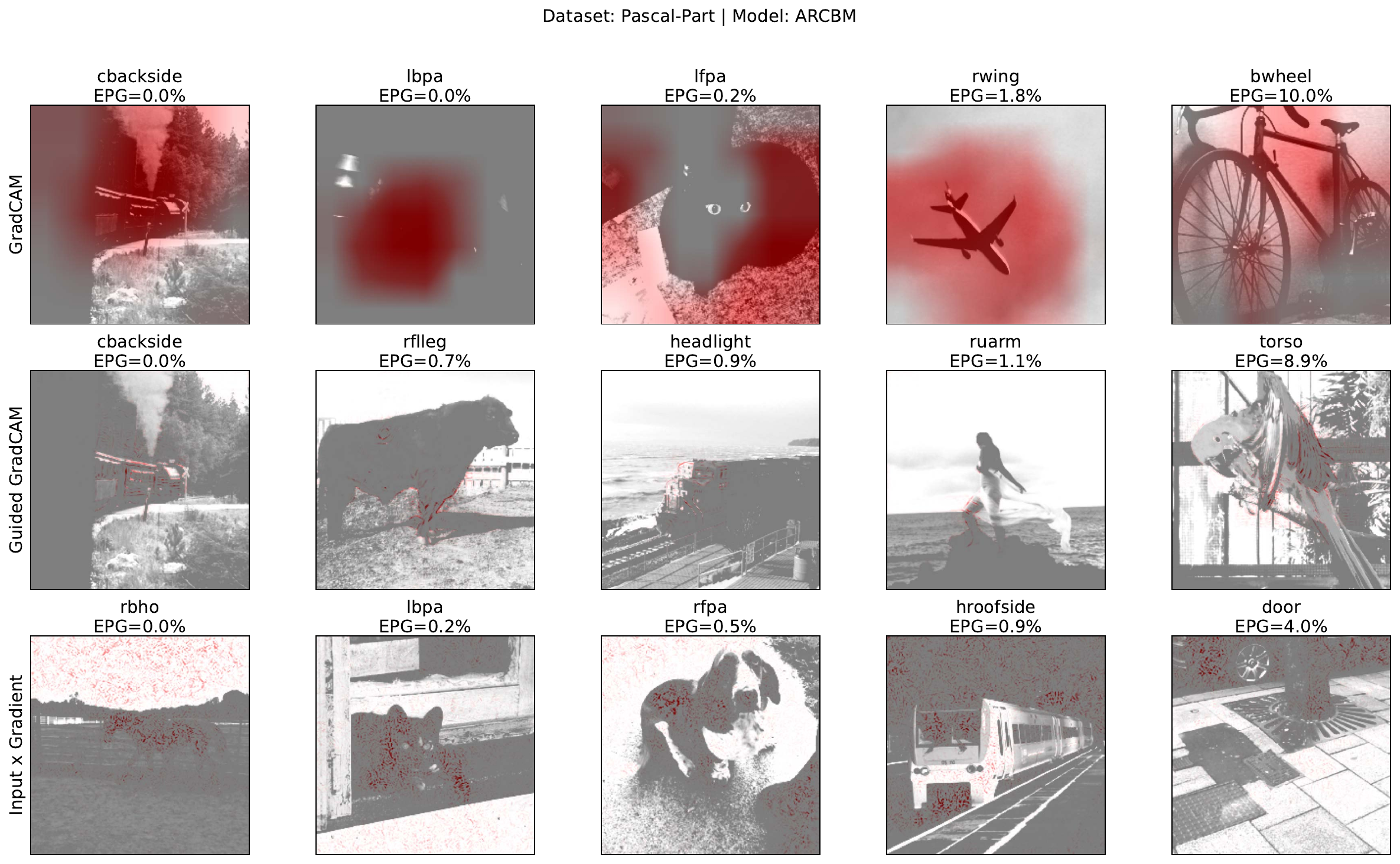}
    \caption{Attribution maps visualization for the Auto-Regressive CBM on the Pascal-Part dataset.}
    \label{app-fig:cbm-pascal_part-attributions}
\end{figure}

\begin{figure}[!htbp]
    \centering
    \captionsetup{font=footnotesize, skip=0.0pt}
    \includegraphics[width=1.0\textwidth]{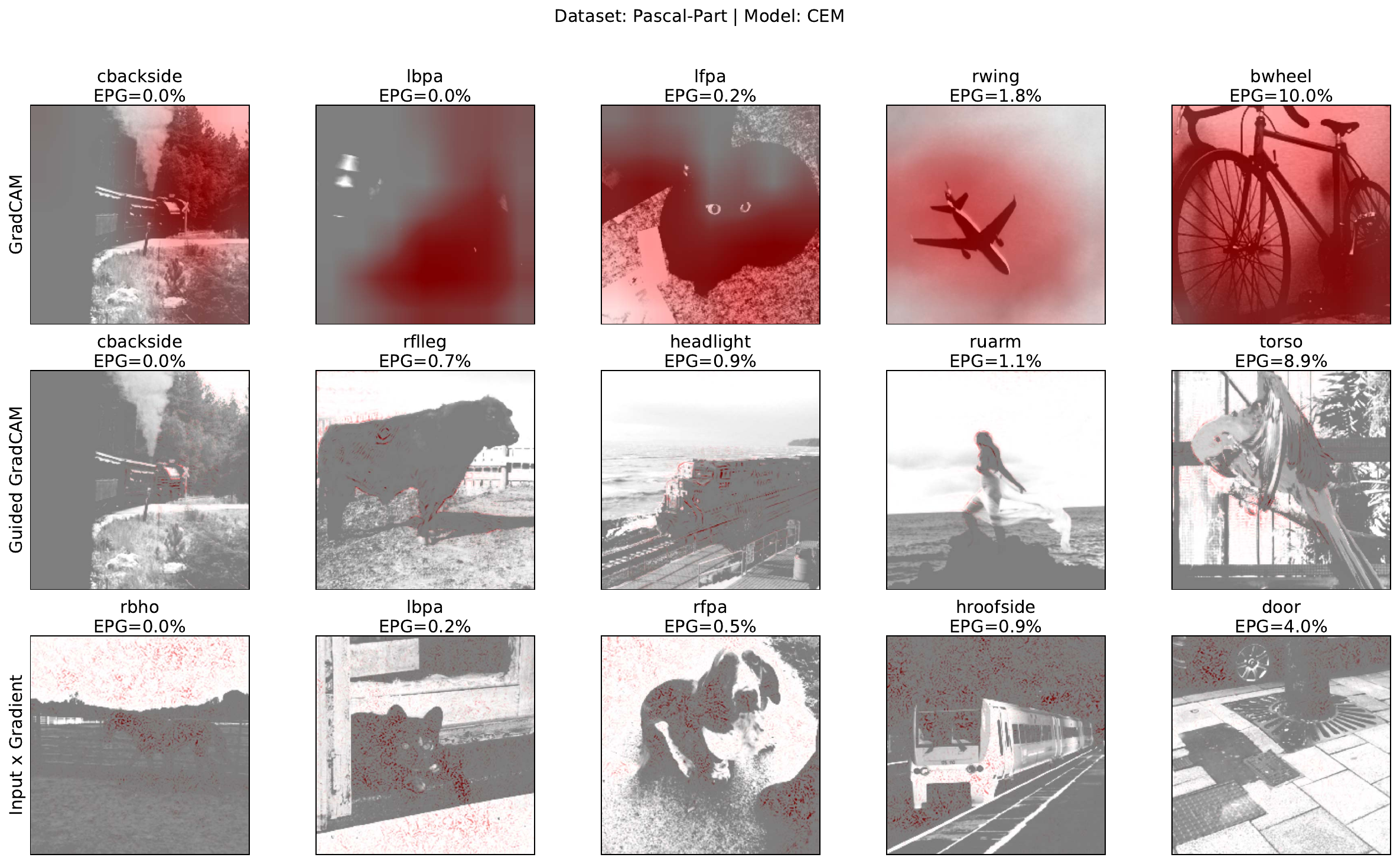}
    \caption{Attribution maps visualization for the CEM on the Pascal-Part dataset.}
    \label{app-fig:cem-pascal_part-attributions}
\end{figure}

\begin{figure}[!htbp]
    \centering
    \captionsetup{font=footnotesize, skip=0.0pt}
    \includegraphics[width=1.0\textwidth]{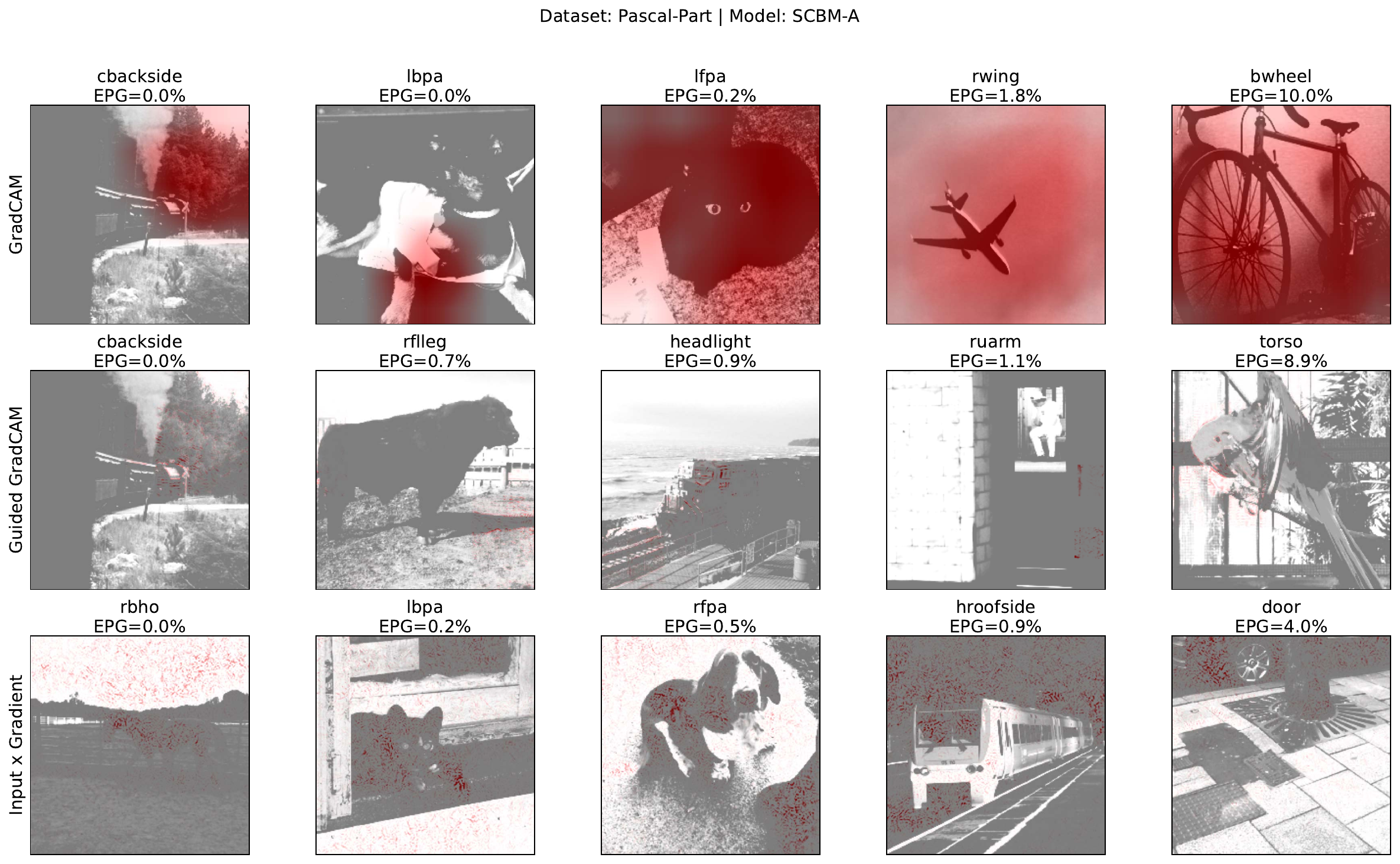}
    \caption{Attribution maps visualization for the SCBM-Amortized on the Pascal-Part dataset.}
    \label{app-fig:scbm_a-pascal_part-attributions}
\end{figure}

\begin{figure}[!htbp]
    \centering
    \captionsetup{font=footnotesize, skip=0.0pt}
    \includegraphics[width=1.0\textwidth]{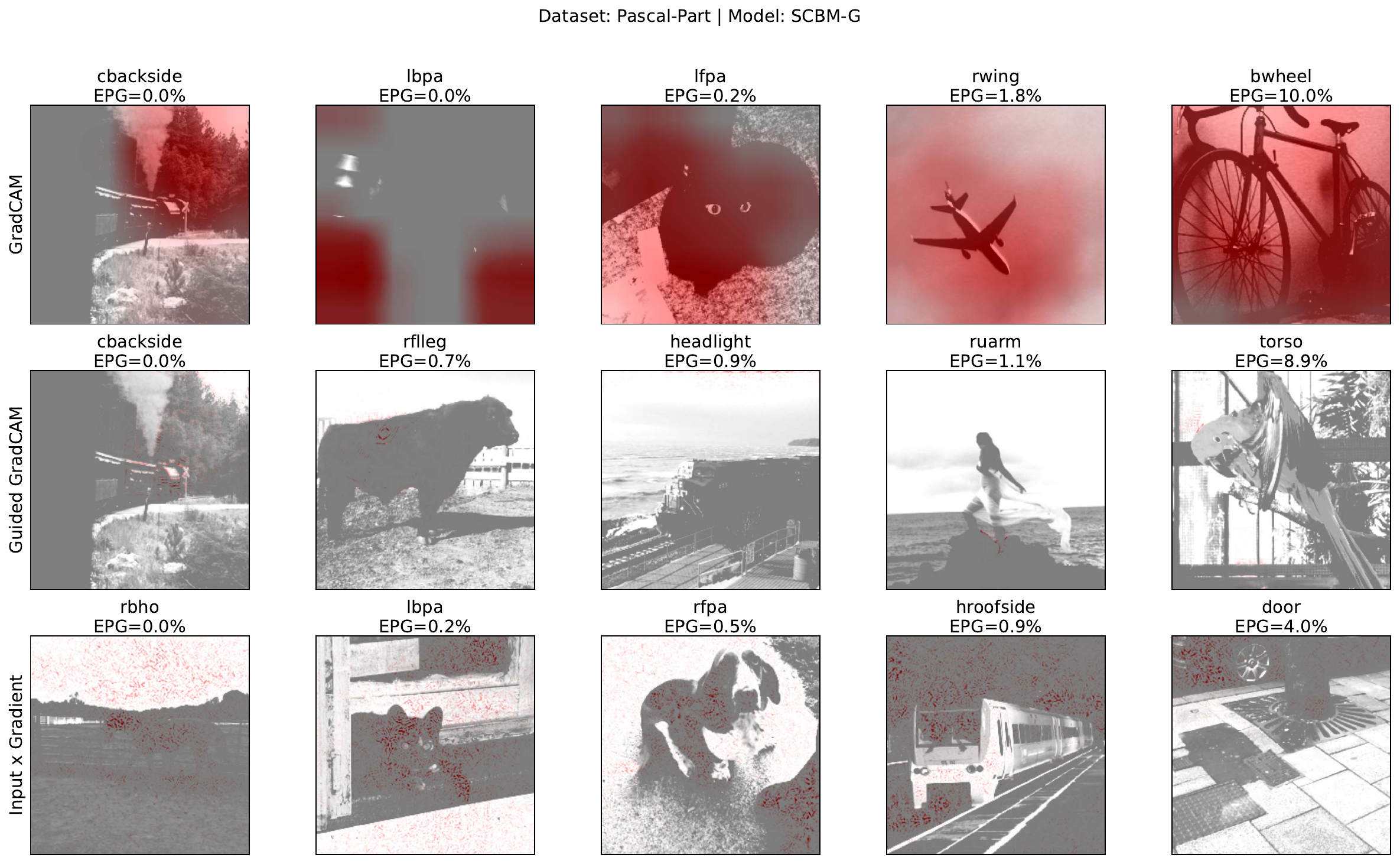}
    \caption{Attribution maps visualization for the SCBM-Global on the Pascal-Part dataset.}
    \label{app-fig:scbm_g-pascal_part-attributions}
\end{figure}
\clearpage
\subsection{Effect of Concept Learning Weight}
\label{appendix:learning_weight}

We ablate the concept-learning weight to quantify its effect on the proposed metrics. For each benchmarked model, we train three random seeds for each of four values, $\alpha \in \{1, 0.1, 0.01, 0.001\}$, and analyze how performance varies with $\alpha$.

\begin{figure}[!htbp]
    \centering
    \captionsetup{font=footnotesize, skip=0.0pt}
    \includegraphics[width=0.8\textwidth]{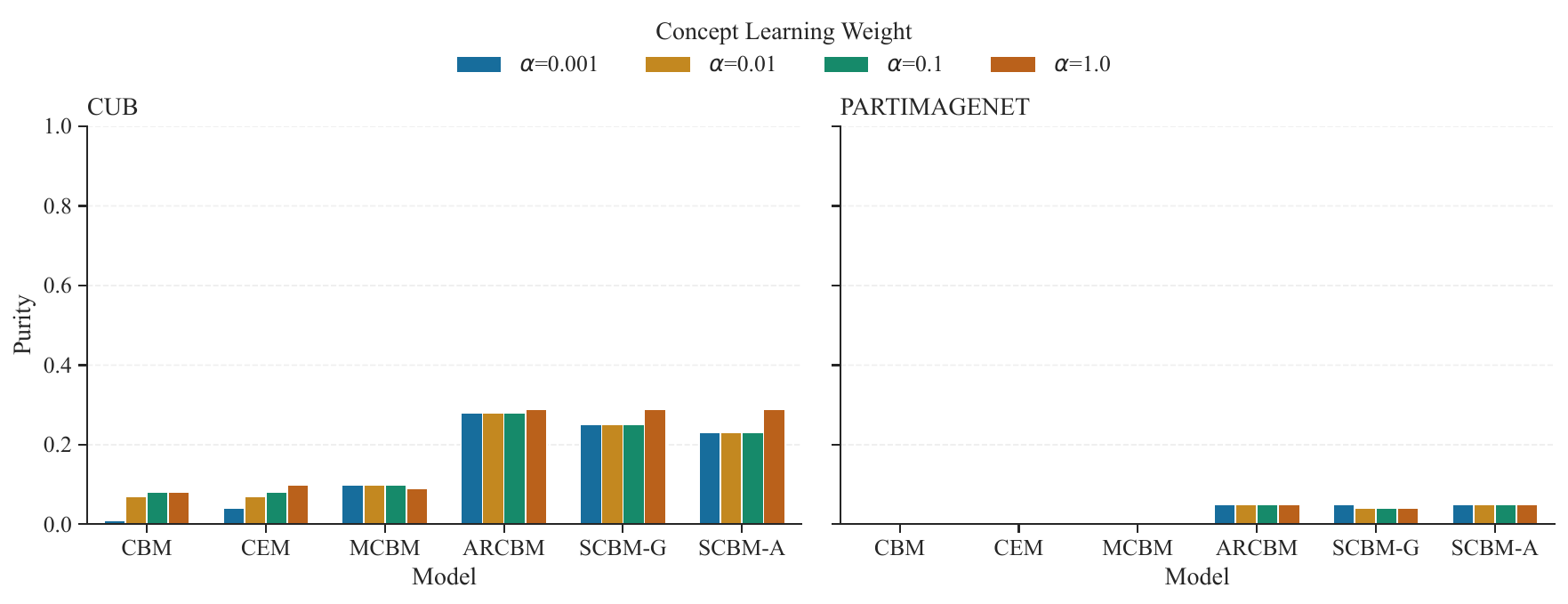}
    \caption{Purity scores of the benchmarked models for different concept-learning weights.}
    \label{app-fig:barplot_disentanglement}
\end{figure}

\noindent\textit{\textbf{Effect on purity.}}
\label{appendix:learning_weight-disentanglement}
As shown in \Cref{app-fig:barplot_disentanglement}, purity improves slightly as $\alpha$ increases. Although the trend is not strictly monotonic, all models attain their highest purity score at $\alpha = 1.0$.

\begin{figure}[!htbp]
    \centering
    \captionsetup{font=footnotesize, skip=0.0pt}
    \includegraphics[width=0.8\textwidth]{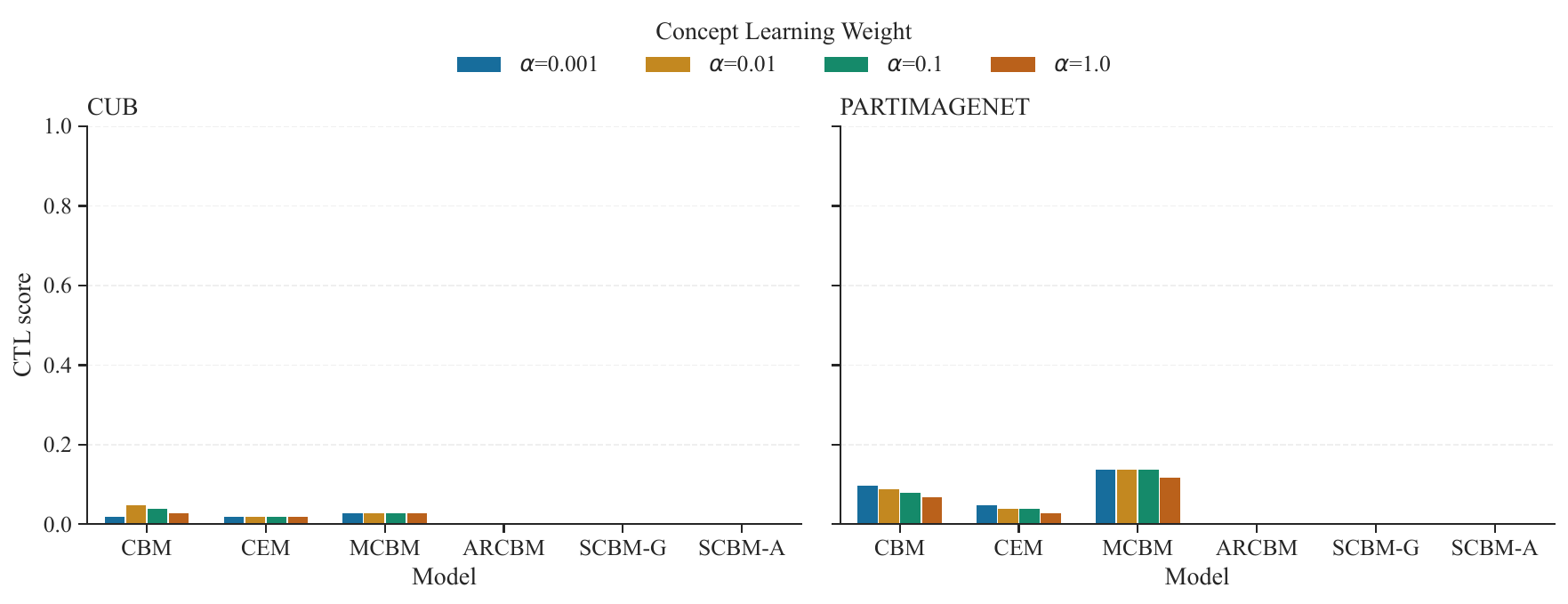}
    \caption{Leakage, measured with CTL~\citep{leakage}, for the benchmarked models under different concept-learning weights.}
    \label{app-fig:barplot_leakage_ctl}
\end{figure}

\noindent\textit{\textbf{Effect on leakage.}}
\label{appendix:learning_weight-leakage}
\Cref{app-fig:barplot_leakage_ctl} shows that the hard CBMs exhibit no leakage. On CUB, the soft models yield similar leakage scores across all values of $\alpha$, indicating limited sensitivity to the concept-learning weight. On PartImageNet, in contrast, leakage decreases progressively as $\alpha$ increases, with the lowest leakage observed at $\alpha = 1.0$. Despite PartImageNet containing only 40 concepts shared across 151 classes, stronger concept supervision still leads to reduced leakage.

\begin{figure}[!htbp]
    \centering
    \captionsetup{font=footnotesize, skip=0.0pt}
    \includegraphics[width=1.0\textwidth]{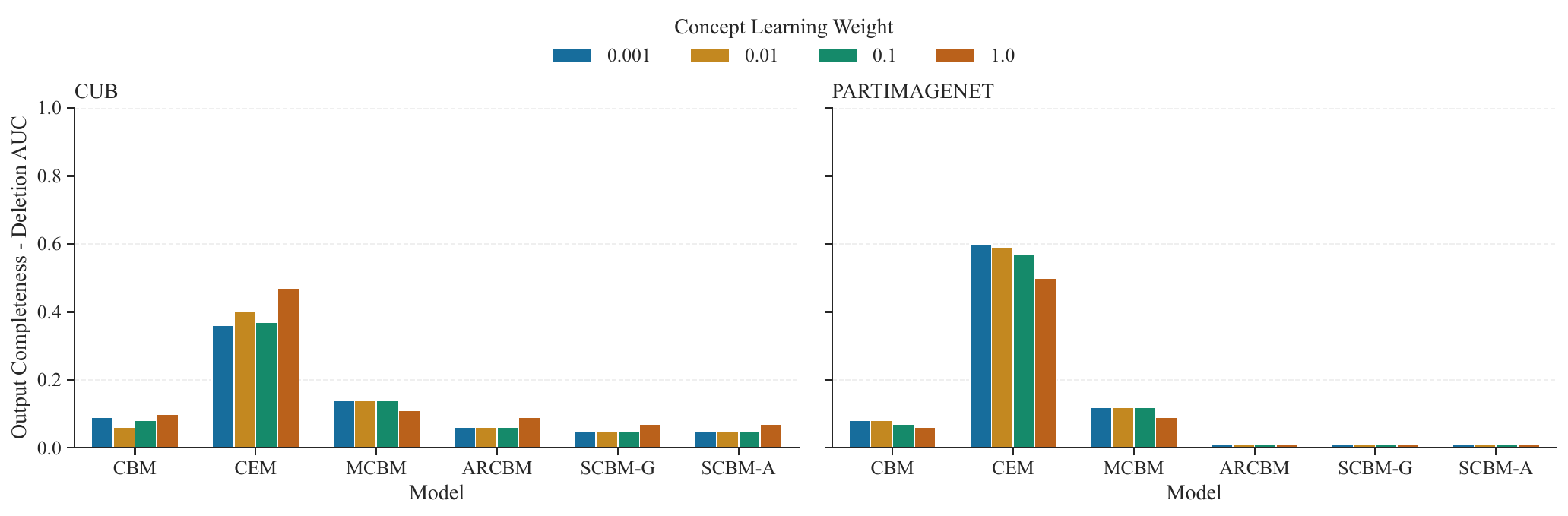}
    \caption{Output completeness: deletion AUC for the benchmarked models under different concept-learning weights (lower is better).}
    \label{app-fig:barplot_output_completeness_del}
\end{figure}

\noindent\textit{\textbf{Effect on output completeness.}}
\label{appendix:learning_weight-output_completeness}
\Cref{app-fig:barplot_output_completeness_del} reports deletion AUC for the output completeness benchmark, where lower values are better. This metric reflects how quickly predictive performance deteriorates when the most useful concepts for each class are removed. Overall, the results suggest that predictions rely primarily on class-relevant concepts. Across both CUB and PartImageNet, the binary models achieve the best deletion AUC, whereas CEM performs worst, indicating greater reliance on a broader set of concepts beyond those associated with the target class.

On CUB, higher values of $\alpha$ generally lead to lower deletion AUC, with MCBM as the main exception. A similar trend is observed on PartImageNet. We note, however, that the hard CBMs (ARCBM, SCBM-G, and SCBM-A) perform poorly in terms of task accuracy on PartImageNet.

\begin{figure}[!htbp]
    \centering
    \captionsetup{font=footnotesize, skip=0.0pt}
    \includegraphics[width=1.0\textwidth]{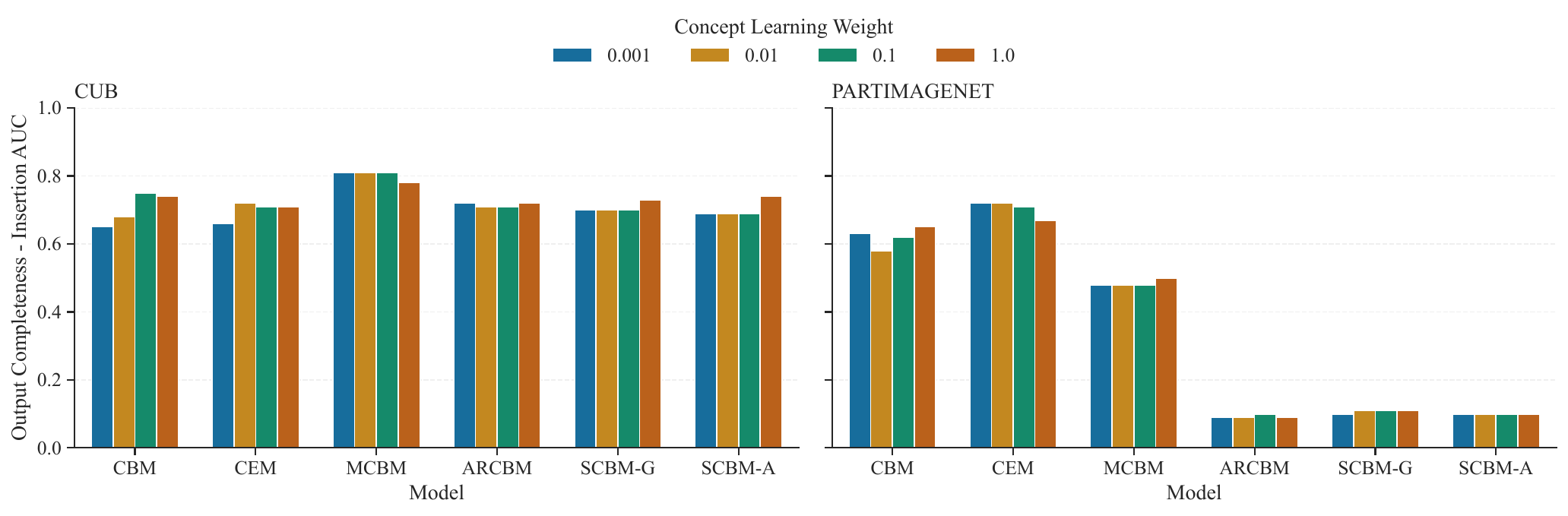}
    \caption{Output completeness: insertion AUC for the benchmarked models under different concept-learning weights (higher is better).}
    \label{app-fig:barplot_output_completeness_ins}
\end{figure}

\Cref{app-fig:barplot_output_completeness_ins} reports the complementary insertion AUC, where concepts are introduced incrementally in decreasing order of utility and higher values are better. Most models achieve high insertion AUC overall, with a mild tendency toward improved performance at larger $\alpha$.

\begin{figure}[!htbp]
    \centering
    \captionsetup{font=footnotesize, skip=0.0pt}
    \includegraphics[width=1.0\textwidth]{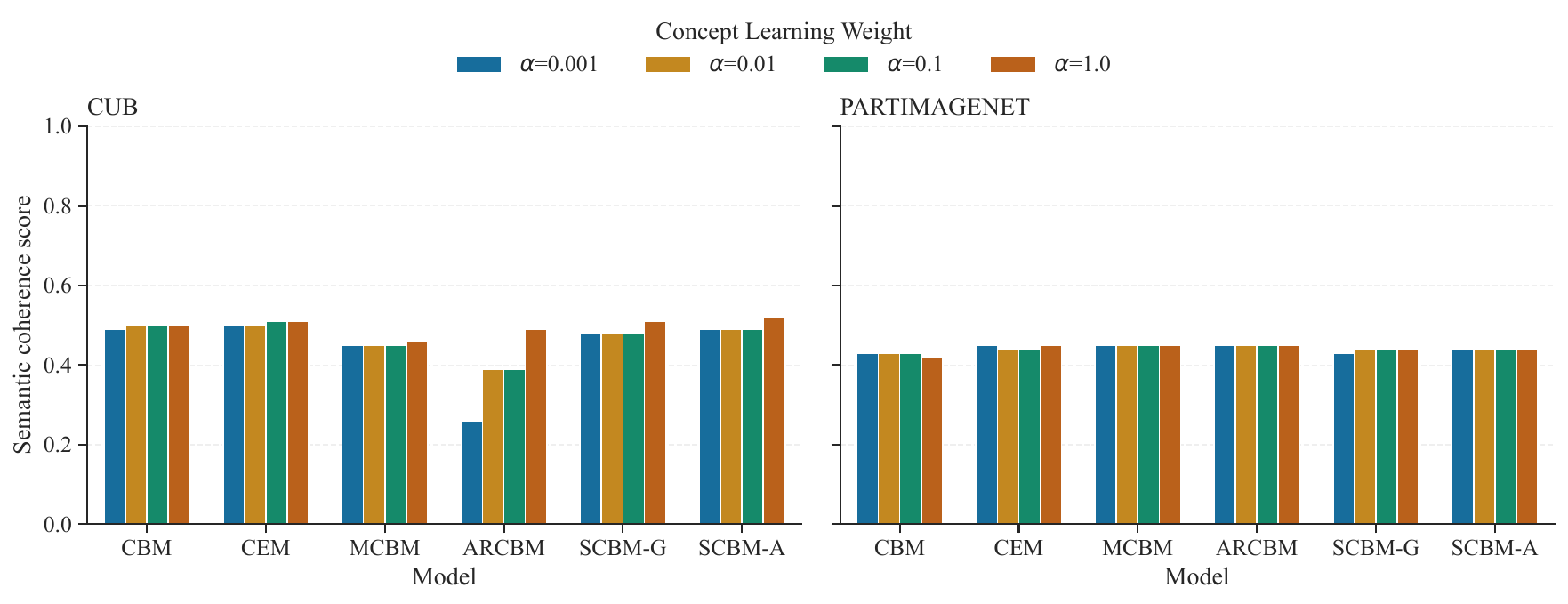}
    \caption{Semantic coherence scores of the benchmarked models for different concept-learning weights.}
    \label{app-fig:barplot_semantic_coherence}
\end{figure}

\noindent\textit{\textbf{Effect on semantic coherence.}}
\label{appendix:learning_weight-semantic_coherence}
As shown in \Cref{app-fig:barplot_semantic_coherence}, semantic coherence is largely insensitive to $\alpha$. The main exception is the hard CBMs on CUB, for which the best scores are obtained at larger values of $\alpha$.

\begin{figure}[!htbp]
    \centering
    \captionsetup{font=footnotesize, skip=0.0pt}

    \begin{subfigure}[t]{0.49\textwidth}
        \centering
        \includegraphics[width=\textwidth]{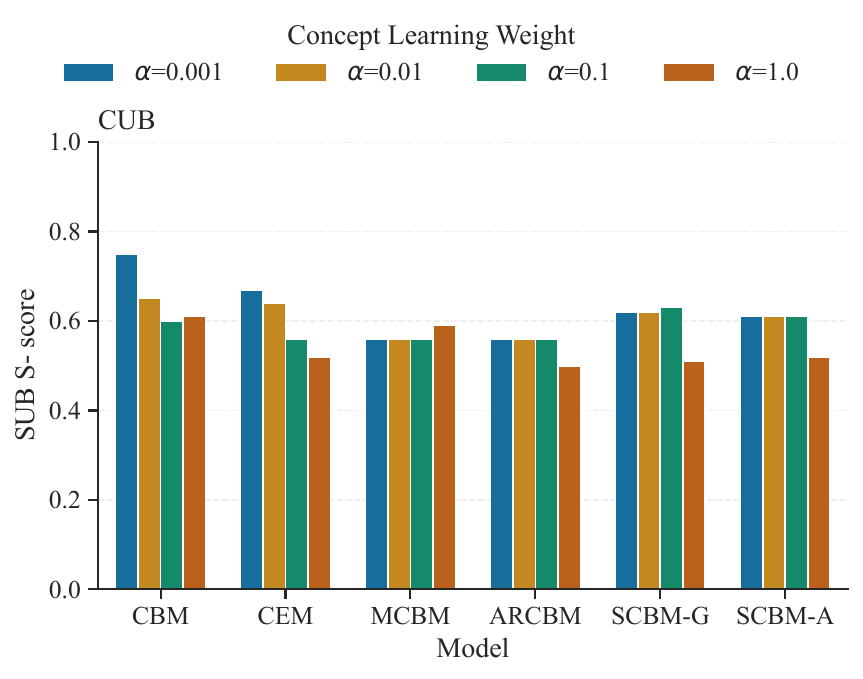}
        \caption{SUB $\mathcal{S}^{-}$ scores for different concept-learning weights.}
        \label{app-fig:sub-s-minus}
    \end{subfigure}
    \hfill
    \begin{subfigure}[t]{0.49\textwidth}
        \centering
        \includegraphics[width=\textwidth]{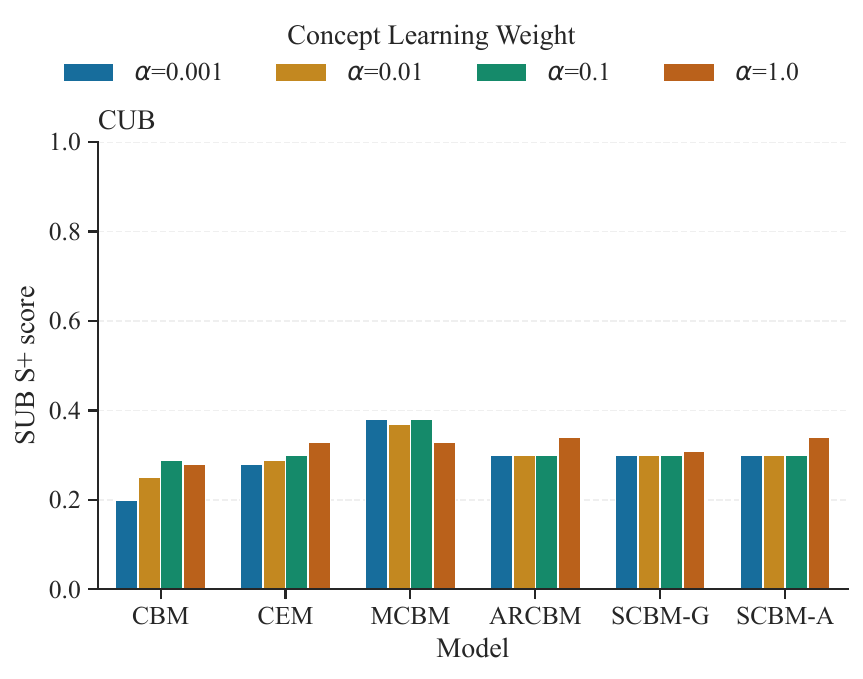}
        \caption{SUB $\mathcal{S}^{+}$ scores for different concept-learning weights.}
        \label{app-fig:sub-s-plus}
    \end{subfigure}

    \caption{Effect of the concept-learning weight on the SUB benchmark.}
    \label{app-fig:concept_learning_weight-SUB}
\end{figure}

\noindent\textit{\textbf{Effect on the SUB benchmark.}}
\label{appendix:learning_weight-sub}
\Cref{app-fig:concept_learning_weight-SUB} shows how varying the concept-learning weight affects the two SUB metrics, $\mathcal{S}^{+}$ and $\mathcal{S}^{-}$. On CUB, smaller values of $\alpha$ lead to better $\mathcal{S}^{-}$ scores, indicating improved recognition of the substituted concept, increasing $\alpha$ consistently reduces this score. The opposite trend is observed for $\mathcal{S}^{+}$, where larger values of $\alpha$ generally improve the recognition of the newly introduced concept, with MCBM as the main exception.

\clearpage
\subsection{Metrics Pair-Wise Correlations}
\label{appendix:metrics_correlations}

Here we report pair-wise scatter plots of model-level metric scores for hard and soft CBMs on each dataset. Each panel compares
two metrics and reports the Pearson correlation coefficient, allowing us to inspect linear associations and potential trade-offs
across models.

The metrics included are task accuracy, concept accuracy, leakage in CTL, purity, EPG, and output completeness. For CUB-trained models, we additionally include results for  SUB S$-$ and SUB S$+$.
All reported scores are averaged over three seeds. EPG is additionally averaged over three attribution methods.

\begin{figure}[!htbp]
    \centering
    \includegraphics[width=\linewidth]{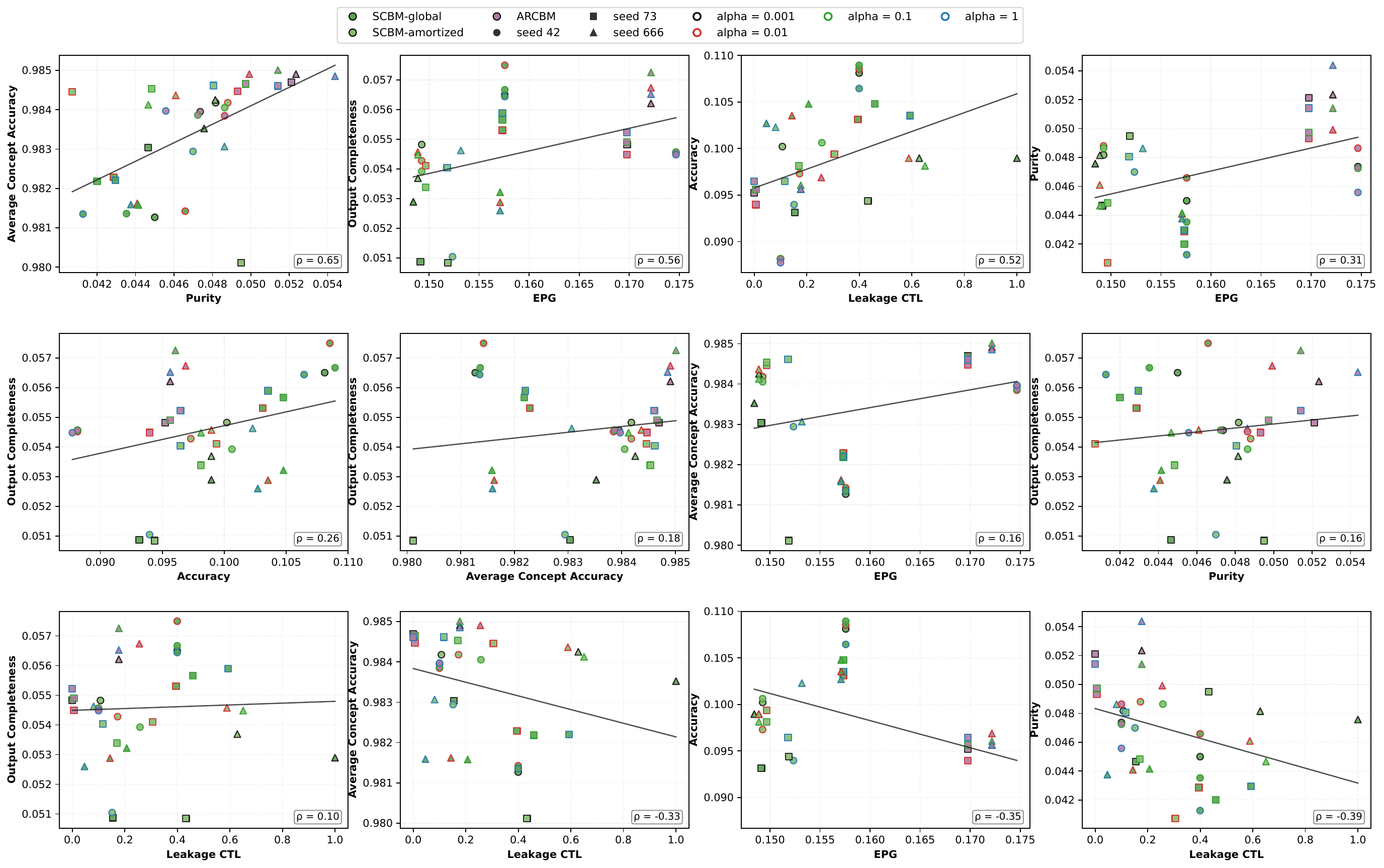}
    \includegraphics[width=\linewidth]{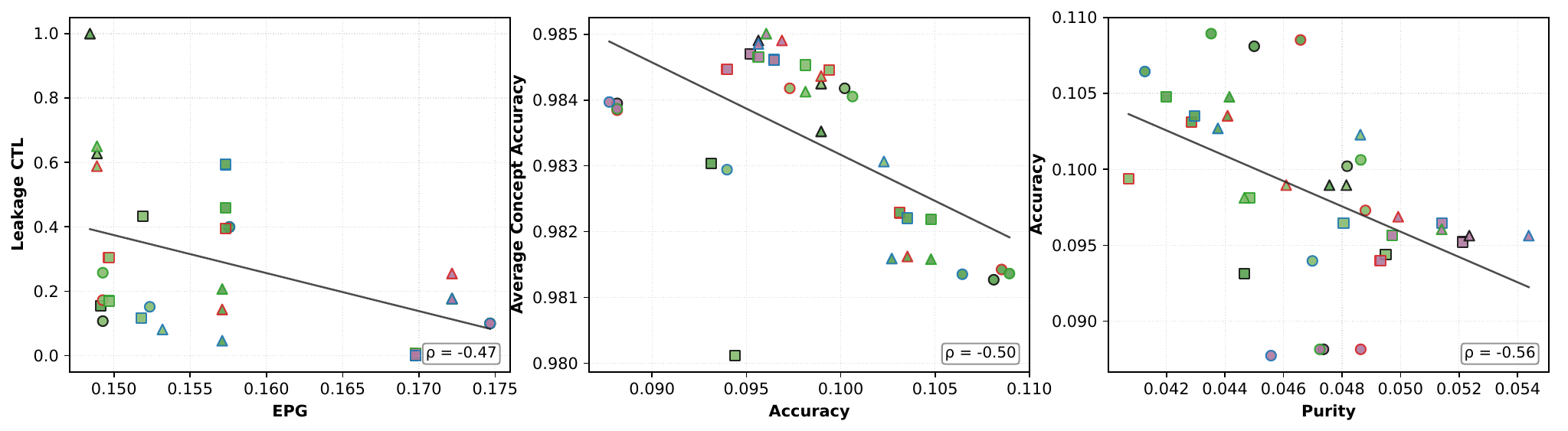}\par\medskip
    \caption{Pair-wise metric correlations for hard CBMs on PartImageNet. The plots compare task accuracy, concept accuracy,
    CTL leakage, purity, EPG, and output completeness; each point corresponds to one model. All scores are averaged
    over three seeds, and EPG is additionally averaged over three attribution methods.}
    \label{app-fig:correlations_partimagenet_hard}
\end{figure}

\begin{figure}[!htbp]
    \centering
    \includegraphics[width=\linewidth]{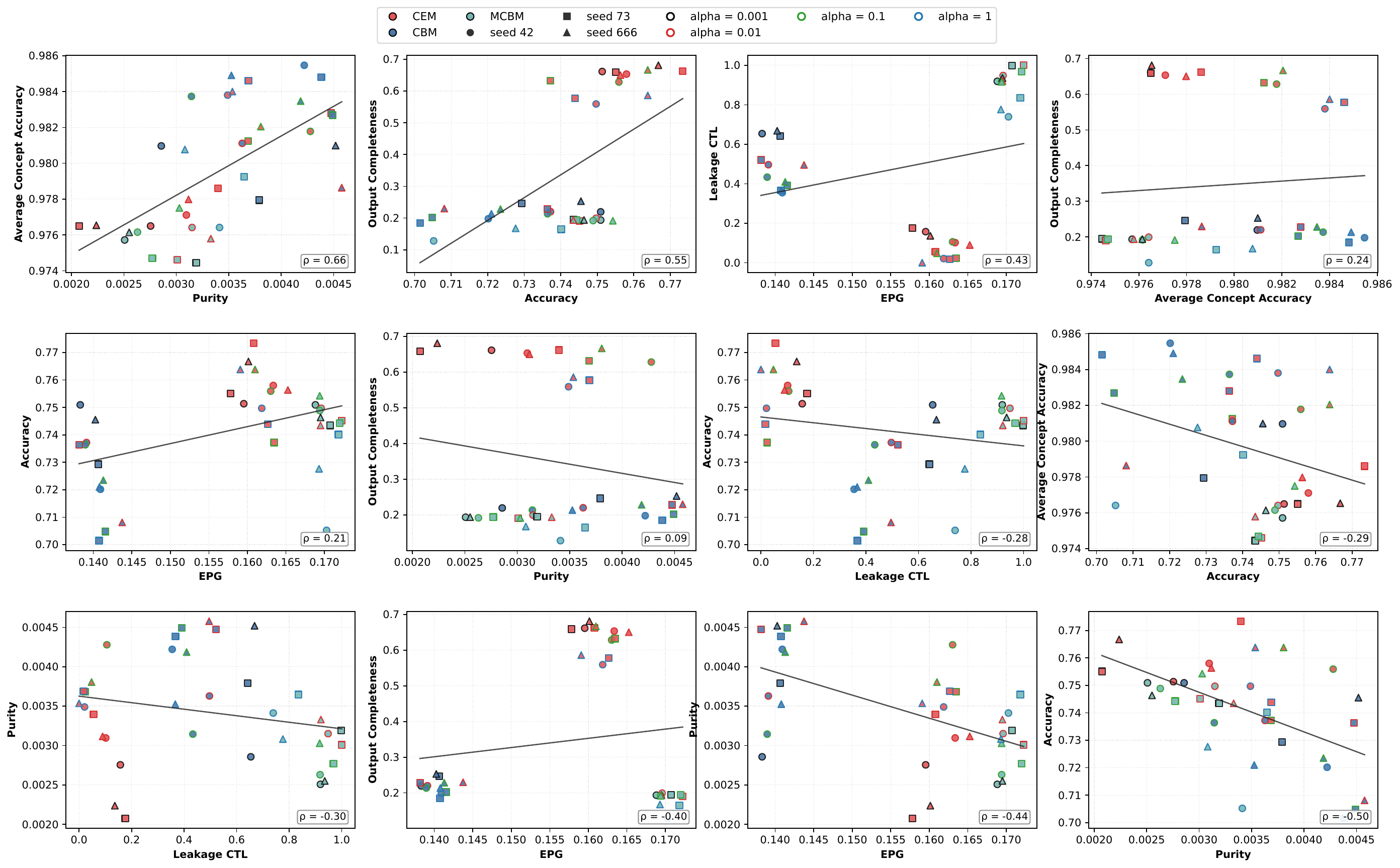}
    \includegraphics[width=\linewidth]{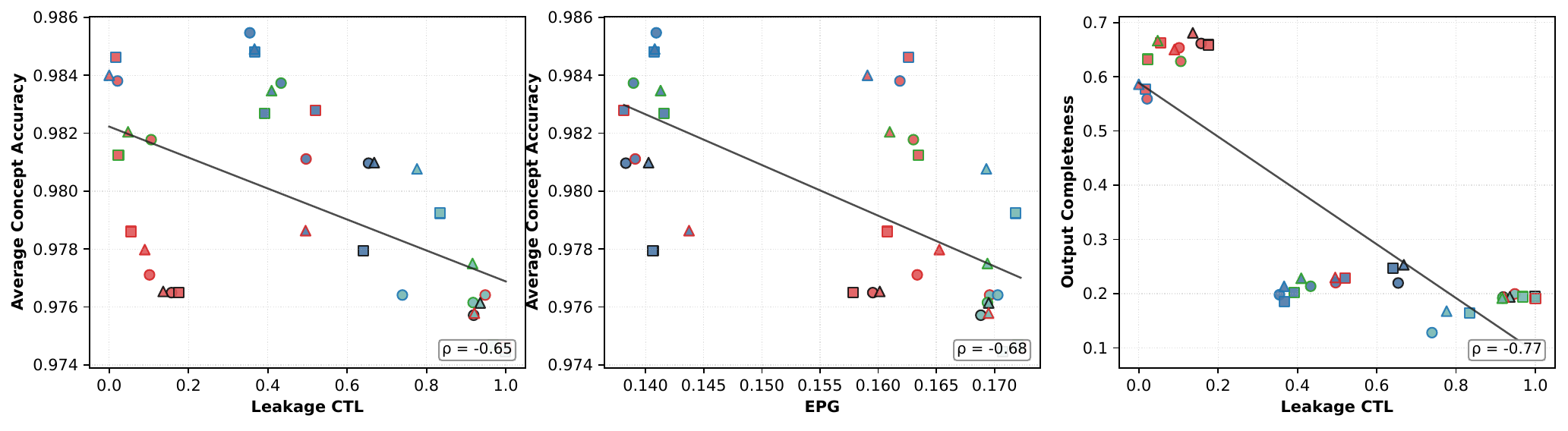}\par\medskip
    \caption{Pair-wise metric correlations for soft CBMs on PartImageNet over the same benchmark metrics (task accuracy,
    concept accuracy, CTL leakage, purity, EPG, and output completeness). All scores are averaged over three seeds,
    and EPG is additionally averaged over three attribution methods.}
    \label{app-fig:correlations_partimagenet_soft}
\end{figure}

\begin{figure}[!htbp]
    \centering
    \includegraphics[width=0.8\linewidth]{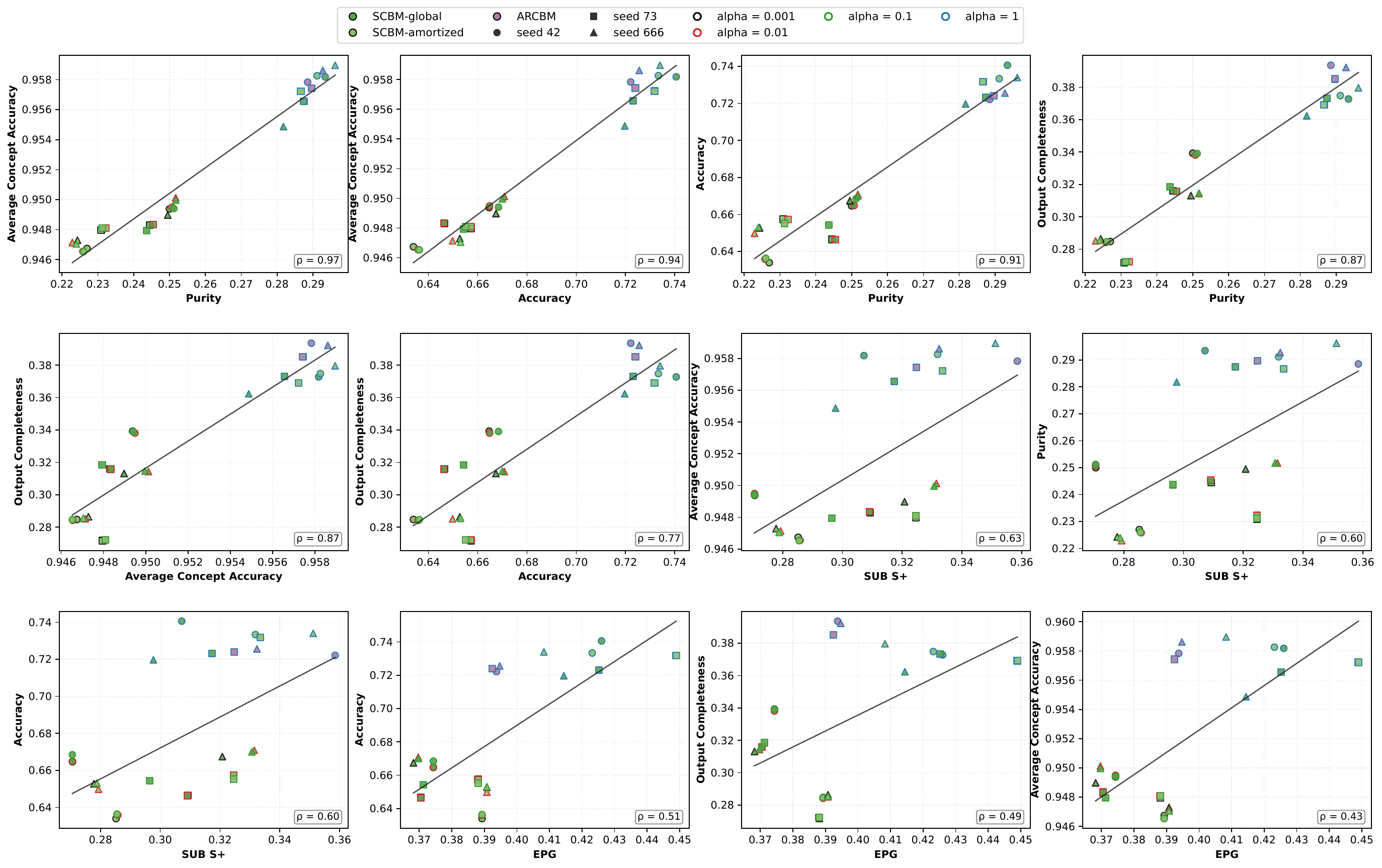}\par\medskip
    \includegraphics[width=0.8\linewidth]{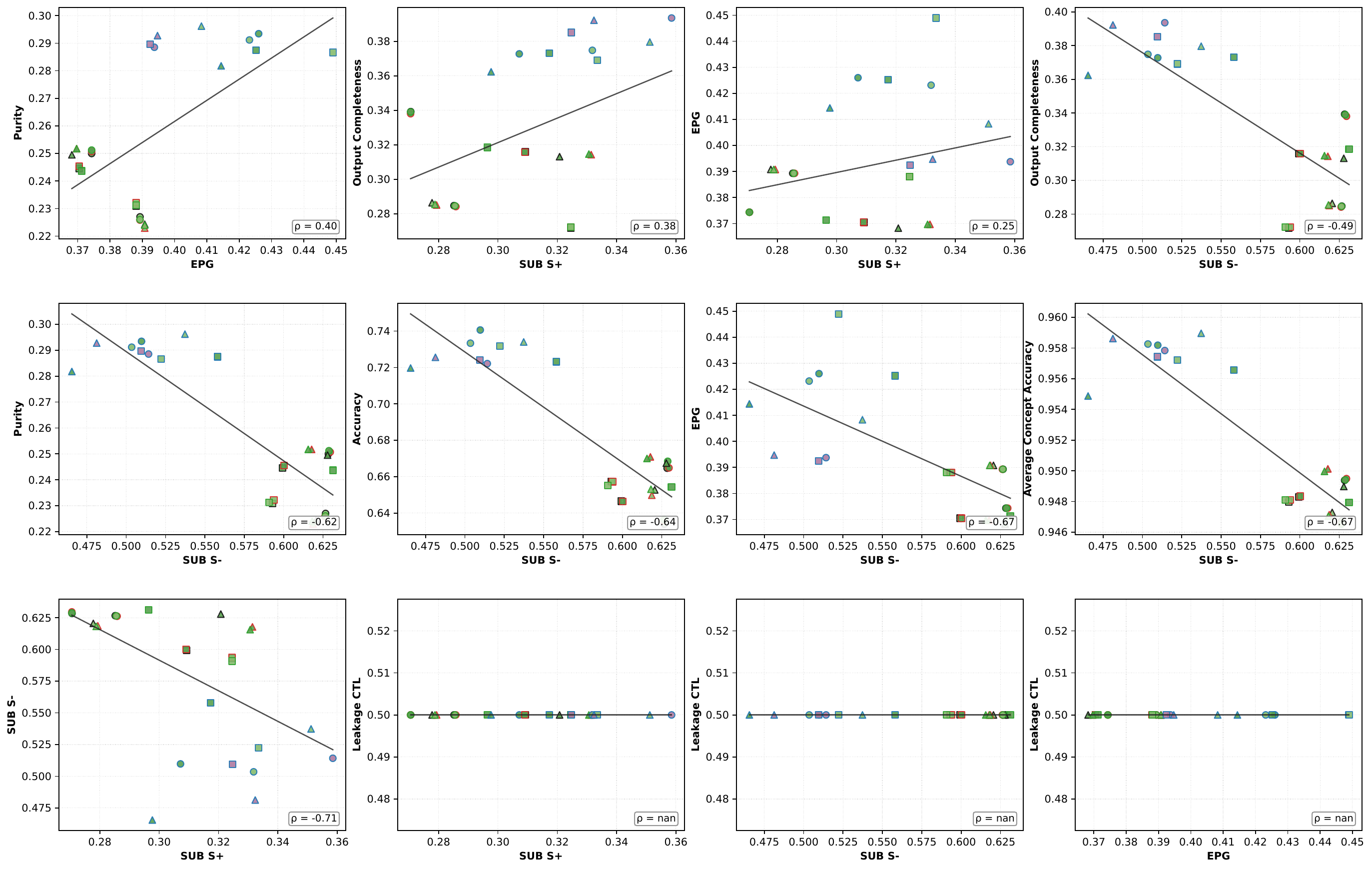}\par\medskip
    \includegraphics[width=0.8\linewidth]{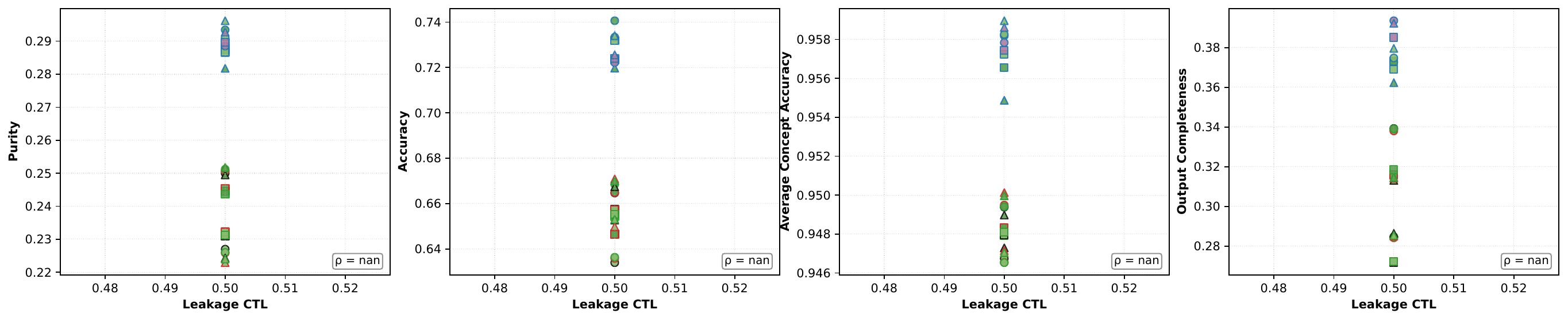}
    \caption{Pair-wise metric correlations for hard CBMs on CUB, including SUB S$-$ and SUB S$+$ scores in addition to task accuracy, concept accuracy, CTL leakage, purity, and EPG. All scores are averaged over three seeds, and EPG is additionally averaged over three attribution methods.}
    \label{app-fig:correlations_cub_hard}
\end{figure}

\begin{figure}[!htbp]
    \centering
    \includegraphics[width=0.8\linewidth]{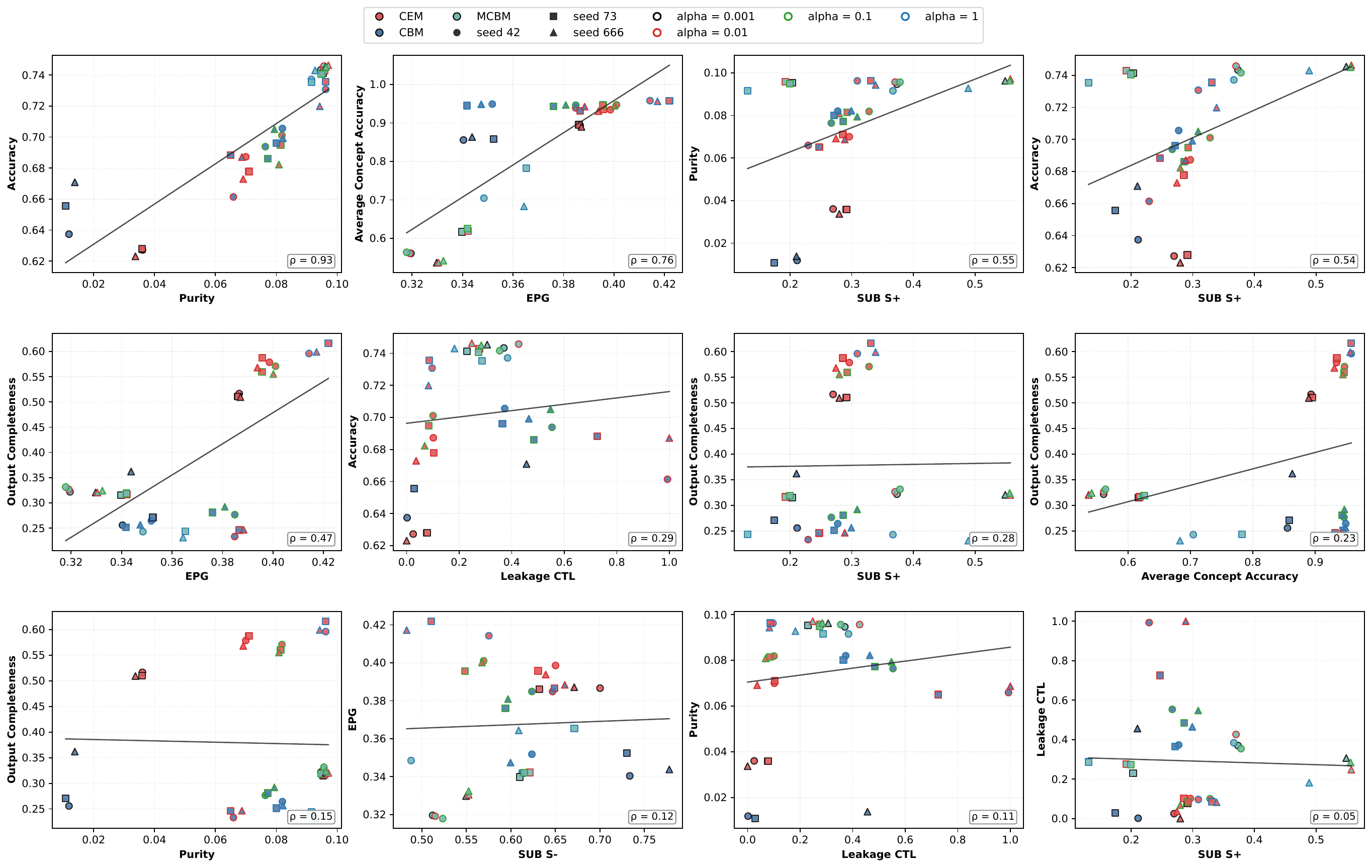}
    \includegraphics[width=0.8\linewidth]{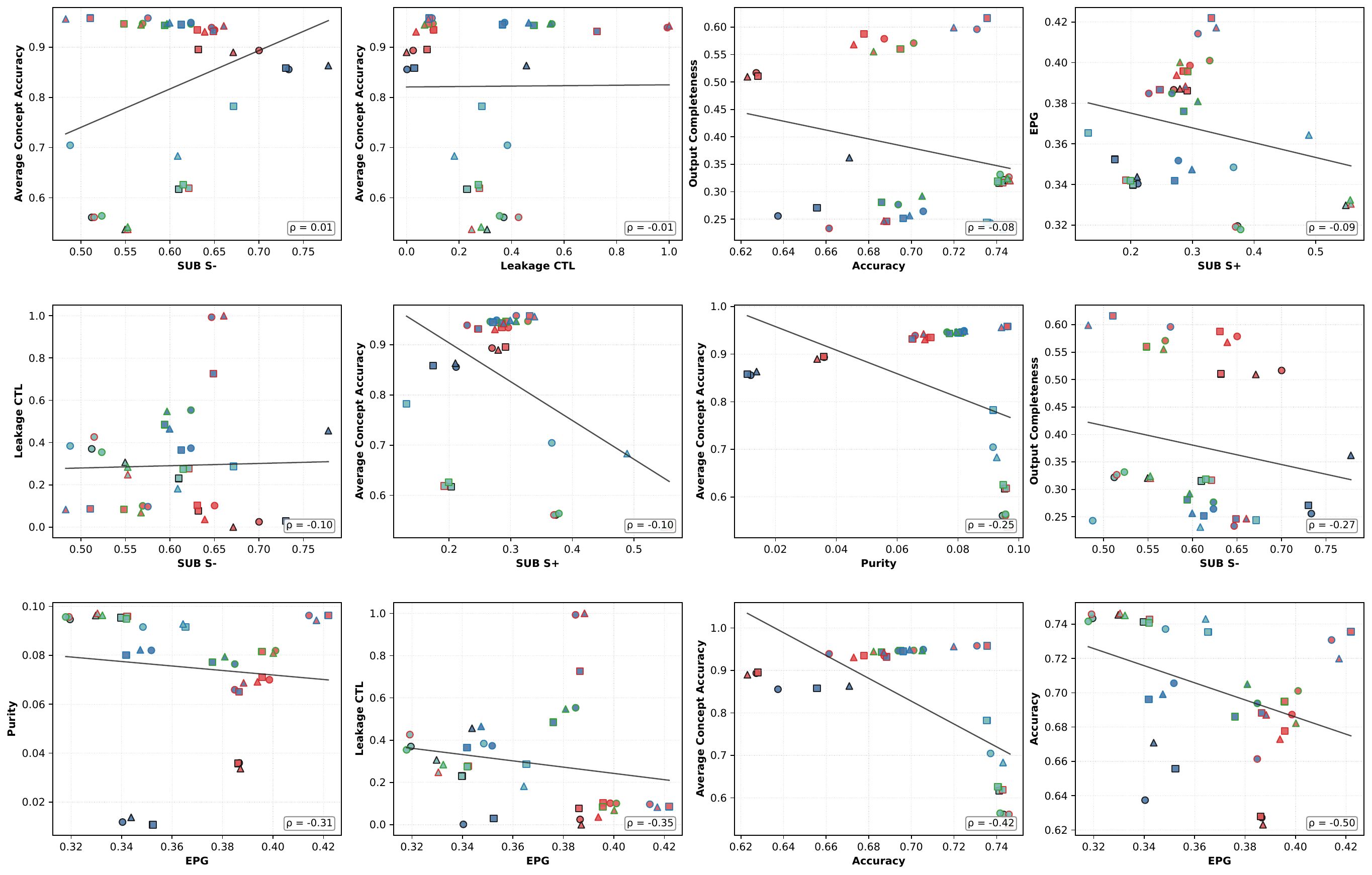}\par\medskip
    \includegraphics[width=0.8\linewidth]{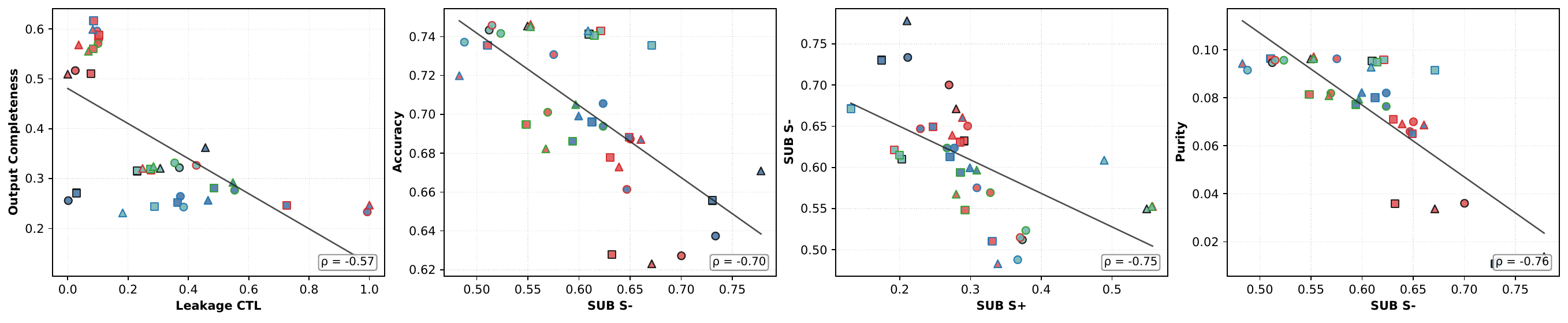}
    \caption{Pair-wise metric correlations for soft CBMs on CUB across task accuracy, concept accuracy, CTL leakage,
    purity, EPG, and output completeness (including SUB S$-$ and SUB S$+$ variants). All scores are averaged over
    three seeds, and EPG is additionally averaged over three attribution methods.}
    \label{app-fig:correlations_cub_soft}
\end{figure}

\begin{figure}[!htbp]
    \centering
    \includegraphics[width=\linewidth]{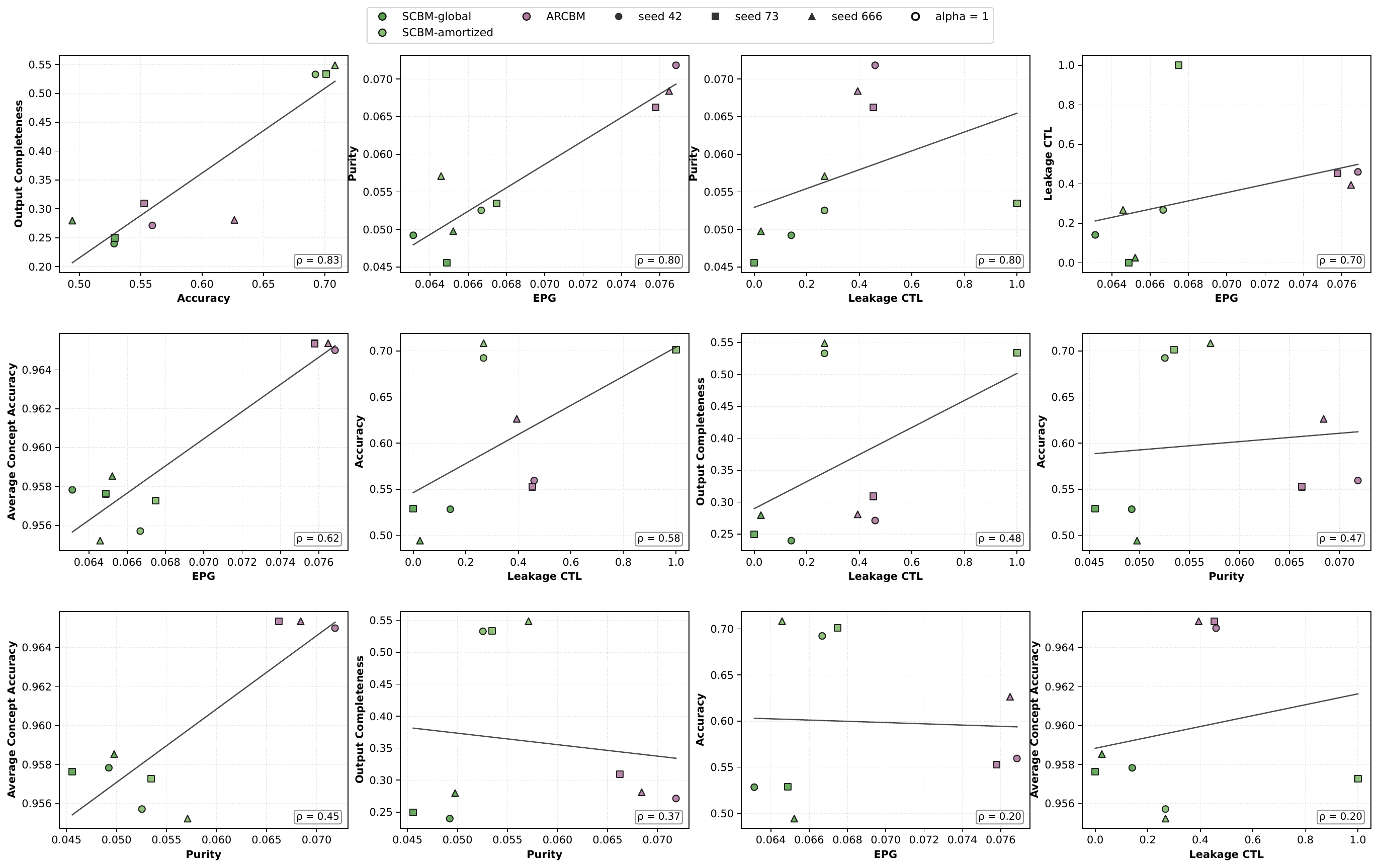}
    \includegraphics[width=\linewidth]{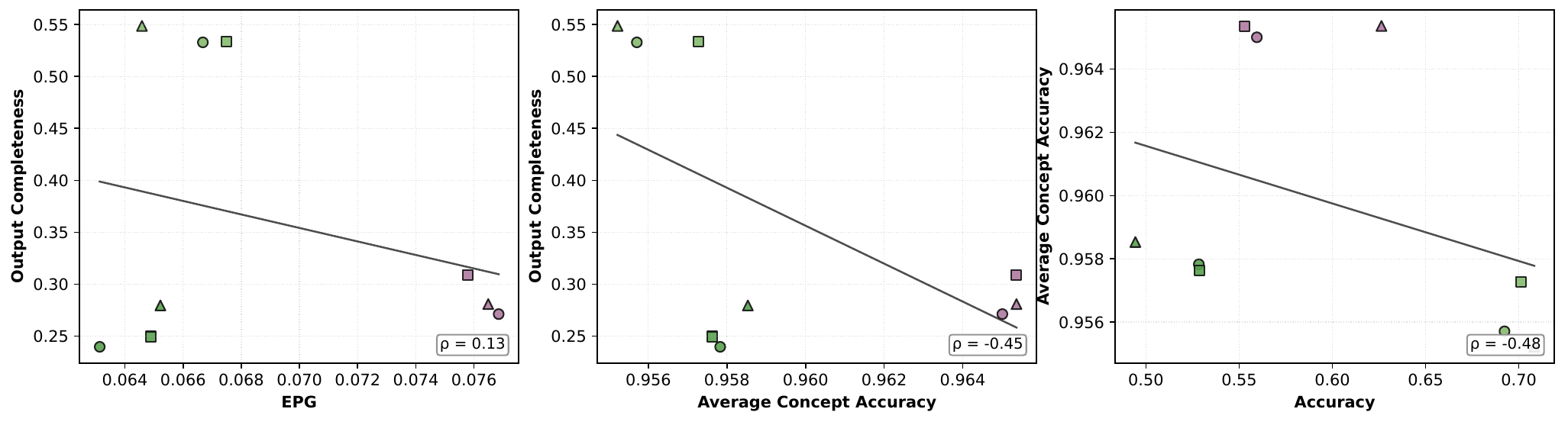}\par\medskip
    \caption{Pair-wise metric correlations for hard CBMs on Pascal-Part, showing interactions among task accuracy,
    concept accuracy, CTL leakage, purity, EPG, and output completeness. All scores are averaged over three seeds,
    and EPG is additionally averaged over three attribution methods.}
    \label{app-fig:correlations_pascalpart_hard}
\end{figure}

\begin{figure}[!htbp]
    \centering
    \includegraphics[width=\linewidth]{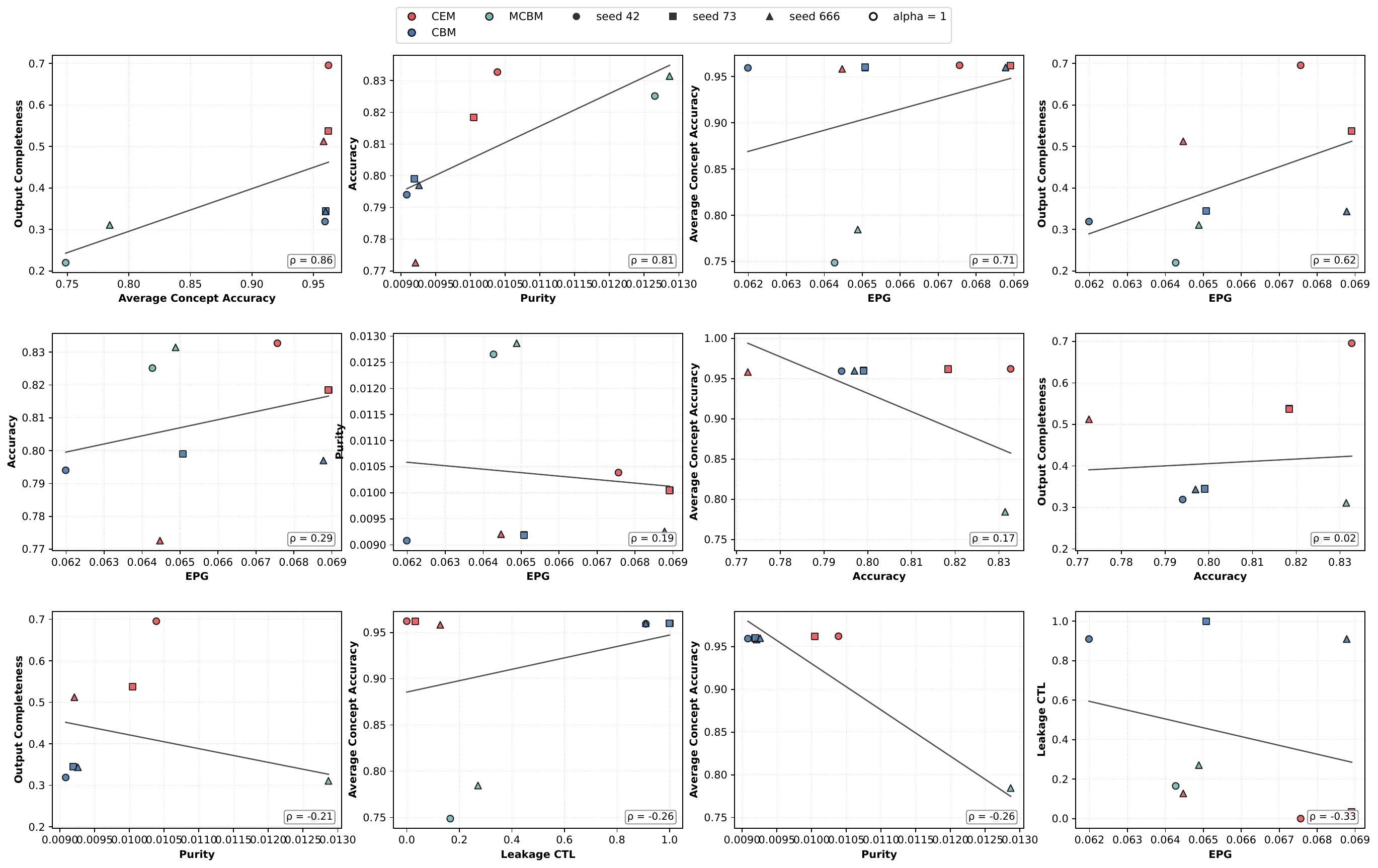}
    \includegraphics[width=\linewidth]{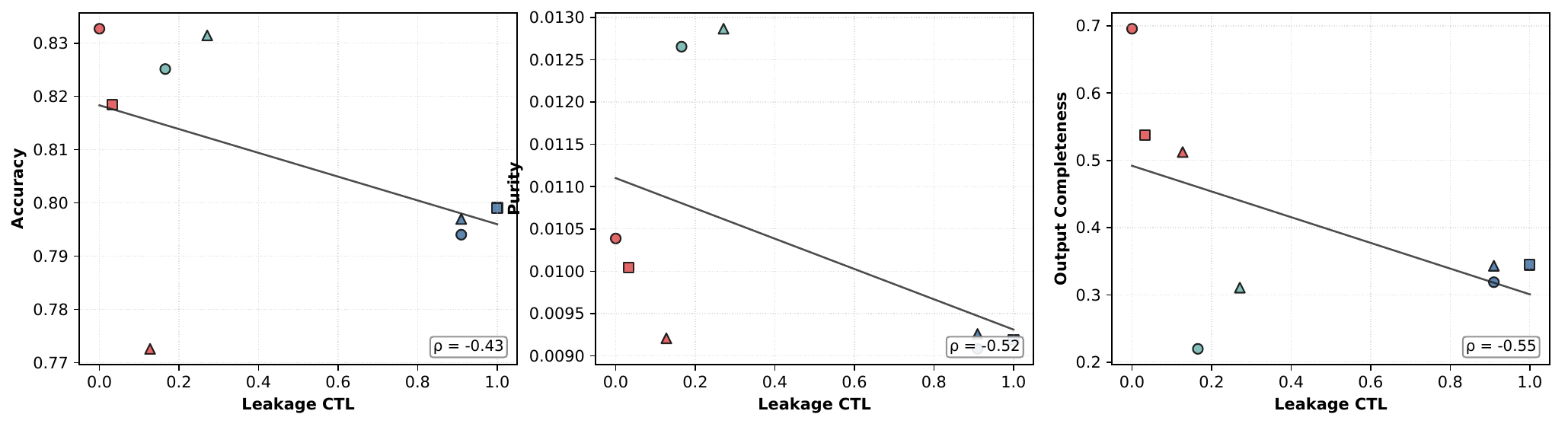}\par\medskip
    \caption{Pair-wise metric correlations for soft CBMs on Pascal-Part over the full benchmark metric set (task accuracy,
    concept accuracy, CTL leakage, purity, EPG, and output completeness). All scores are averaged over three seeds,
    and EPG is additionally averaged over three attribution methods.}
    \label{app-fig:correlations_pascalpart_soft}
\end{figure}
\clearpage
\subsection{Concept Statistics}
\label{appendix:concept_stats}

Here we summarize annotation mask-size statistics on the training split of each dataset. For each concept (or class in CUB), we report the mean and standard deviation of the mask area as a percentage of the full image area, together with the number of images in which that concept/class is present.

\begingroup
\setlength{\LTleft}{0pt}
\setlength{\LTright}{0pt}
\setlength{\LTcapwidth}{\linewidth}
\setlength{\tabcolsep}{3pt}
\begin{longtable}{@{}>{\raggedright\arraybackslash}p{0.12\linewidth}>{\raggedright\arraybackslash}p{0.34\linewidth}>{\centering\arraybackslash}p{0.15\linewidth}>{\centering\arraybackslash}p{0.15\linewidth}>{\centering\arraybackslash}p{0.12\linewidth}@{}}
\caption{PascalPart training split: per-concept part-mask area statistics (percentage of image area).}\label{tab:pascalpart-concept-sizes-train}\\
\toprule
\textbf{Concept ID} & \textbf{Concept Name} & \textbf{Mean Mask Ratio (\%)} & \textbf{Std. Dev. (\%)} & \textbf{Images Present} \\
\midrule
\endfirsthead
\toprule
\textbf{Concept ID} & \textbf{Concept Name} & \textbf{Mean Mask Ratio (\%)} & \textbf{Std. Dev. (\%)} & \textbf{Images Present} \\
\midrule
\endhead
\midrule \multicolumn{5}{r}{\emph{Continued on next page}}\\
\endfoot
\bottomrule
\endlastfoot
0 & backside & 15.42 & 23.49 & 53 \\
1 & beak & 0.71 & 1.82 & 233 \\
2 & bliplate & 0.87 & 1.69 & 30 \\
3 & body & 17.40 & 18.36 & 258 \\
4 & bwheel & 11.66 & 15.07 & 118 \\
5 & cap & 1.12 & 2.25 & 43 \\
6 & cbackside & 18.26 & 17.60 & 4 \\
7 & cfrontside & 1.22 & 1.97 & 18 \\
8 & chainwheel & 3.36 & 9.00 & 58 \\
9 & cleftside & 8.37 & 15.29 & 43 \\
10 & coach & 17.97 & 25.16 & 101 \\
11 & crightside & 4.15 & 6.32 & 26 \\
12 & croofside & 4.71 & 10.28 & 9 \\
13 & door & 9.69 & 10.22 & 93 \\
14 & engine & 2.67 & 2.65 & 141 \\
15 & fliplate & 0.63 & 0.82 & 60 \\
16 & frontside & 27.55 & 27.01 & 124 \\
17 & fwheel & 10.15 & 11.94 & 113 \\
18 & hair & 7.67 & 10.56 & 170 \\
19 & handlebar & 1.83 & 2.72 & 55 \\
20 & hbackside & 4.69 & 4.00 & 6 \\
21 & head & 19.24 & 22.46 & 1437 \\
22 & headlight & 1.31 & 3.16 & 222 \\
23 & hfrontside & 15.66 & 16.62 & 83 \\
24 & hleftside & 17.80 & 14.66 & 55 \\
25 & hrightside & 19.58 & 15.56 & 43 \\
26 & hroofside & 4.03 & 2.72 & 13 \\
27 & lbho & 0.14 & 0.11 & 32 \\
28 & lbleg & 2.06 & 2.88 & 247 \\
29 & lblleg & 0.69 & 0.64 & 76 \\
30 & lbpa & 0.49 & 0.66 & 180 \\
31 & lbuleg & 0.89 & 0.77 & 75 \\
32 & lear & 1.74 & 2.66 & 842 \\
33 & lebrow & 0.09 & 0.18 & 144 \\
34 & leftmirror & 0.25 & 0.31 & 98 \\
35 & leftside & 21.90 & 18.78 & 93 \\
36 & leye & 0.35 & 2.58 & 932 \\
37 & lfho & 0.15 & 0.13 & 27 \\
38 & lfleg & 3.20 & 3.64 & 459 \\
39 & lflleg & 0.58 & 0.61 & 84 \\
40 & lfoot & 0.38 & 0.47 & 136 \\
41 & lfpa & 1.02 & 1.25 & 321 \\
42 & lfuleg & 1.04 & 1.01 & 90 \\
43 & lhand & 1.40 & 2.26 & 113 \\
44 & lhorn & 0.57 & 0.77 & 35 \\
45 & llarm & 1.86 & 2.45 & 125 \\
46 & lleg & 0.52 & 0.60 & 140 \\
47 & llleg & 1.69 & 2.64 & 67 \\
48 & luarm & 2.67 & 2.86 & 158 \\
49 & luleg & 2.57 & 3.16 & 85 \\
50 & lwing & 3.13 & 3.90 & 248 \\
51 & mouth & 0.41 & 0.76 & 171 \\
52 & muzzle & 6.91 & 9.34 & 504 \\
53 & neck & 3.32 & 4.32 & 751 \\
54 & nose & 0.71 & 1.32 & 841 \\
55 & plant & 29.95 & 29.96 & 37 \\
56 & pot & 29.22 & 31.76 & 39 \\
57 & rbho & 0.15 & 0.12 & 29 \\
58 & rbleg & 2.17 & 2.74 & 240 \\
59 & rblleg & 0.63 & 0.62 & 72 \\
60 & rbpa & 0.48 & 0.54 & 175 \\
61 & rbuleg & 0.80 & 0.77 & 75 \\
62 & rear & 1.93 & 2.84 & 844 \\
63 & rebrow & 0.11 & 0.19 & 135 \\
64 & reye & 0.28 & 0.51 & 933 \\
65 & rfho & 0.15 & 0.16 & 28 \\
66 & rfleg & 3.15 & 3.44 & 444 \\
67 & rflleg & 0.52 & 0.55 & 82 \\
68 & rfoot & 0.36 & 0.57 & 138 \\
69 & rfpa & 0.98 & 1.28 & 318 \\
70 & rfuleg & 1.03 & 1.18 & 91 \\
71 & rhand & 1.28 & 1.63 & 109 \\
72 & rhorn & 1.19 & 2.95 & 35 \\
73 & rightmirror & 0.26 & 0.39 & 98 \\
74 & rightside & 25.84 & 21.06 & 79 \\
75 & rlarm & 1.78 & 1.82 & 113 \\
76 & rleg & 0.46 & 0.55 & 142 \\
77 & rlleg & 1.63 & 2.00 & 59 \\
78 & roofside & 2.42 & 3.13 & 27 \\
79 & ruarm & 2.28 & 2.62 & 146 \\
80 & ruleg & 2.56 & 2.73 & 81 \\
81 & rwing & 3.65 & 4.76 & 264 \\
82 & saddle & 2.09 & 5.47 & 55 \\
83 & screen & 23.25 & 21.44 & 51 \\
84 & stern & 3.35 & 3.87 & 188 \\
85 & tail & 1.63 & 2.70 & 583 \\
86 & torso & 16.28 & 12.86 & 1308 \\
87 & wheel & 2.09 & 3.60 & 303 \\
88 & window & 7.98 & 7.26 & 187 \\
\end{longtable}
\endgroup

\begingroup
\setlength{\LTleft}{0pt}
\setlength{\LTright}{0pt}
\setlength{\LTcapwidth}{\linewidth}
\setlength{\tabcolsep}{3pt}
\begin{longtable}{@{}>{\raggedright\arraybackslash}p{0.12\linewidth}>{\raggedright\arraybackslash}p{0.34\linewidth}>{\centering\arraybackslash}p{0.15\linewidth}>{\centering\arraybackslash}p{0.15\linewidth}>{\centering\arraybackslash}p{0.12\linewidth}@{}}
\caption{PartImageNet training split: per-concept part-mask area statistics (percentage of image area).}\label{tab:partimagenet-concept-sizes-train}\\
\toprule
\textbf{Concept ID} & \textbf{Concept Name} & \textbf{Mean Mask Ratio (\%)} & \textbf{Std. Dev. (\%)} & \textbf{Images Present} \\
\midrule
\endfirsthead
\toprule
\textbf{Concept ID} & \textbf{Concept Name} & \textbf{Mean Mask Ratio (\%)} & \textbf{Std. Dev. (\%)} & \textbf{Images Present} \\
\midrule
\endhead
\midrule \multicolumn{5}{r}{\emph{Continued on next page}}\\
\endfoot
\bottomrule
\endlastfoot
0 & Quadruped Head & 14.61 & 16.34 & 6059 \\
1 & Quadruped Body & 12.57 & 8.72 & 5784 \\
2 & Quadruped Foot & 5.85 & 4.87 & 4645 \\
3 & Quadruped Tail & 2.01 & 3.35 & 2202 \\
4 & Biped Head & 11.96 & 16.55 & 2132 \\
5 & Biped Body & 10.57 & 10.11 & 2043 \\
6 & Biped Hand & 7.68 & 7.39 & 1797 \\
7 & Biped Foot & 4.44 & 3.95 & 1300 \\
8 & Biped Tail & 2.01 & 2.00 & 656 \\
9 & Fish Head & 6.81 & 10.81 & 1162 \\
10 & Fish Body & 8.27 & 6.74 & 1126 \\
11 & Fish Fin & 2.80 & 2.74 & 1100 \\
12 & Fish Tail & 2.04 & 2.46 & 822 \\
13 & Bird Head & 3.88 & 7.13 & 1985 \\
14 & Bird Body & 7.35 & 6.88 & 1922 \\
15 & Bird Wing & 7.66 & 6.32 & 512 \\
16 & Bird Foot & 0.84 & 0.83 & 1435 \\
17 & Bird Tail & 1.57 & 1.69 & 1585 \\
18 & Snake Head & 5.50 & 9.10 & 2312 \\
19 & Snake Body & 24.48 & 19.88 & 2360 \\
20 & Reptile Head & 7.95 & 12.07 & 3024 \\
21 & Reptile Body & 11.12 & 10.04 & 2958 \\
22 & Reptile Foot & 4.79 & 4.33 & 2821 \\
23 & Reptile Tail & 2.69 & 2.72 & 1349 \\
24 & Car Body & 31.90 & 18.74 & 1878 \\
25 & Car Tier & 3.84 & 3.98 & 1724 \\
26 & Car Side Mirror & 0.47 & 1.24 & 1258 \\
27 & Bicycle Body & 5.98 & 6.90 & 517 \\
28 & Bicycle Head & 1.33 & 1.73 & 415 \\
29 & Bicycle Seat & 0.98 & 1.48 & 354 \\
30 & Bicycle Tier & 7.69 & 8.11 & 531 \\
31 & Boat Body & 9.43 & 10.71 & 470 \\
32 & Boat Sail & 15.00 & 10.92 & 363 \\
33 & Aeroplane Head & 3.22 & 5.98 & 250 \\
34 & Aeroplane Body & 7.87 & 8.17 & 259 \\
35 & Aeroplane Engine & 1.99 & 2.55 & 218 \\
36 & Aeroplane Wing & 3.67 & 3.51 & 240 \\
37 & Aeroplane Tail & 1.97 & 2.42 & 226 \\
38 & Bottle Mouth & 5.13 & 10.67 & 341 \\
39 & Bottle Body & 23.78 & 17.26 & 402 \\
\end{longtable}
\endgroup

\begingroup
\setlength{\LTleft}{0pt}
\setlength{\LTright}{0pt}
\setlength{\LTcapwidth}{\linewidth}
\setlength{\tabcolsep}{3pt}
\begin{longtable}{@{}>{\raggedright\arraybackslash}p{0.12\linewidth}>{\raggedright\arraybackslash}p{0.34\linewidth}>{\centering\arraybackslash}p{0.15\linewidth}>{\centering\arraybackslash}p{0.15\linewidth}>{\centering\arraybackslash}p{0.12\linewidth}@{}}
\caption{CUB training split: per-class bird-mask area statistics (percentage of image area).}\label{tab:cub-class-sizes-train}\\
\toprule
\textbf{Class ID} & \textbf{Class Name} & \textbf{Mean Mask Ratio (\%)} & \textbf{Std. Dev. (\%)} & \textbf{Images} \\
\midrule
\endfirsthead
\toprule
\textbf{Class ID} & \textbf{Class Name} & \textbf{Mean Mask Ratio (\%)} & \textbf{Std. Dev. (\%)} & \textbf{Images} \\
\midrule
\endhead
\midrule \multicolumn{5}{r}{\emph{Continued on next page}}\\
\endfoot
\bottomrule
\endlastfoot
0 & Black\_\allowbreak{}footed\_\allowbreak{}Albatross & 12.85 & 8.70 & 26 \\
1 & Laysan\_\allowbreak{}Albatross & 17.16 & 10.28 & 22 \\
2 & Sooty\_\allowbreak{}Albatross & 13.57 & 12.75 & 23 \\
3 & Groove\_\allowbreak{}billed\_\allowbreak{}Ani & 10.90 & 5.89 & 24 \\
4 & Crested\_\allowbreak{}Auklet & 19.00 & 11.31 & 23 \\
5 & Least\_\allowbreak{}Auklet & 19.26 & 11.41 & 27 \\
6 & Parakeet\_\allowbreak{}Auklet & 19.11 & 9.97 & 28 \\
7 & Rhinoceros\_\allowbreak{}Auklet & 18.34 & 15.01 & 24 \\
8 & Brewer\_\allowbreak{}Blackbird & 15.43 & 7.40 & 21 \\
9 & Red\_\allowbreak{}winged\_\allowbreak{}Blackbird & 12.13 & 7.73 & 22 \\
10 & Rusty\_\allowbreak{}Blackbird & 12.03 & 6.08 & 25 \\
11 & Yellow\_\allowbreak{}headed\_\allowbreak{}Blackbird & 9.88 & 5.46 & 26 \\
12 & Bobolink & 10.85 & 5.44 & 27 \\
13 & Indigo\_\allowbreak{}Bunting & 9.79 & 7.30 & 23 \\
14 & Lazuli\_\allowbreak{}Bunting & 8.08 & 3.23 & 22 \\
15 & Painted\_\allowbreak{}Bunting & 10.79 & 5.87 & 24 \\
16 & Cardinal & 13.01 & 5.96 & 24 \\
17 & Spotted\_\allowbreak{}Catbird & 18.57 & 9.12 & 25 \\
18 & Gray\_\allowbreak{}Catbird & 12.82 & 5.09 & 22 \\
19 & Yellow\_\allowbreak{}breasted\_\allowbreak{}Chat & 7.68 & 4.11 & 28 \\
20 & Eastern\_\allowbreak{}Towhee & 12.40 & 6.69 & 27 \\
21 & Chuck\_\allowbreak{}will\_\allowbreak{}Widow & 20.53 & 9.56 & 23 \\
22 & Brandt\_\allowbreak{}Cormorant & 15.47 & 11.44 & 27 \\
23 & Red\_\allowbreak{}faced\_\allowbreak{}Cormorant & 16.09 & 10.67 & 22 \\
24 & Pelagic\_\allowbreak{}Cormorant & 12.48 & 6.56 & 27 \\
25 & Bronzed\_\allowbreak{}Cowbird & 13.69 & 7.38 & 24 \\
26 & Shiny\_\allowbreak{}Cowbird & 10.70 & 5.19 & 22 \\
27 & Brown\_\allowbreak{}Creeper & 10.08 & 3.91 & 23 \\
28 & American\_\allowbreak{}Crow & 15.68 & 6.49 & 23 \\
29 & Fish\_\allowbreak{}Crow & 17.91 & 9.91 & 23 \\
30 & Black\_\allowbreak{}billed\_\allowbreak{}Cuckoo & 13.51 & 8.67 & 22 \\
31 & Mangrove\_\allowbreak{}Cuckoo & 11.15 & 5.21 & 19 \\
32 & Yellow\_\allowbreak{}billed\_\allowbreak{}Cuckoo & 9.64 & 5.54 & 28 \\
33 & Gray\_\allowbreak{}crowned\_\allowbreak{}Rosy\_\allowbreak{}Finch & 14.62 & 6.89 & 20 \\
34 & Purple\_\allowbreak{}Finch & 12.39 & 5.08 & 21 \\
35 & Northern\_\allowbreak{}Flicker & 16.10 & 6.33 & 22 \\
36 & Acadian\_\allowbreak{}Flycatcher & 11.91 & 7.41 & 24 \\
37 & Great\_\allowbreak{}Crested\_\allowbreak{}Flycatcher & 11.95 & 7.51 & 25 \\
38 & Least\_\allowbreak{}Flycatcher & 13.74 & 9.28 & 23 \\
39 & Olive\_\allowbreak{}sided\_\allowbreak{}Flycatcher & 9.08 & 4.60 & 24 \\
40 & Scissor\_\allowbreak{}tailed\_\allowbreak{}Flycatcher & 5.44 & 3.56 & 22 \\
41 & Vermilion\_\allowbreak{}Flycatcher & 10.02 & 5.66 & 16 \\
42 & Yellow\_\allowbreak{}bellied\_\allowbreak{}Flycatcher & 14.98 & 7.86 & 27 \\
43 & Frigatebird & 15.49 & 13.84 & 23 \\
44 & Northern\_\allowbreak{}Fulmar & 13.10 & 8.42 & 25 \\
45 & Gadwall & 12.39 & 6.45 & 22 \\
46 & American\_\allowbreak{}Goldfinch & 11.55 & 6.12 & 23 \\
47 & European\_\allowbreak{}Goldfinch & 11.37 & 7.59 & 22 \\
48 & Boat\_\allowbreak{}tailed\_\allowbreak{}Grackle & 12.98 & 7.23 & 22 \\
49 & Eared\_\allowbreak{}Grebe & 9.86 & 4.19 & 26 \\
50 & Horned\_\allowbreak{}Grebe & 10.74 & 3.94 & 22 \\
51 & Pied\_\allowbreak{}billed\_\allowbreak{}Grebe & 9.93 & 5.63 & 23 \\
52 & Western\_\allowbreak{}Grebe & 9.37 & 5.18 & 27 \\
53 & Blue\_\allowbreak{}Grosbeak & 9.37 & 4.73 & 23 \\
54 & Evening\_\allowbreak{}Grosbeak & 12.47 & 4.79 & 28 \\
55 & Pine\_\allowbreak{}Grosbeak & 13.25 & 5.43 & 27 \\
56 & Rose\_\allowbreak{}breasted\_\allowbreak{}Grosbeak & 14.61 & 8.00 & 20 \\
57 & Pigeon\_\allowbreak{}Guillemot & 13.92 & 10.29 & 23 \\
58 & California\_\allowbreak{}Gull & 11.97 & 6.89 & 23 \\
59 & Glaucous\_\allowbreak{}winged\_\allowbreak{}Gull & 16.25 & 8.20 & 26 \\
60 & Heermann\_\allowbreak{}Gull & 16.50 & 8.90 & 26 \\
61 & Herring\_\allowbreak{}Gull & 11.63 & 5.26 & 20 \\
62 & Ivory\_\allowbreak{}Gull & 11.95 & 5.77 & 27 \\
63 & Ring\_\allowbreak{}billed\_\allowbreak{}Gull & 15.94 & 6.84 & 26 \\
64 & Slaty\_\allowbreak{}backed\_\allowbreak{}Gull & 21.49 & 9.54 & 25 \\
65 & Western\_\allowbreak{}Gull & 15.63 & 5.83 & 21 \\
66 & Anna\_\allowbreak{}Hummingbird & 11.66 & 5.89 & 21 \\
67 & Ruby\_\allowbreak{}throated\_\allowbreak{}Hummingbird & 10.56 & 6.28 & 23 \\
68 & Rufous\_\allowbreak{}Hummingbird & 10.72 & 5.37 & 22 \\
69 & Green\_\allowbreak{}Violetear & 12.46 & 5.79 & 24 \\
70 & Long\_\allowbreak{}tailed\_\allowbreak{}Jaeger & 10.03 & 4.49 & 22 \\
71 & Pomarine\_\allowbreak{}Jaeger & 13.67 & 6.54 & 24 \\
72 & Blue\_\allowbreak{}Jay & 12.90 & 6.89 & 26 \\
73 & Florida\_\allowbreak{}Jay & 10.95 & 6.73 & 27 \\
74 & Green\_\allowbreak{}Jay & 14.32 & 8.10 & 25 \\
75 & Dark\_\allowbreak{}eyed\_\allowbreak{}Junco & 14.89 & 9.80 & 26 \\
76 & Tropical\_\allowbreak{}Kingbird & 9.67 & 4.83 & 24 \\
77 & Gray\_\allowbreak{}Kingbird & 10.00 & 4.88 & 23 \\
78 & Belted\_\allowbreak{}Kingfisher & 8.70 & 4.40 & 19 \\
79 & Green\_\allowbreak{}Kingfisher & 8.83 & 5.60 & 28 \\
80 & Pied\_\allowbreak{}Kingfisher & 9.64 & 5.43 & 26 \\
81 & Ringed\_\allowbreak{}Kingfisher & 8.72 & 4.28 & 27 \\
82 & White\_\allowbreak{}breasted\_\allowbreak{}Kingfisher & 10.08 & 6.51 & 19 \\
83 & Red\_\allowbreak{}legged\_\allowbreak{}Kittiwake & 16.44 & 13.23 & 24 \\
84 & Horned\_\allowbreak{}Lark & 12.09 & 6.43 & 27 \\
85 & Pacific\_\allowbreak{}Loon & 12.73 & 11.73 & 27 \\
86 & Mallard & 20.37 & 6.47 & 27 \\
87 & Western\_\allowbreak{}Meadowlark & 11.96 & 8.37 & 24 \\
88 & Hooded\_\allowbreak{}Merganser & 14.79 & 7.67 & 23 \\
89 & Red\_\allowbreak{}breasted\_\allowbreak{}Merganser & 8.56 & 3.65 & 22 \\
90 & Mockingbird & 12.29 & 7.99 & 26 \\
91 & Nighthawk & 14.13 & 8.53 & 23 \\
92 & Clark\_\allowbreak{}Nutcracker & 13.56 & 6.09 & 25 \\
93 & White\_\allowbreak{}breasted\_\allowbreak{}Nuthatch & 12.71 & 6.95 & 22 \\
94 & Baltimore\_\allowbreak{}Oriole & 10.63 & 5.73 & 26 \\
95 & Hooded\_\allowbreak{}Oriole & 11.06 & 5.42 & 25 \\
96 & Orchard\_\allowbreak{}Oriole & 7.98 & 4.39 & 28 \\
97 & Scott\_\allowbreak{}Oriole & 11.77 & 7.27 & 23 \\
98 & Ovenbird & 11.01 & 5.28 & 26 \\
99 & Brown\_\allowbreak{}Pelican & 18.18 & 7.81 & 25 \\
100 & White\_\allowbreak{}Pelican & 17.88 & 12.65 & 22 \\
101 & Western\_\allowbreak{}Wood\_\allowbreak{}Pewee & 10.33 & 3.50 & 24 \\
102 & Sayornis & 10.66 & 4.69 & 26 \\
103 & American\_\allowbreak{}Pipit & 9.31 & 4.92 & 24 \\
104 & Whip\_\allowbreak{}poor\_\allowbreak{}Will & 21.82 & 12.72 & 26 \\
105 & Horned\_\allowbreak{}Puffin & 17.96 & 11.08 & 27 \\
106 & Common\_\allowbreak{}Raven & 14.91 & 6.70 & 25 \\
107 & White\_\allowbreak{}necked\_\allowbreak{}Raven & 15.97 & 10.60 & 26 \\
108 & American\_\allowbreak{}Redstart & 11.17 & 4.63 & 22 \\
109 & Geococcyx & 10.92 & 5.88 & 23 \\
110 & Loggerhead\_\allowbreak{}Shrike & 8.62 & 3.89 & 23 \\
111 & Great\_\allowbreak{}Grey\_\allowbreak{}Shrike & 11.05 & 5.45 & 23 \\
112 & Baird\_\allowbreak{}Sparrow & 14.12 & 9.32 & 21 \\
113 & Black\_\allowbreak{}throated\_\allowbreak{}Sparrow & 10.54 & 6.06 & 24 \\
114 & Brewer\_\allowbreak{}Sparrow & 11.26 & 5.32 & 25 \\
115 & Chipping\_\allowbreak{}Sparrow & 9.37 & 6.14 & 26 \\
116 & Clay\_\allowbreak{}colored\_\allowbreak{}Sparrow & 10.59 & 5.32 & 26 \\
117 & House\_\allowbreak{}Sparrow & 11.72 & 4.43 & 25 \\
118 & Field\_\allowbreak{}Sparrow & 11.08 & 5.89 & 27 \\
119 & Fox\_\allowbreak{}Sparrow & 15.10 & 6.49 & 25 \\
120 & Grasshopper\_\allowbreak{}Sparrow & 12.63 & 8.82 & 25 \\
121 & Harris\_\allowbreak{}Sparrow & 11.74 & 5.19 & 19 \\
122 & Henslow\_\allowbreak{}Sparrow & 12.45 & 6.05 & 24 \\
123 & Le\_\allowbreak{}Conte\_\allowbreak{}Sparrow & 12.63 & 6.98 & 25 \\
124 & Lincoln\_\allowbreak{}Sparrow & 12.12 & 5.38 & 22 \\
125 & Nelson\_\allowbreak{}Sharp\_\allowbreak{}tailed\_\allowbreak{}Sparrow & 9.77 & 6.29 & 20 \\
126 & Savannah\_\allowbreak{}Sparrow & 12.06 & 5.34 & 26 \\
127 & Seaside\_\allowbreak{}Sparrow & 12.16 & 6.39 & 22 \\
128 & Song\_\allowbreak{}Sparrow & 14.71 & 5.46 & 20 \\
129 & Tree\_\allowbreak{}Sparrow & 15.73 & 8.10 & 21 \\
130 & Vesper\_\allowbreak{}Sparrow & 9.76 & 7.56 & 21 \\
131 & White\_\allowbreak{}crowned\_\allowbreak{}Sparrow & 13.08 & 6.32 & 24 \\
132 & White\_\allowbreak{}throated\_\allowbreak{}Sparrow & 14.66 & 5.86 & 24 \\
133 & Cape\_\allowbreak{}Glossy\_\allowbreak{}Starling & 13.19 & 5.82 & 23 \\
134 & Bank\_\allowbreak{}Swallow & 9.86 & 4.40 & 26 \\
135 & Barn\_\allowbreak{}Swallow & 13.79 & 8.81 & 28 \\
136 & Cliff\_\allowbreak{}Swallow & 11.95 & 6.52 & 22 \\
137 & Tree\_\allowbreak{}Swallow & 11.72 & 6.71 & 25 \\
138 & Scarlet\_\allowbreak{}Tanager & 9.45 & 5.71 & 24 \\
139 & Summer\_\allowbreak{}Tanager & 10.54 & 4.06 & 26 \\
140 & Artic\_\allowbreak{}Tern & 10.94 & 4.92 & 23 \\
141 & Black\_\allowbreak{}Tern & 11.47 & 5.84 & 20 \\
142 & Caspian\_\allowbreak{}Tern & 9.58 & 6.44 & 25 \\
143 & Common\_\allowbreak{}Tern & 8.50 & 6.51 & 22 \\
144 & Elegant\_\allowbreak{}Tern & 10.98 & 4.67 & 23 \\
145 & Forsters\_\allowbreak{}Tern & 10.77 & 5.58 & 26 \\
146 & Least\_\allowbreak{}Tern & 10.51 & 5.65 & 24 \\
147 & Green\_\allowbreak{}tailed\_\allowbreak{}Towhee & 11.46 & 5.47 & 21 \\
148 & Brown\_\allowbreak{}Thrasher & 9.71 & 5.95 & 25 \\
149 & Sage\_\allowbreak{}Thrasher & 11.74 & 4.89 & 25 \\
150 & Black\_\allowbreak{}capped\_\allowbreak{}Vireo & 16.58 & 14.03 & 24 \\
151 & Blue\_\allowbreak{}headed\_\allowbreak{}Vireo & 9.93 & 6.54 & 24 \\
152 & Philadelphia\_\allowbreak{}Vireo & 12.85 & 5.92 & 21 \\
153 & Red\_\allowbreak{}eyed\_\allowbreak{}Vireo & 11.08 & 3.43 & 26 \\
154 & Warbling\_\allowbreak{}Vireo & 9.35 & 4.68 & 24 \\
155 & White\_\allowbreak{}eyed\_\allowbreak{}Vireo & 11.15 & 4.54 & 21 \\
156 & Yellow\_\allowbreak{}throated\_\allowbreak{}Vireo & 11.60 & 10.26 & 20 \\
157 & Bay\_\allowbreak{}breasted\_\allowbreak{}Warbler & 11.54 & 5.06 & 20 \\
158 & Black\_\allowbreak{}and\_\allowbreak{}white\_\allowbreak{}Warbler & 12.13 & 5.64 & 28 \\
159 & Black\_\allowbreak{}throated\_\allowbreak{}Blue\_\allowbreak{}Warbler & 10.63 & 4.46 & 23 \\
160 & Blue\_\allowbreak{}winged\_\allowbreak{}Warbler & 10.72 & 8.78 & 22 \\
161 & Canada\_\allowbreak{}Warbler & 11.71 & 11.25 & 25 \\
162 & Cape\_\allowbreak{}May\_\allowbreak{}Warbler & 9.76 & 6.14 & 23 \\
163 & Cerulean\_\allowbreak{}Warbler & 13.30 & 7.07 & 21 \\
164 & Chestnut\_\allowbreak{}sided\_\allowbreak{}Warbler & 12.56 & 10.81 & 25 \\
165 & Golden\_\allowbreak{}winged\_\allowbreak{}Warbler & 14.05 & 7.89 & 26 \\
166 & Hooded\_\allowbreak{}Warbler & 10.89 & 6.45 & 24 \\
167 & Kentucky\_\allowbreak{}Warbler & 12.69 & 8.03 & 27 \\
168 & Magnolia\_\allowbreak{}Warbler & 11.35 & 4.37 & 22 \\
169 & Mourning\_\allowbreak{}Warbler & 13.01 & 6.57 & 26 \\
170 & Myrtle\_\allowbreak{}Warbler & 12.44 & 6.98 & 27 \\
171 & Nashville\_\allowbreak{}Warbler & 9.91 & 6.20 & 25 \\
172 & Orange\_\allowbreak{}crowned\_\allowbreak{}Warbler & 11.55 & 6.49 & 25 \\
173 & Palm\_\allowbreak{}Warbler & 10.19 & 5.23 & 21 \\
174 & Pine\_\allowbreak{}Warbler & 13.44 & 5.18 & 22 \\
175 & Prairie\_\allowbreak{}Warbler & 10.14 & 5.74 & 26 \\
176 & Prothonotary\_\allowbreak{}Warbler & 9.63 & 6.57 & 24 \\
177 & Swainson\_\allowbreak{}Warbler & 16.86 & 12.98 & 26 \\
178 & Tennessee\_\allowbreak{}Warbler & 10.25 & 4.26 & 25 \\
179 & Wilson\_\allowbreak{}Warbler & 10.27 & 4.63 & 29 \\
180 & Worm\_\allowbreak{}eating\_\allowbreak{}Warbler & 13.47 & 7.01 & 25 \\
181 & Yellow\_\allowbreak{}Warbler & 9.26 & 5.64 & 26 \\
182 & Northern\_\allowbreak{}Waterthrush & 9.87 & 4.14 & 22 \\
183 & Louisiana\_\allowbreak{}Waterthrush & 11.16 & 6.15 & 24 \\
184 & Bohemian\_\allowbreak{}Waxwing & 13.35 & 5.29 & 25 \\
185 & Cedar\_\allowbreak{}Waxwing & 11.38 & 5.97 & 24 \\
186 & American\_\allowbreak{}Three\_\allowbreak{}toed\_\allowbreak{}Woodpecker & 15.76 & 5.48 & 27 \\
187 & Pileated\_\allowbreak{}Woodpecker & 11.89 & 5.35 & 23 \\
188 & Red\_\allowbreak{}bellied\_\allowbreak{}Woodpecker & 11.85 & 5.20 & 20 \\
189 & Red\_\allowbreak{}cockaded\_\allowbreak{}Woodpecker & 11.68 & 6.37 & 25 \\
190 & Red\_\allowbreak{}headed\_\allowbreak{}Woodpecker & 10.41 & 5.14 & 25 \\
191 & Downy\_\allowbreak{}Woodpecker & 17.83 & 6.64 & 26 \\
192 & Bewick\_\allowbreak{}Wren & 14.57 & 9.61 & 20 \\
193 & Cactus\_\allowbreak{}Wren & 12.74 & 6.11 & 23 \\
194 & Carolina\_\allowbreak{}Wren & 14.86 & 7.29 & 24 \\
195 & House\_\allowbreak{}Wren & 10.63 & 6.08 & 25 \\
196 & Marsh\_\allowbreak{}Wren & 11.20 & 4.72 & 24 \\
197 & Rock\_\allowbreak{}Wren & 10.32 & 3.30 & 24 \\
198 & Winter\_\allowbreak{}Wren & 10.50 & 5.31 & 26 \\
199 & Common\_\allowbreak{}Yellowthroat & 13.04 & 7.94 & 26 \\
\end{longtable}
\endgroup

\end{document}